%% file: main.tex
\documentclass[10pt,twocolumn,letterpaper]{article}

\input{header}
\edef\loadcvpr{\noexpand\usepackage[\paperoptions]{cvpr}}\loadcvpr

\input{preamble}

\ifarxiv\else\usepackage{xr-hyper}\fi
\definecolor{cvprblue}{rgb}{0.21,0.49,0.74}
\usepackage[pagebackref,breaklinks,colorlinks,allcolors=cvprblue]{hyperref}
\ifarxiv\else\fi
\ifx\papermode\papermodereview\else\hypersetup{urlcolor=magenta}\fi
\ifarxiv\usepackage{placeins}\fi

\def\confName{CVPR}
\def\confYear{2027}

\title{\papertitle}
\author{\paperauthors}

\begin{document}
\twocolumn[{%
\renewcommand\twocolumn[1][]{#1}%
\maketitle
\input{figures/teaser}
}]
\input{sec/0_abstract}
\input{sec/1_intro}
\input{sec/2_related}
\input{sec/3_background}
\input{sec/4_method}

\input{sec/5_experiments}
\input{sec/6_conclusion}
\label{sec:body_end}
\clearpage
\ifx\papermode\papermodereview\else\input{sec/8_acknowledgements}\fi
\input{sec/7_llm_usage}
{
    \small
    \bibliographystyle{ieeenat_fullname}
    \bibliography{main}
}

\ifarxiv
  \maketitlesupplementary
  \appendix
  \supplementarynumbering
  \setlength{\textfloatsep}{12pt plus 2pt minus 2pt}
  \setlength{\dbltextfloatsep}{12pt plus 2pt minus 2pt}
  \input{sec/X_suppl}

\fi

\end{document}

%% file: header.tex
\input{mode}
\def\papermodereview{review}

\def\papermodearxiv{arxiv}
\newif\ifarxiv
\ifx\papermode\papermodearxiv\arxivtrue\fi

\newcommand{\paperoptions}{%
  \ifx\papermode\papermodereview review\else
  \ifx\papermode\papermodearxiv pagenumbers\fi\fi}

\newcommand{\supplementarynumbering}{%
  \setcounter{figure}{0}\setcounter{table}{0}\setcounter{equation}{0}%
  \renewcommand{\thefigure}{S\arabic{figure}}%
  \renewcommand{\thetable}{S\arabic{table}}%
  \renewcommand{\theequation}{S\arabic{equation}}%
  \renewcommand{\theHfigure}{S\arabic{figure}}%
  \renewcommand{\theHtable}{S\arabic{table}}%
  \renewcommand{\theHequation}{S\arabic{equation}}}

\AtBeginDocument{%
  \crefname{appendix}{Sec.}{Secs.}%
  \Crefname{appendix}{Section}{Sections}}
\newcommand{\mainref}[1]{\ifarxiv\cref{#1}\else main paper \cref*{M-#1}\fi}

\newcommand{\suppref}[1]{\ifarxiv\cref{#1}\else supplementary \cref*{S-#1}\fi}
\newcommand{\Suppref}[1]{\ifarxiv\Cref{#1}\else Supplementary \cref*{S-#1}\fi}

\newcommand{\projecturl}{\url{https://memo-4d.github.io}}
\newcommand{\resultspage}{%
  \ifx\papermode\papermodereview the HTML page included in this supplementary material\else
  \url{https://memo-4d.github.io/comparisons.html}\fi}
\newcommand{\abstractlink}{%
  \ifx\papermode\papermodereview\else\par\noindent Project page: \projecturl\fi}

\newcommand{\papertitle}{A Mesh for Every Moment}

\newcommand{\paperauthors}{Roy Or-El\textsuperscript{1} \qquad
        Jeong Joon Park\textsuperscript{2} \qquad
        Ira Kemelmacher-Shlizerman\textsuperscript{1} \\
        \textsuperscript{1}University of Washington \qquad
        \textsuperscript{2}University of Michigan}

%% file: mode.tex
\def\papermode{arxiv}

%% file: preamble.tex
\usepackage{microtype}

\usepackage{xspace}
\usepackage{array}
\usepackage{colortbl}
\definecolor{rankgold}{RGB}{255,217,102}
\definecolor{ranksilver}{RGB}{217,217,217}
\definecolor{rankbronze}{RGB}{240,192,150}
\newcommand{\gold}[1]{\cellcolor{rankgold}#1}
\newcommand{\silver}[1]{\cellcolor{ranksilver}#1}
\newcommand{\bronze}[1]{\cellcolor{rankbronze}#1}
\newcommand{\swatch}[1]{\textcolor{#1}{\rule[-0.1em]{1.2em}{0.8em}}}
\newcommand{\ourmethod}{MEMO\xspace}
\newcommand{\benchmark}{Blender-Moments\xspace}

\makeatletter
\renewcommand\paragraph{\@startsection{paragraph}{4}{\z@}%
  {1ex \@plus 0.5ex \@minus 0.2ex}{-1em}{\normalfont\normalsize\bfseries}}
\makeatother

\usepackage{tikz}
\usetikzlibrary{arrows.meta,calc,fit,positioning}

%% file: figures/teaser.tex
\begin{center}
  \captionsetup{type=figure}
  \includegraphics[width=\linewidth]{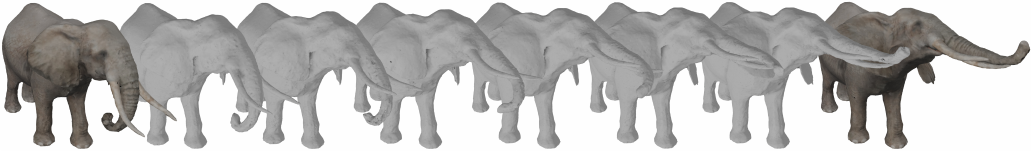}\\[2pt]
  \includegraphics[width=\linewidth]{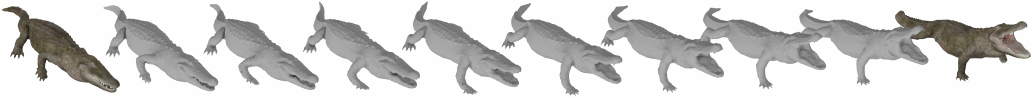}\\[2pt]
  \includegraphics[width=\linewidth]{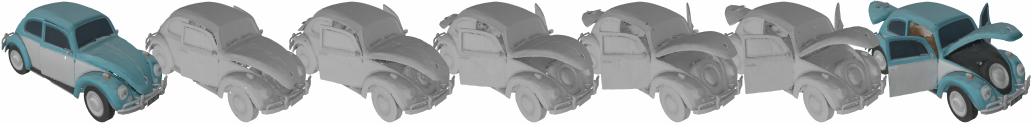}\\[2pt]
  \begin{tikzpicture}
    \draw[line width=1.2pt, -{Latex[length=6pt, width=5pt]}] (0, 0) -- (\linewidth, 0)
      node[midway, fill=white, inner xsep=4pt, inner ysep=0pt, font=\footnotesize] {Time};
  \end{tikzpicture}
  \caption{\ourmethod reconstructs a mesh of a dynamic scene at every frame, directly from its Gaussians.
    Shown are three \benchmark scenes over time, with the first and last frames in vertex colors and the frames between in gray.}
  \label{fig:teaser}
\end{center}

%% file: sec/0_abstract.tex
\begin{abstract}
We present \ourmethod (a Mesh for Every MOment), a method that reconstructs the geometry of complete dynamic scenes, captured by multiple synchronized cameras, and extracts a mesh at every frame.
Existing dynamic Gaussian splatting methods either model appearance alone or target isolated objects segmented by foreground masks, and all of them recover meshes by TSDF fusion or Poisson reconstruction, whose fixed grids limit the detail they can capture.
At the core of \ourmethod is a per-Gaussian motion model inspired by Gaussian processes, which represents each Gaussian's position and rotation over time as a combination of temporal Gaussian kernels.
The model keeps every Gaussian a valid surface element throughout the sequence and carries an implicit representation of the dynamic geometry, so a mesh can be derived at any time from a field defined by the Gaussians.
Although common in real captures, large and fast motion poses a challenge that dynamic Gaussian splatting methods usually overlook.
To account for it, we design a track-based initialization scheme that fits each Gaussian's initial trajectory to point tracks in closed form.
We evaluate \ourmethod on \benchmark, a new benchmark of nine animated scenes with ground-truth meshes at every frame, and on two real-world datasets.
\ourmethod reconstructs the most accurate geometry among state-of-the-art methods, and remains competitive in novel-view synthesis and training time.
\abstractlink
\vspace*{-10pt}
\end{abstract}

%% file: sec/1_intro.tex
\section{Introduction}
\label{sec:intro}

Capturing a dynamic scene, a person in motion, a mechanism at work, or an animal on the move, and turning it into a digital asset is the first step toward bringing the physical world into the virtual world.
It is a long-standing goal of computer vision and graphics, and recent progress in novel-view synthesis has brought it within reach.
Neural radiance fields~\cite{mildenhall2020nerf} first reconstructed static scenes from photographs with striking realism, and 3D Gaussian splatting~\cite{kerbl20233d} has since made it fast enough to become the representation of choice.
Dynamic variants of Gaussian splatting bring this to videos, allowing immersive rendering of the scene from novel views and camera trajectories.
However, these dynamic methods optimize for appearance, and the surfaces extracted from them are often noisy and incomplete.
This matters for downstream applications that need accurate per-frame geometry, like physical interaction with virtual objects, relighting, and AR/VR scene composition.

Current dynamic Gaussian methods, whether built on deformation fields~\cite{wu20244d,yang2024deformable}, 4D primitives~\cite{yang2024realtime}, sparse control points~\cite{huang2024scgs}, or per-Gaussian trajectories~\cite{luiten2024dynamic,li2024spacetime,wang2025freetimegs,tzathas2026adaptive}, encode appearance and motion but do not have a representation of the surface, either implicit or explicit.
To reconstruct the scene's surfaces, a volume has to be built from the trained model after the fact, by TSDF volume integration~\cite{curless1996volumetric} on depth maps rendered from many viewpoints and passed to marching cubes~\cite{lorensen1987marching} on a fixed-resolution grid.
Since the optimization constrains geometry only through appearance, the models carry many floaters, which distort the input volumes of the marching cubes and lead to noisy reconstructions.

For static scenes, this problem has been addressed by training Gaussians to represent geometry alongside appearance.
One family of methods regularizes the Gaussians so that their rendered depth, and hence its fusion, is accurate~\cite{huang20242d,chen2025pgsr,zhang2026radegs}.
A second family defines a field directly from the Gaussians and extracts its level set with marching tetrahedra~\cite{yu2024gaussian,radl2025sof,guedon2025milo,gomez2026blobs}, which removes depth fusion altogether and adapts the mesh resolution to the Gaussians.
For dynamic scenes, the few methods that reconstruct surfaces~\cite{liu2025dynamic,zhang2025dynamic,wu20264dsurf} are object-centric, relying on foreground masks to isolate a single object, and remain slow to train.

In this work, we aim to bring the field-based extraction of the second family to complete, unmasked dynamic scenes captured by multiple synchronized cameras, allowing the full scene geometry to be extracted at every frame.
We build on three key insights.
First, every Gaussian must remain a valid surface element at every frame of the sequence.
Second, the model should carry an implicit representation of the scene's dynamic geometry.
Third, training struggles to recover large and fast motion without a good starting point.
To realize the first two insights, we propose a per-Gaussian temporal motion model that represents each Gaussian's changes in position and rotation over time as a linear combination of temporal Gaussian kernels, inspired by radial basis function (RBF) networks~\cite{broomhead1988multivariable,moody1989fast} and Gaussian processes (GPs)~\cite{rasmussen2006gaussian}.
To realize the third, we introduce a track-based initialization scheme that assigns each initial Gaussian a trajectory consistent with the observed motion.

Since the temporal Gaussian kernels span the whole timeline, each Gaussian can follow complex motion and remains a valid surface element throughout the sequence.
It can therefore carry an implicit representation of the surface, such as oriented normals~\cite{gomez2026blobs}, along its trajectory.
Polynomial trajectories~\cite{li2024spacetime,wang2025freetimegs}, in contrast, hold only within short time windows and introduce geometric noise outside them.
Since each trajectory is linear in its kernel weights, our initialization scheme fits it in closed form by GP regression to point tracks spanning the whole sequence, so even large and fast motion is captured from the start.
With our model, the surface is defined at any time, and a mesh follows at every frame from marching tetrahedra, with no depth rendering or fusion (\cref{fig:teaser}).

Evaluating dynamic geometry requires ground-truth meshes that evolve with the scene, and no public dataset pairs synchronized multi-camera captures with exact time-varying geometry.
We therefore build \benchmark, a benchmark of nine animated scenes, rendered from multiple synchronized cameras, whose ground-truth meshes can be extracted at any frame.
We evaluate our method on this new dataset and on two existing real-world datasets, Neural 3D Video~\cite{li2022neural} and SelfCap~\cite{xu2024representing}, and compare it to state-of-the-art dynamic methods, including ones designed for surface reconstruction.
\ourmethod reconstructs the most accurate dynamic geometry, and its novel-view synthesis rivals that of the strongest baselines.
Its advantage is largest on scenes with large and fast motion.
It also trains in a time comparable to the fastest dynamic splatting methods.

In summary, our contributions are:
\begin{itemize}
  \item A framework for joint appearance and per-frame mesh reconstruction of complete dynamic scenes, achieving state-of-the-art surface reconstruction.
  \item A novel motion model and initialization technique inspired by Gaussian processes.
  \item A multi-view benchmark for dynamic surface reconstruction, with ground-truth meshes at every frame.
\end{itemize}

%% file: sec/2_related.tex
\section{Related Work}
\label{sec:related}

\paragraph{Static scenes.}
\label{par:related_static}
Novel-view synthesis methods such as neural radiance fields~\cite{mildenhall2020nerf} and 3D Gaussian splatting~\cite{kerbl20233d} are optimized for appearance alone and do not model the scene's geometry.
Many works have addressed this by extending these representations to recover surfaces.
For radiance fields, several works~\cite{wang2021neus,yariv2021volume,li2023neuralangelo,wang2023neus2} map an implicit surface representation to volume density, so that volume rendering recovers the surface together with appearance.
For Gaussians, two classes of methods have emerged.
The first regularizes the Gaussians so that their rendered depth is accurate, and extracts meshes by TSDF volume integration of that depth.
2DGS~\cite{huang20242d} pioneered this class by flattening the Gaussians into 2D surfels with well-defined normals.
Later works adopt planar~\cite{chen2025pgsr} or quadratic~\cite{zhang2025quadratic} primitives, rasterize depth and normals in closed form~\cite{zhang2026radegs}, supervise normals with monocular priors~\cite{chen2024vcrgaus}, or treat Gaussians as stochastic solids~\cite{zhang2026geometry}.

The second class defines a field directly from the Gaussians and extracts its level set with marching tetrahedra on a Delaunay tetrahedralization of Gaussian-derived points.
GOF~\cite{yu2024gaussian} introduced this approach with an opacity field, which SOF~\cite{radl2025sof} evaluates faster by sorting the Gaussians along each ray.
MILo~\cite{guedon2025milo} learns signed distances at nine pivots per Gaussian and extracts a mesh during training, whose rendered depth and normals regularize the Gaussians.
Gaussian Wrapping~\cite{gomez2026blobs} treats each Gaussian as an oriented surface element whose vacancy field yields detailed meshes.
Other works jointly optimize the Gaussians with a neural signed~\cite{chen2023neusg,yu2024gsdf,lyu20243dgsr,zhang2024neural} or unsigned~\cite{li2025gaussianudf} distance field, or align the Gaussians with the surface and reconstruct a mesh by Poisson reconstruction~\cite{guedon2024sugar}.
All of these methods assume a static scene.

\paragraph{Dynamic scenes.}
\label{par:related_dynamic}
Both radiance fields and Gaussian splatting have been extended from static to dynamic scenes.
For radiance fields, a common approach~\cite{pumarola2021dnerf,park2021nerfies,tretschk2021nonrigid,park2021hypernerf} is to warp rays into a canonical space with a learned deformation field.
Other works condition the field on time~\cite{lombardi2019neural,li2022neural} or factorize it into space-time feature planes~\cite{fridovichkeil2023kplanes,cao2023hexplane}.
For Gaussians, 4DGS~\cite{wu20244d} and Deformable-3DGS~\cite{yang2024deformable} carry over the canonical-deformation approach~\cite{pumarola2021dnerf,park2021nerfies,tretschk2021nonrigid,park2021hypernerf}, and many later methods follow it~\cite{liang2025gaufre,bae2024pergaussian,zhu2024motiongs}.
Other methods drive the Gaussians by blending the rigid motions of a sparse set of control nodes~\cite{huang2024scgs,lei2025mosca,wang2025shape}.
Another line of works models per-Gaussian motion explicitly.
Dynamic3DGS~\cite{luiten2024dynamic} pioneered this direction, storing each Gaussian's position and rotation at every frame.
Later methods make this motion more compact and expressive, either with Gaussians whose position and opacity vary over time~\cite{yang2024realtime,li2024spacetime,duan20244drotor,xu2024representing,wang2025freetimegs}, or by parameterizing each trajectory with keyframes, splines, or other temporal bases~\cite{katsumata2024compact,lee2024fully,park2025splinegs,tzathas2026adaptive}.

The methods above are trained for appearance alone, with no dedicated representation of the surface.
Dynamic surface reconstruction methods initially addressed this with neural implicit surfaces that deform or evolve a signed distance function over time~\cite{cai2022neural,shao2023tensor4d,wang2023neus2,mao2024neural,ebbed2026moangelo,yao2024dynosurf}.
More recently, Gaussian-based dynamic surface reconstruction methods have emerged, in three forms.
Hybrid NeRF--Gaussian methods~\cite{li2025dgns,kim2026dysurface} couple dynamic Gaussians with a volume-rendered signed distance function.
Hybrid mesh--Gaussian methods~\cite{ma2025mags,zheng2025gaustar,tao2026tessellation} bind Gaussians to an input mesh that they deform or track over time.
Lastly, Gaussian-only methods~\cite{liu2025dynamic,cai2024dynasurfgs,zhang2025dynamic,wang2024spacetime,chen2025adaptive,wu20264dsurf} represent the surface with oriented or flattened Gaussians and extract each frame's mesh by volume integration of rendered depth or by Poisson reconstruction.
Across all three groups, most methods target isolated objects with foreground masks and rely on bounded volumetric grids or input meshes to recover the surface, which keeps them from reconstructing complete dynamic scenes.

%% file: sec/3_background.tex
\section{Background}
\label{sec:background}
\input{figures/pipeline}

3D Gaussian splatting~\cite{kerbl20233d} represents a scene as a set of anisotropic 3D Gaussians,
\begin{equation}
  G_i(\mathbf{x}) = \exp\!\left(-\tfrac{1}{2}(\mathbf{x}-\boldsymbol{\mu}_i)^\top \boldsymbol{\Sigma}_i^{-1} (\mathbf{x}-\boldsymbol{\mu}_i)\right),
  \label{eq:gaussian}
\end{equation}
where $\boldsymbol{\mu}_i \in \mathbb{R}^3$ is the mean and $\boldsymbol{\Sigma}_i = \mathbf{R}_i \mathbf{S}_i \mathbf{S}_i^\top \mathbf{R}_i^\top$ is the covariance matrix, factored into a rotation matrix $\mathbf{R}_i$ given by a unit quaternion $\mathbf{q}_i$ and a diagonal scaling matrix $\mathbf{S}_i = \operatorname{diag}(\mathbf{s}_i)$.
Each Gaussian also carries an opacity $o_i \in [0,1]$ and spherical-harmonic coefficients that define its view-dependent color $\mathbf{c}_i$.
To render an image, each Gaussian is projected onto the image plane with the 2D covariance $\boldsymbol{\Sigma}'_i = \mathbf{J}\mathbf{W}\boldsymbol{\Sigma}_i\mathbf{W}^\top\mathbf{J}^\top$, where $\mathbf{W}$ is the world-to-camera transformation and $\mathbf{J}$ is the Jacobian matrix of the projection's local affine approximation~\cite{zwicker2001ewa}.
The projected Gaussians are sorted by depth, and the color of pixel $\mathbf{p}$ is obtained by front-to-back alpha compositing,
\begin{equation}
  \hspace{-0.56em}\mathbf{C}(\mathbf{p}) = \sum_i \mathbf{c}_i\alpha_i(\mathbf{p})T_i(\mathbf{p}), \quad T_i(\mathbf{p}) = \prod_{j<i}\bigl(1-\alpha_j(\mathbf{p})\bigr),
  \label{eq:compositing}
\end{equation}
where $\alpha_i(\mathbf{p}) = o_i\,G'_i(\mathbf{p})$ is the opacity of the projected Gaussian $G'_i$ at $\mathbf{p}$, and $T_i(\mathbf{p})$ is the transmittance accumulated by the Gaussians in front of it.
The parameters are optimized with the photometric loss
\begin{equation}
  \mathcal{L}_{\text{RGB}} = (1-\lambda)\,\bigl\|\mathbf{I}_{\text{r}} - \mathbf{I}_{\text{gt}}\bigr\|_1 + \lambda\,\bigl(1 - \operatorname{SSIM}(\mathbf{I}_{\text{r}}, \mathbf{I}_{\text{gt}})\bigr),
  \label{eq:rgb_loss}
\end{equation}
where $\mathbf{I}_{\text{r}}$ and $\mathbf{I}_{\text{gt}}$ are the rendered and ground-truth images, and $\lambda = 0.2$.
During optimization, adaptive densification clones and splits Gaussians in under-reconstructed regions and prunes nearly transparent ones.

%% file: figures/pipeline.tex
\begin{figure*}[t]
  \centering
  \definecolor{gradred}{RGB}{200, 40, 40}
  \newcommand{\fire}[1][1.8ex]{\raisebox{-0.25ex}{\includegraphics[height=#1]{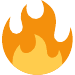}}}
  \begin{tikzpicture}[
      x=1cm, y=1cm,
      panel/.style={inner sep=0pt, outer sep=0pt},
      lbl/.style={font=\scriptsize, inner sep=1pt},
      title/.style={lbl, text height=1.6ex, text depth=0.4ex},
      flow/.style={-{Stealth[length=4pt]}, line width=0.6pt, draw=black!70},
      loss/.style={line width=0.5pt, draw=black!70},
      box/.style={draw=black!70, line width=0.6pt, rounded corners=3pt},
      grad/.style={-{Stealth[length=4pt]}, line width=0.6pt, draw=gradred,
                   dash pattern=on 2.5pt off 1.5pt},
    ]
    \def\dy{0.5}
    \def\ry{0.3}
    \node[panel] (can) at (1.45, 0.7+\dy)
      {\includegraphics[width=2.9cm]{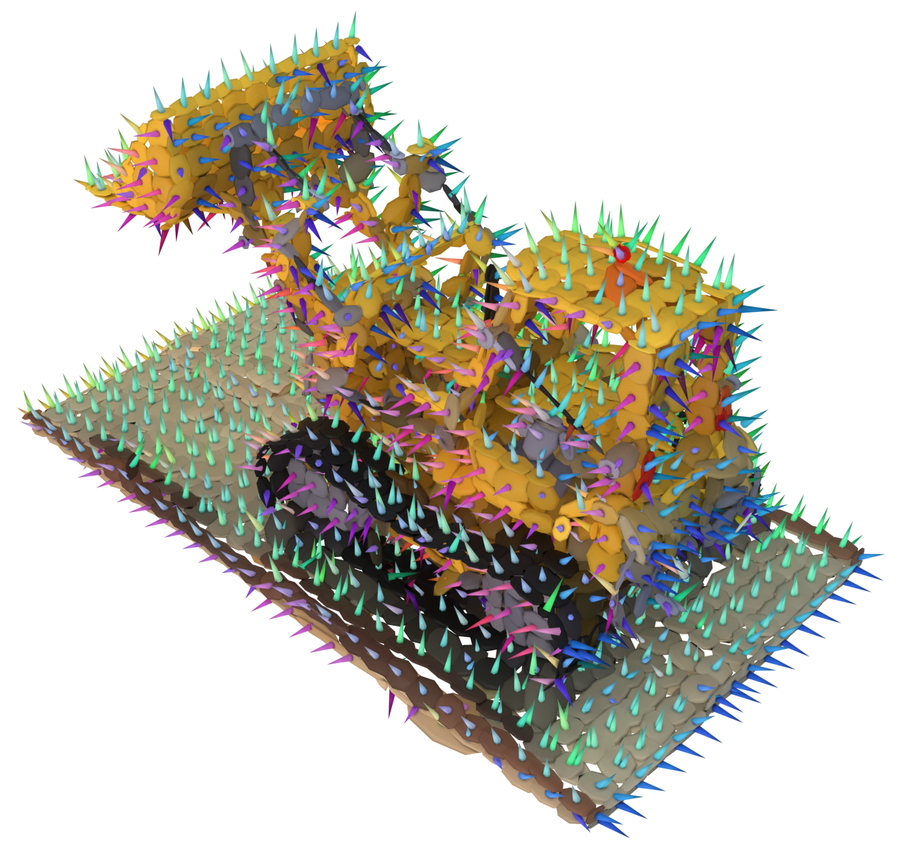}};
    \node[title, below=1pt of can] (canT) {Canonical Gaussians~\fire};
    \node[box, fill=black!4, minimum width=1.5cm, minimum height=1.1cm,
          align=center, font=\scriptsize] (motion) at (4.1, 0.7+\dy)
      {Motion\\model\\(\cref{subsec:motion_model})};
    \node[lbl] (t) at (3.7, -0.4+\dy) {$t$};
    \draw[flow] (t) -- (t |- motion.south);
    \node[panel] (gt1) at (6.75, 0.7+\dy)
      {\includegraphics[width=2.9cm]{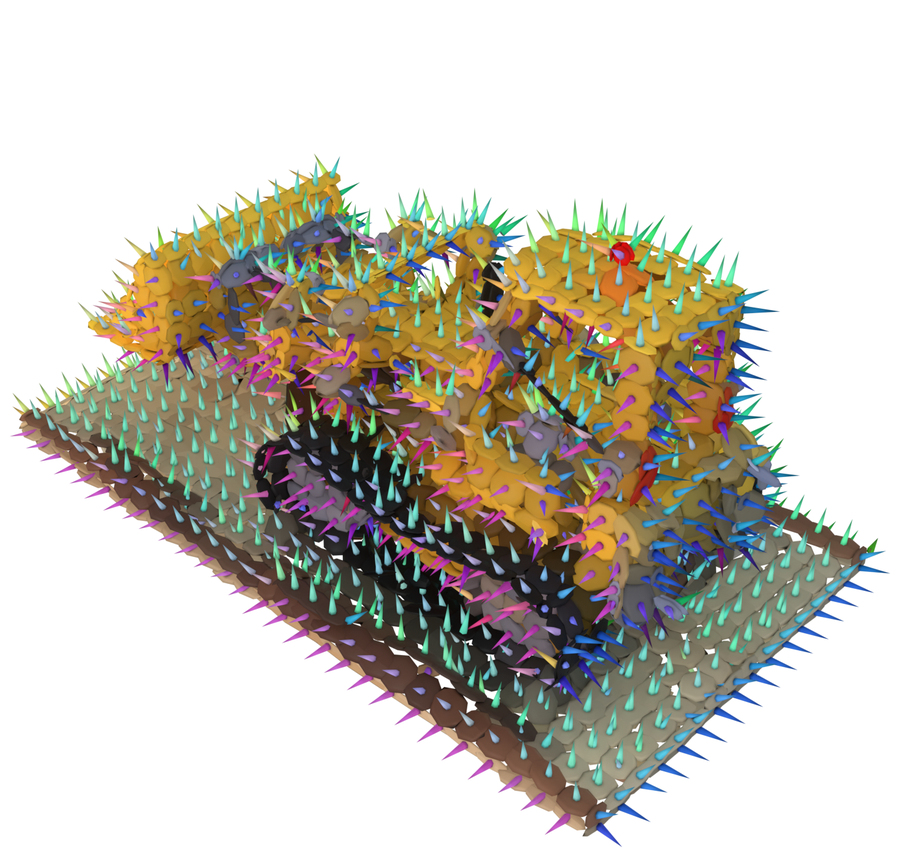}};
    \node[title, below=1pt of gt1] {Gaussians at time $t$};
    \draw[flow] (can.east) -- (motion.west);
    \draw[flow] (motion.east) -- (gt1.west);
    \node[box, fill=black!4, align=center, font=\scriptsize] (init) at (1.45, -2.05)
      {Initialization (\cref{subsec:initialization})};
    \draw[flow] (init.north) -- (canT.south);
    \draw[flow] (init.east) -| ($(motion.south)+(0.4, 0)$);

    \def\w{1.57cm}
    \node[panel] (render) at (9.55, 1.35+\ry) {\includegraphics[width=\w]{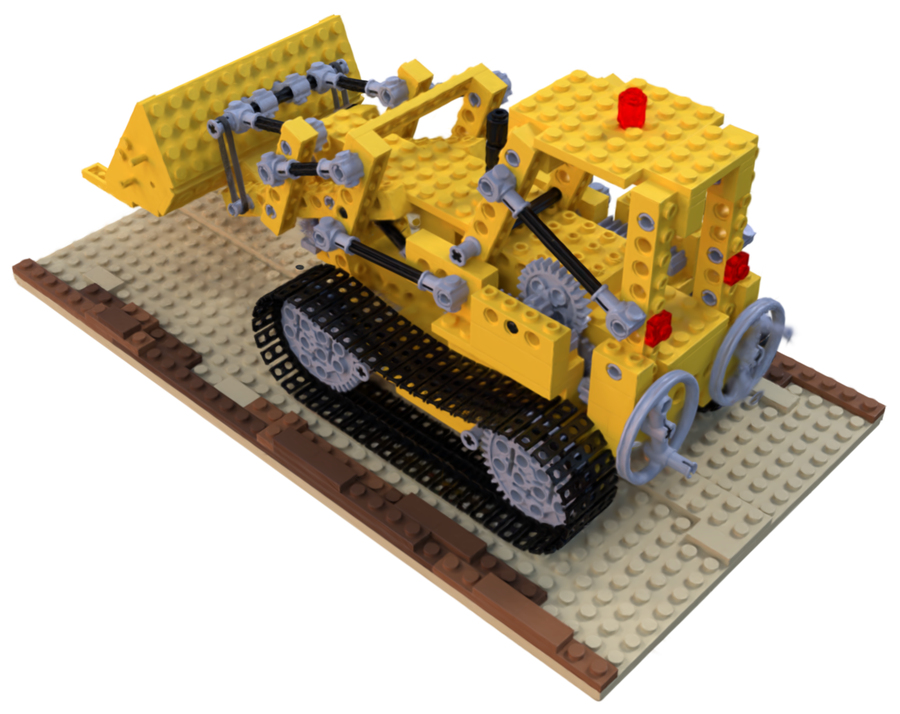}};
    \node[panel, right=0.35cm of render] (gt) {\includegraphics[width=\w]{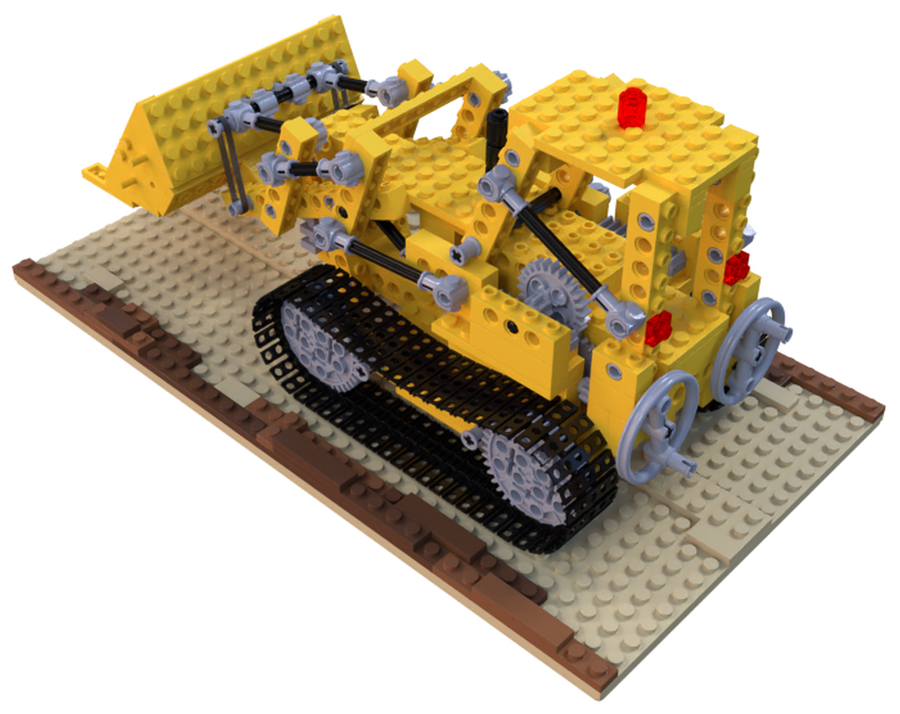}};
    \node[panel, right=0.07cm of gt] (ng) {\includegraphics[width=\w]{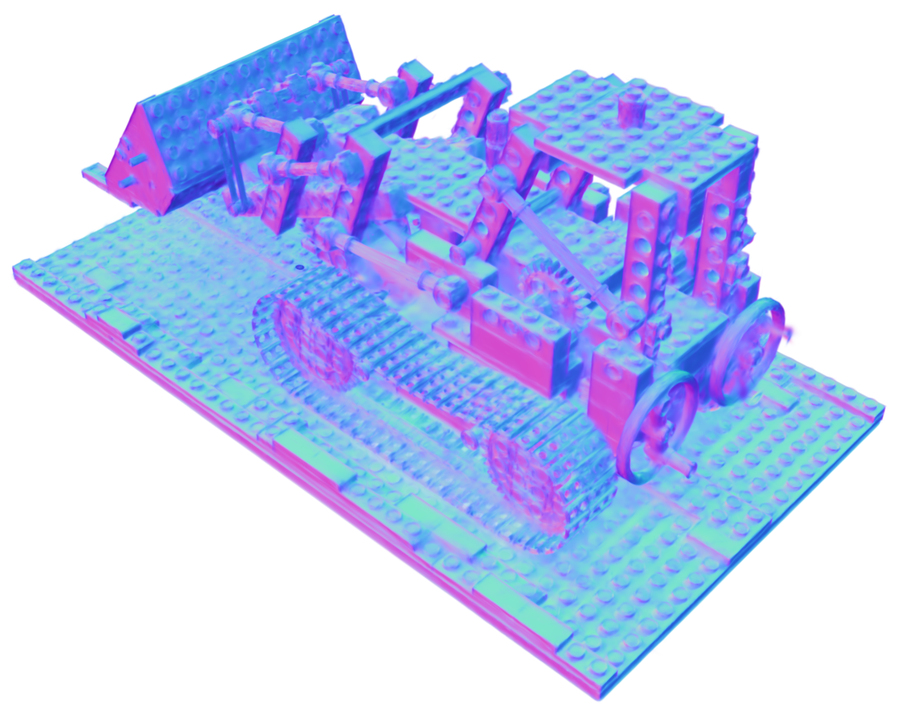}};
    \node[panel, right=0.07cm of ng] (nd) {\includegraphics[width=\w]{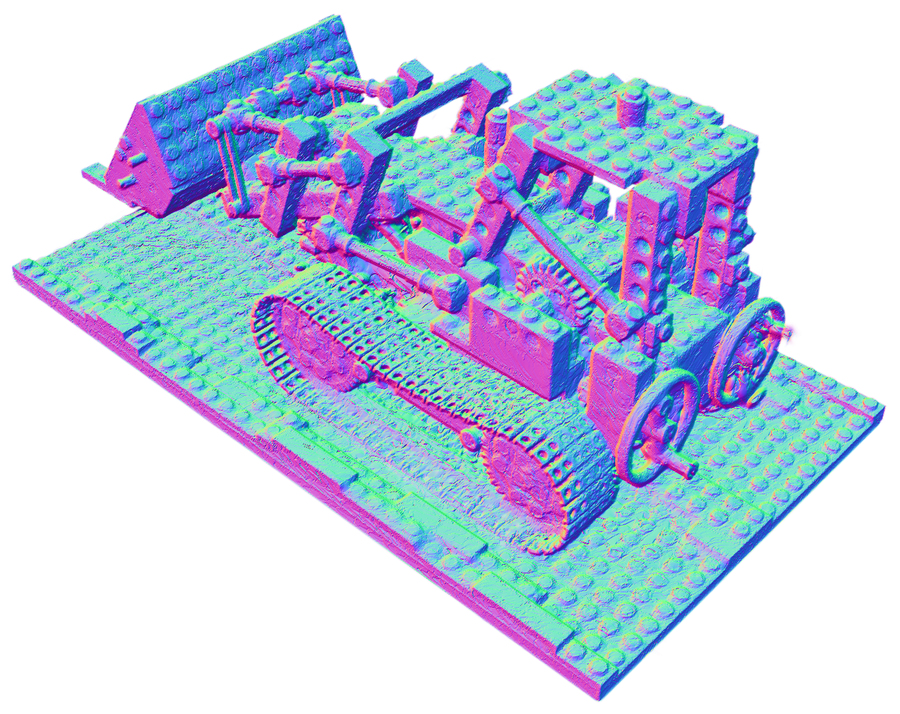}};
    \node[panel, right=0.07cm of nd] (no) {\includegraphics[width=\w]{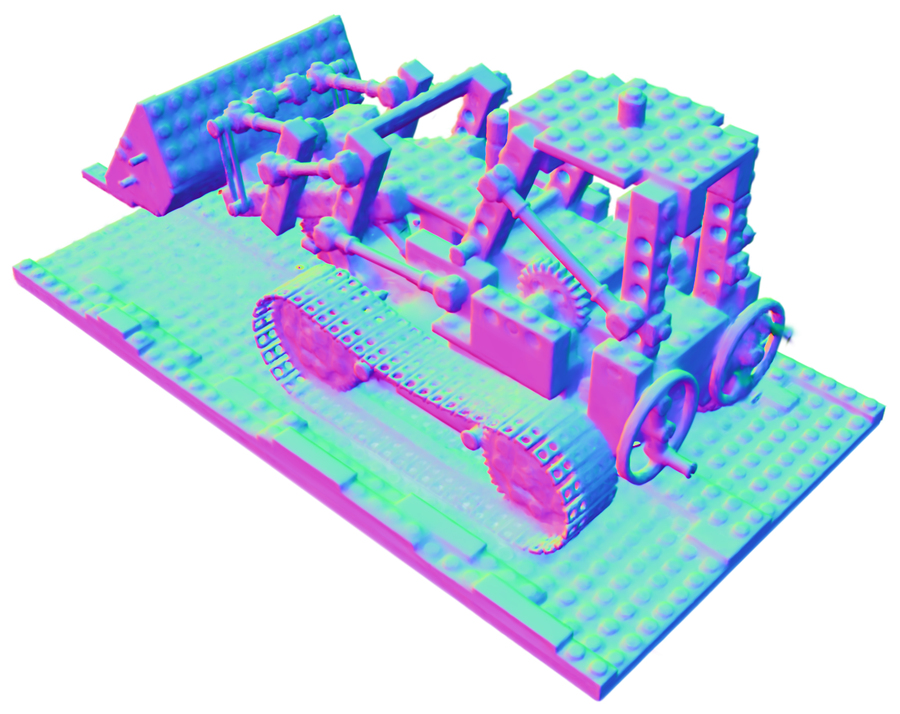}};
    \foreach \n/\t in {render/Render, gt/Ground truth, ng/$\mathbf{N}_g$,
                       nd/$\mathbf{N}_d$, no/$\mathbf{N}_o$}
      \node[title, below=0pt of \n] {\t};
    \coordinate (lb) at ($(render.north)+(0, 0.08)$);
    \foreach \a/\b/\l/\n in {render/gt/{\mathcal{L}_{\text{RGB}}}/Lrgb,
                             ng/nd/{\mathcal{L}_D}/LD, nd/no/{\mathcal{L}_N}/LN} {
      \draw[loss] ($(\a.north |- lb)+(0.05, 0)$) -- ++(0, 0.12)
        -| ($(\b.north |- lb)+(-0.05, 0)$)
        node[pos=0.25, above=1.5pt, lbl] (\n) {$\l$};
    }
    \def\off{0.16}
    \draw[grad] ($(render.west)+(0, -\off)$) -- (8.38+\off, 1.35+\ry-\off)
      -- (8.38+\off, 0.7+\dy-\off) -- ($(gt1.east)+(0, -\off)$);
    \draw[grad] ($(gt1.west)+(0, -\off)$) -- ($(motion.east)+(0, -\off)$);
    \draw[grad] ($(motion.west)+(0, -\off)$) -- ($(can.east)+(0, -\off)$);
    \node[inner sep=0pt] at (motion.north east) {\fire[0.32cm]};
    \draw[box, line width=1pt, dash pattern=on 5pt off 3pt]
      (-0.1, -1.6) -- (8.6, -1.6) -- (8.6, 0.1) -- (17.33, 0.1)
      -- (17.33, 3.5) -- (-0.1, 3.5) -- cycle;
    \node[lbl, anchor=north east, font=\scriptsize\itshape] at (17.28, 3.45)
      {Training (\cref{subsec:training})};
    \node[lbl, anchor=north west] (key) at (0.0, 3.45)
      {\fire~Learned\quad
       \tikz[baseline=-0.6ex]\draw[grad] (0, 0) -- (0.6, 0);~Gradients};

    \node[panel, anchor=north west] (tets) at (9.2, -0.1)
      {\includegraphics[width=2.4cm]{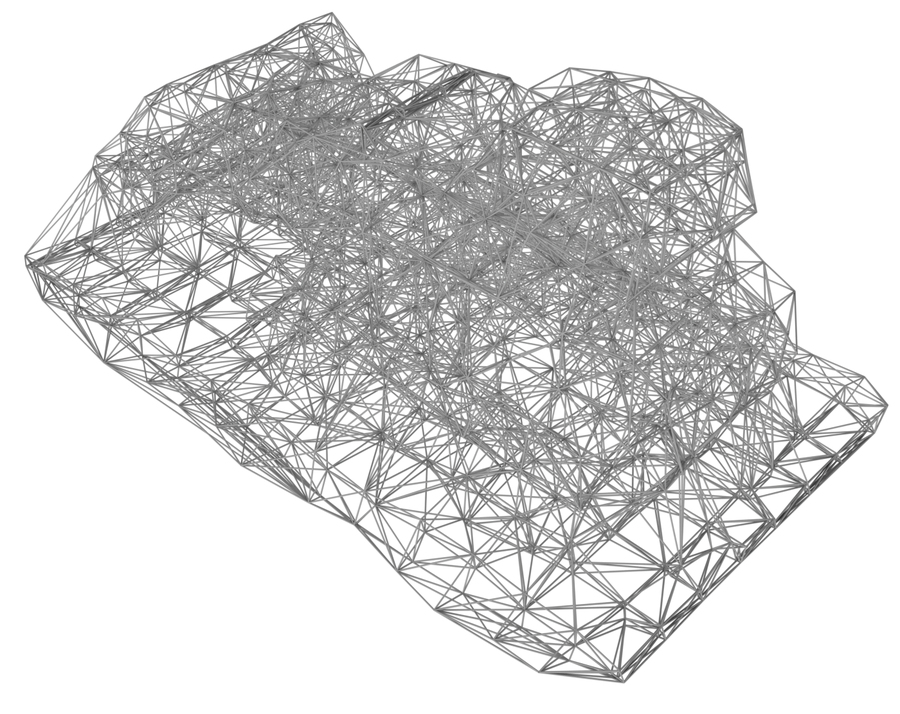}};
    \node[panel, right=0.45cm of tets] (vacancy)
      {\includegraphics[width=2.4cm]{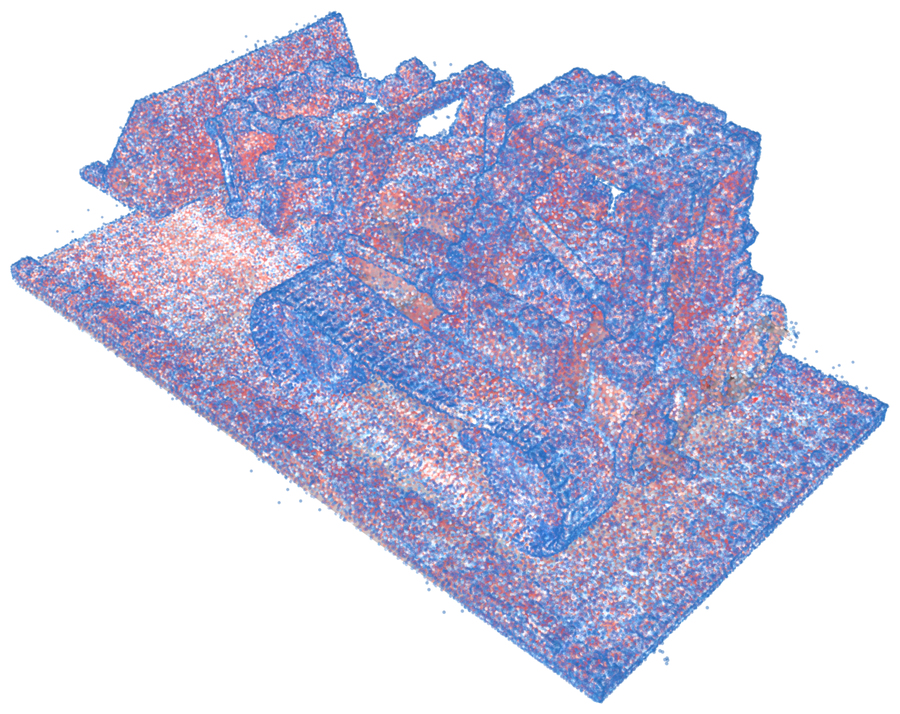}};
    \node[panel, right=0.45cm of vacancy] (mesh)
      {\includegraphics[width=2.4cm]{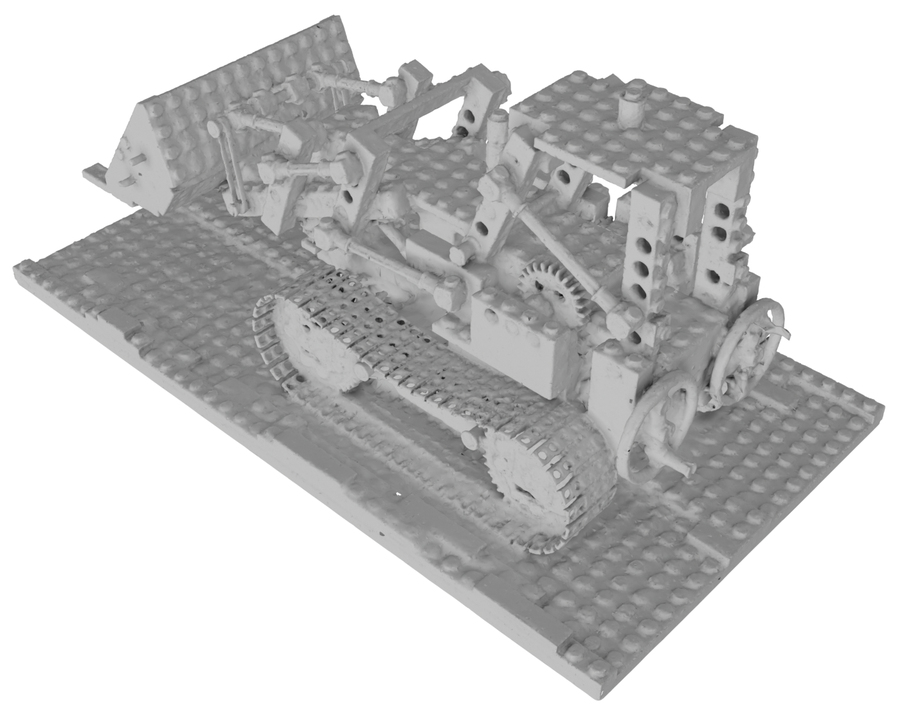}};
    \draw[flow] (tets.east) -- (vacancy.west);
    \draw[flow] (vacancy.east) -- (mesh.west);
    \path let \p1 = (tets.south), \p2 = (vacancy.south), \p3 = (mesh.south) in
      coordinate (rowbottom) at (0, {min(\y1, \y2, \y3)});
    \node[title, anchor=north] at (tets.south |- rowbottom) {Tetrahedralization};
    \node[title, anchor=north] at (vacancy.south |- rowbottom) {Vacancy field};
    \node[title, anchor=north] at (mesh.south |- rowbottom) {Mesh};

    \coordinate (fork) at (8.38, 0.7+\dy);
    \draw[line width=0.6pt, draw=black!70] (gt1.east) -- (fork);
    \draw[flow] (fork) |- (render.west);
    \draw[flow] (fork) |- (tets.west);
  \end{tikzpicture}
  \caption{Overview of \ourmethod.
    The motion model translates and rotates each Gaussian and its oriented normal (arrows) from the canonical state to time~$t$.
    Training backpropagates (red) $\mathcal{L}_{\text{RGB}}$, $\mathcal{L}_D$ between the geometric and depth normals $\mathbf{N}_g$ and $\mathbf{N}_d$, and $\mathcal{L}_N$ between $\mathbf{N}_d$ and the oriented normals $\mathbf{N}_o$.
    Mesh extraction follows Gaussian Wrapping~\cite{gomez2026blobs}: tetrahedralizing the pivots, evaluating the vacancy field at each pivot (red occupied, blue vacant), and recovering its 0.5 level set with marching tetrahedra.}
  \label{fig:pipeline}
  \vspace{-8pt}
\end{figure*}

%% file: sec/4_method.tex
\section{Method}
\label{sec:method}

Given multi-view videos of a dynamic scene captured by synchronized and calibrated cameras, \ourmethod reconstructs persistent oriented Gaussians whose geometry is defined at every time $t \in [0,1]$.
Geometry is represented with the oriented normals proposed in Gaussian Wrapping~\cite{gomez2026blobs}, and our motion model computes each Gaussian's position, rotation, and normal at any time $t$ (\cref{subsec:motion_model}).
Before training, we initialize each Gaussian's trajectory by fitting its motion model to point tracks spanning the sequence (\cref{subsec:initialization}).
Training then optimizes all parameters with photometric and geometric losses (\cref{subsec:training}).
After training, a mesh can be extracted directly at any frame with marching tetrahedra on the Gaussians evaluated at that time.
An overview of our full pipeline is shown in \cref{fig:pipeline}.

\subsection{Motion Model}
\label{subsec:motion_model}

\ourmethod attaches a temporal model to every Gaussian, encoding motion as displacements from a base position and rotation at each query time $t$.
Assuming locally rigid motion, the scale of each Gaussian stays constant.
The displacements are modeled as a linear combination of temporal Gaussian kernels, where each kernel is centered around a different inducing time.
Since the kernels cover the whole timeline, the model holds at every frame, so opacity can remain constant as well.
This ensures that every Gaussian remains active at every frame, whereas prior works vary opacity over time~\cite{yang2024realtime,li2024spacetime,duan20244drotor,xu2024representing,wang2025freetimegs} because their motion models hold only within short windows.

Our model is inspired by Gaussian processes (GPs)~\cite{rasmussen2006gaussian}, whose posterior mean is also a weighted sum of kernels.
While a GP solves for its weights in closed form with fixed inducing points and kernel lengthscale, we relax both and learn them jointly with the weights during training.
Unlike deformation fields~\cite{wu20244d,yang2024deformable} and control points~\cite{huang2024scgs}, which share motion among neighboring Gaussians, and polynomial trajectories~\cite{li2024spacetime,wang2025freetimegs}, which hold only within short windows, our model gives each Gaussian its own motion over the whole sequence.
Concretely, each Gaussian $G_i$ carries two sets of $M_{\text{pos}}$ and $M_{\text{rot}}$ temporal Gaussian kernels, for position displacements and changes in rotation (represented as quaternions), respectively.
Each kernel takes the form
\begin{equation}
  k_{i,m}(t) = \exp\!\left(-\frac{(t-\tau_{i,m})^2}{2\ell_i^2}\right),
  \label{eq:kernel}
\end{equation}
where $\tau_{i,m}$ is the kernel's learnable inducing time (its center) and $\ell_i$ is a learnable lengthscale, one per set, that controls its width.
A set's inducing times are confined through a sigmoid to disjoint bins of width $1/M_{\text{pos}}$ or $1/M_{\text{rot}}$ across the unit interval, preventing clustering and ensuring full coverage of the timeline.
Similarly, the lengthscale is a sigmoid of a learned parameter, initialized to the bin width of its set so that neighboring kernels overlap.

Given query time $t$, the position displacement and rotation delta of $G_i$ are computed from weighted sums of its position and rotation kernels, $k^{\text{pos}}_{i,m}$ and $k^{\text{rot}}_{i,m}$,
\begin{equation}
  \begin{aligned}
    &\Delta\boldsymbol{\mu}_i(t) = \sum_{m=1}^{K_{\text{pos}}} k^{\text{pos}}_{i,m}(t)\, \mathbf{w}^{\text{pos}}_{i,m}, \\
    &\Delta\mathbf{q}_i(t) = \operatorname{normalize}\Bigl(\mathbf{q}_{\text{id}} + \sum_{m=1}^{K_{\text{rot}}} k^{\text{rot}}_{i,m}(t)\, \mathbf{w}^{\text{rot}}_{i,m}\Bigr),
  \end{aligned}
  \label{eq:displacements}
\end{equation}
where $\mathbf{w}^{\text{pos}}_{i,m} \in \mathbb{R}^3$ and $\mathbf{w}^{\text{rot}}_{i,m} \in \mathbb{R}^4$ are learnable weights, $\mathbf{q}_{\text{id}}$ is the identity quaternion, and the normalization to unit length makes $\Delta\mathbf{q}_i(t)$ a valid rotation, which equals the identity when the weights are zero.
\Cref{fig:motion_model} illustrates the model.
The final position at time $t$ is $\boldsymbol{\mu}_i(t) = \boldsymbol{\mu}_i + \Delta\boldsymbol{\mu}_i(t)$.
For rotation, the time-dependent quaternion is
\begin{equation}
  \mathbf{q}_i(t) = \Delta\mathbf{q}_i(t) \otimes \mathbf{q}_i,
  \label{eq:rotation_motion}
\end{equation}
where $\otimes$ denotes quaternion multiplication.

With inducing times only inside the sequence, the first and last frames are covered by kernels on one side only, so their motion is extrapolated rather than interpolated and their reconstruction quality drops (\suppref{fig:ablations_boundary}).
We therefore add $B = 2$ inducing times beyond each end of the sequence, which restores two-sided kernel support at the boundary frames and gives $K_{\text{pos}} = M_{\text{pos}} + 2B$ and $K_{\text{rot}} = M_{\text{rot}} + 2B$ kernels in total.

To represent the scene's geometry, we follow Gaussian Wrapping~\cite{gomez2026blobs} and give each Gaussian $G_i$ a learned normal $\mathbf{n}_i$ in its base state, alongside its base position $\boldsymbol{\mu}_i$ and rotation $\mathbf{q}_i$.
The normal's orientation is learned as well and determines which side of the surface is inside and which is outside.
Since the rotation delta is applied in the world frame, the normal at time $t$ is rotated by the same delta,
\begin{equation}
  \mathbf{n}_i(t) = \mathbf{R}\bigl(\Delta\mathbf{q}_i(t)\bigr)\, \mathbf{n}_i,
  \label{eq:normal_motion}
\end{equation}
where $\mathbf{R}(\cdot)$ converts a quaternion to a rotation matrix.

\input{figures/motion_model}

Oriented normals are an ideal fit for our motion model, since they move with the rotation delta it already computes, with no modification or additional parameters, and thus keep the scene's implicit geometry valid at every frame.
They also keep training fast, since they do not require the repeated costly mesh extractions of MILo~\cite{guedon2025milo} and DG-Mesh~\cite{liu2025dynamic}, and they define a vacancy field~\cite{miller2024objects,gomez2026blobs} that serves as the implicit geometry for direct extraction.

\paragraph{Mesh extraction.}
At inference, we evaluate the motion model at frame $t$ to get the state of the Gaussians, and extract the frame's mesh following the pivot-based method of Gaussian Wrapping.
We spawn two pivots from each Gaussian and compute the Delaunay tetrahedralization of all pivots.
We evaluate the vacancy field at the pivots and extract its 0.5 isosurface with marching tetrahedra.

\input{tables/main_results}

\subsection{Initialization}
\label{subsec:initialization}

Dynamic scenes often contain large and fast motion, such as swinging limbs or spinning parts.
Such motion is challenging for any motion model, since photometric gradients only guide a Gaussian toward nearby image content.
One way to mitigate this challenge is to start from a good initial approximation of the motion.
Most prior methods, however, forgo this approach and start from a static model~\cite{wu20244d,yang2024deformable,liu2025dynamic} or from zero motion~\cite{tzathas2026adaptive,huang2024scgs,zhang2025dynamic}.
Others approximate the motion only between consecutive frames~\cite{luiten2024dynamic,wang2025freetimegs}, where large displacements can divert trajectories.
Our initialization scheme instead finds for each Gaussian $G_i$ a 3D trajectory spanning the entire sequence, using a closed-form fit to tracks built from per-frame SfM points, so that large and fast motion is captured before training begins.

Our initialization runs once before training and takes as input a sparse COLMAP~\cite{schonberger2016structure} point cloud at every frame, as in~\cite{li2024spacetime,wang2025freetimegs}.
First, we link these points into 3D tracks over the entire sequence.
We begin by selecting one or more anchor frames according to the scene.
A single anchor at the first or middle frame suffices when the moving content is visible there, whereas scenes whose content is revealed over time use several anchors spread across the sequence.
Since every anchor starts a new track for each of its points, including static ones, scenes dominated by a static background use few anchors.
From each anchor, a Kalman filter~\cite{kalman1960new} then runs forward and backward through the sequence, associating each point with the nearest point to its predicted position.
Implausible matches are rejected with a Mahalanobis gate, in which case the track follows the motion of its nearest accepted neighbors instead of being diverted.
The tracks of all anchors are united.

Second, we fit the position weights of each Gaussian to its track in closed form by GP regression~\cite{rasmussen2006gaussian}, with the inducing times and lengthscales fixed at the bin centers and the bin width, respectively, and set its rotation deltas to the identity quaternion.
Finally, if the tracks yield fewer than 50K Gaussians, we densify the initialization by sampling additional Gaussians uniformly within a shell around the tracked points.
Each spawned Gaussian inherits the motion weights of its nearest tracked Gaussian.
\Cref{fig:initialization} shows the Kalman tracks and the trajectories fitted to them.

\input{figures/initialization}

\input{figures/geometry}

\subsection{Training}
\label{subsec:training}

We optimize all parameters by rendering each training view at its capture time and minimizing a photometric loss together with geometric losses on depth and normals.
The total loss is
\begin{equation}
  \mathcal{L} = \mathcal{L}_{\text{RGB}} + \lambda_{D}\mathcal{L}_{D} + \lambda_N\mathcal{L}_N,
  \label{eq:total_loss}
\end{equation}
where $\mathcal{L}_{\text{RGB}}$ is the photometric loss of \cref{eq:rgb_loss}, $\mathcal{L}_{D}$ is the depth-normal consistency loss of RaDe-GS~\cite{zhang2026radegs}, and $\mathcal{L}_N$ is the normal alignment loss of Gaussian Wrapping~\cite{gomez2026blobs}.

$\mathcal{L}_D$ makes the rendered depth consistent with the shape of the Gaussians.
It compares the geometric unit normals $\mathbf{N}_g$, rendered from the shape of each Gaussian~\cite{zhang2026radegs}, with $\mathbf{N}_d$, the unit surface normal computed from the rendered depth map, which blends the expected and median depths,
\begin{equation}
  \mathcal{L}_D = \sum_{p} \alpha(p)\bigl(1 - \mathbf{N}_g(p)^\top \mathbf{N}_d(p)\bigr),
  \label{eq:d_loss}
\end{equation}
where $\alpha(p)$ is the rendered alpha at pixel $p$.

$\mathcal{L}_N$ aligns the oriented normals of the Gaussians with the surface.
It compares the alpha-blended oriented normals $\mathbf{N}_o$ with the same depth normal $\mathbf{N}_d$,
\begin{equation}
  \mathcal{L}_N = \sum_{p} \bigl(\alpha(p) - \mathbf{N}_o(p)^\top \mathbf{N}_d(p)\bigr).
  \label{eq:n_loss}
\end{equation}

Accurate meshes also require the Gaussians to densely cover fine geometry, which static Gaussian Wrapping~\cite{gomez2026blobs} obtains from the multi-view loss of GGGS~\cite{zhang2026geometry}.
We instead adopt the importance-based densification and simplification of Mini-Splatting2~\cite{fang2026efficient}, lifted to 4D (\suppref{subsec:impl_lifting}), which trains faster and avoids holes in textureless regions (\suppref{fig:ablations}).

\input{figures/nvs}

%% file: figures/motion_model.tex
\begin{figure}[t]
  \centering
  \includegraphics[width=\linewidth]{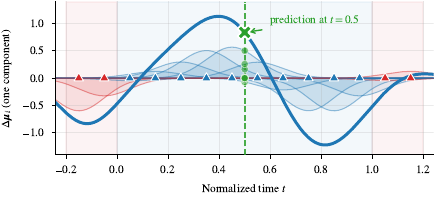}
  \caption{Motion model for one coordinate of the position displacement $\Delta\boldsymbol{\mu}_i(t)$ of a Gaussian $G_i$.
    At a query time $t$, the prediction is the sum of the $M_{\text{pos}} + 2B$ weighted interior (blue) and boundary (red) kernels.}
  \label{fig:motion_model}
  \vspace{-8pt}
\end{figure}

%% file: tables/main_results.tex
\begin{table*}[t]
  \centering
  \begin{subtable}{\linewidth}
    \centering
    \footnotesize
    \setlength{\tabcolsep}{3.5pt}
    \begin{tabular}{@{}lccccccccccccc}
      \toprule
      & \multicolumn{7}{c}{\benchmark} & \multicolumn{3}{c}{Neural 3D Video} & \multicolumn{3}{c}{SelfCap} \\
      \cmidrule(lr){2-8} \cmidrule(lr){9-11} \cmidrule(lr){12-14}
      Method & PSNR$\uparrow$ & SSIM$\uparrow$ & LPIPS$\downarrow$ & CD$\downarrow$ & NC$\downarrow$ & F$_\text{S}\uparrow$ & F$_\text{R}\uparrow$ & PSNR$\uparrow$ & SSIM$\uparrow$ & LPIPS$\downarrow$ & PSNR$\uparrow$ & SSIM$\uparrow$ & LPIPS$\downarrow$ \\
      \midrule
      4DGS~\cite{wu20244d} & 33.25 & 0.971 & 0.043 & 1.14 & 38.0 & 0.283 & 0.359 & \gold{31.29} & \bronze{0.939} & 0.222 & \silver{26.78} & 0.881 & 0.255 \\
      Deformable-3DGS~\cite{yang2024deformable} & 32.59 & 0.976 & \bronze{0.034} & 1.24 & 37.9 & 0.285 & 0.365 & 30.28 & 0.937 & 0.220 & 26.51 & \bronze{0.897} & \bronze{0.212} \\
      SC-GS~\cite{huang2024scgs} & 32.91 & \bronze{0.977} & \silver{0.033} & 1.16 & 36.4 & 0.321 & 0.413 & 26.86 & 0.914 & 0.249 & 24.36 & 0.876 & 0.247 \\
      FreeTimeGS~\cite{wang2025freetimegs} & \silver{33.51} & 0.976 & 0.036 & 1.61 & 45.1 & 0.161 & 0.238 & 29.93 & 0.932 & \gold{0.209} & \bronze{26.77} & \silver{0.905} & \gold{0.195} \\
      AST-3DGS~\cite{tzathas2026adaptive} & 29.37 & 0.965 & 0.050 & 1.87 & 40.1 & 0.256 & 0.324 & \silver{30.80} & \silver{0.940} & \bronze{0.214} & 26.26 & 0.886 & 0.244 \\
      D-2DGS~\cite{zhang2025dynamic} & \bronze{33.38} & \silver{0.978} & \silver{0.033} & \gold{0.46} & \bronze{34.6} & \bronze{0.365} & \silver{0.541} & 25.99 & 0.900 & 0.283 & 21.96 & 0.822 & 0.331 \\
      DG-Mesh~\cite{liu2025dynamic} & 24.90 & 0.945 & 0.066 & \bronze{0.81} & \silver{32.3} & \silver{0.410} & \bronze{0.477} & -- & -- & -- & -- & -- & -- \\
      \midrule
      \ourmethod & \gold{34.96} & \gold{0.982} & \gold{0.025} & \silver{0.51} & \gold{24.6} & \gold{0.556} & \gold{0.746} & \bronze{30.74} & \gold{0.941} & \silver{0.212} & \gold{27.14} & \gold{0.913} & \silver{0.201} \\
      \ourmethod{} w/ TSDF & & & & 0.50 & 29.1 & 0.468 & 0.663 & & & & & & \\
      \ourmethod{} w/ Poisson & & & & 1.03 & 31.1 & 0.393 & 0.426 & & & & & & \\
      \bottomrule
    \end{tabular}
    \caption{Novel-view synthesis and geometry. CD is the Chamfer distance ($\times 100$), NC the normal Chamfer distance in degrees, and F$_\text{S}$ and F$_\text{R}$ the joint F-scores on surface samples and on normals rendered at the test views. \ourmethod{} w/ TSDF and w/ Poisson mesh the same models by TSDF fusion and by the Poisson reconstruction of DG-Mesh, and are not ranked.}
    \label{tab:main_quality}
  \end{subtable}

  \vspace{0.6em}

  \begin{subtable}{\linewidth}
    \centering
    \footnotesize
    \setlength{\tabcolsep}{3.5pt}
    \begin{tabular*}{\linewidth}{@{\extracolsep{\fill}}lcccccccc@{}}
      \toprule
      Dataset & 4DGS & Deformable-3DGS & SC-GS & FreeTimeGS & AST-3DGS & D-2DGS & DG-Mesh & \ourmethod \\
      \midrule
      \benchmark & \silver{16m} & 25m & 26m & \gold{15m} & 31m & 33m & 142m & \bronze{18m} \\
      Neural 3D Video & \bronze{48m} & 52m & 60m & \gold{32m} & 69m & 90m & -- & \silver{38m} \\
      SelfCap & 79m & 129m & 138m & \gold{36m} & \bronze{78m} & 149m & -- & \silver{49m} \\
      \bottomrule
    \end{tabular*}
    \caption{Training time.}
    \label{tab:main_time}
  \end{subtable}
  \caption{Comparison on the three datasets. \swatch{rankgold}\,=\,best, \swatch{ranksilver}\,=\,second best, \swatch{rankbronze}\,=\,third best.}
  \label{tab:main_results}
  \vspace{-8pt}
\end{table*}

%% file: figures/initialization.tex
\begin{figure}[t]
  \centering
  \includegraphics[width=0.75\linewidth]{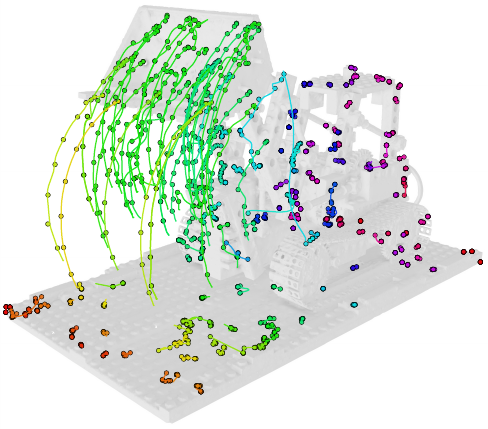}
  \caption{200 sampled tracks of Lego from our initialization with a single first-frame anchor, over the first-frame mesh.
    The fitted trajectories follow the bucket through its full sweep before training begins.
    Dots are track positions every tenth frame, and curves of the same color are the fitted trajectories.}
  \label{fig:initialization}
  \vspace{-8pt}
\end{figure}

%% file: figures/geometry.tex
\begin{figure*}[t]
  \centering
  \newlength{\geomw}\newlength{\geomgap}\newlength{\geompw}\newlength{\geomrw}
  \setlength{\geomw}{\dimexpr\linewidth-0.2in\relax}
  \setlength{\geomgap}{0.03in}
  \setlength{\geompw}{\dimexpr(\geomw-4\geomgap)/5\relax}
  \setlength{\geomrw}{\dimexpr(\geomw-2\geomgap)/3\relax}
  \newcommand{\geomhead}[5]{\makebox[\linewidth][l]{\makebox[0.2in]{}\makebox[\geompw]{#1}\hspace{\geomgap}\makebox[\geompw]{#2}\hspace{\geomgap}\makebox[\geompw]{#3}\hspace{\geomgap}\makebox[\geompw]{#4}\hspace{\geomgap}\makebox[\geompw]{#5}}}
  \newcommand{\geomreal}[3]{\makebox[\geomrw]{#1}\hspace{\geomgap}\makebox[\geomrw]{#2}\hspace{\geomgap}\makebox[\geomrw]{#3}}
  \newcommand{\geomrow}[2]{\makebox[0.2in]{\rotatebox[origin=c]{90}{#1}}\raisebox{-0.5\height}{\includegraphics[width=\geomw]{images/geometry/#2}}}
  {\footnotesize
  \geomhead{Deformable-3DGS}{DG-Mesh}{D-2DGS}{\ourmethod}{Ground truth}\\[1pt]
  \geomrow{Lego}{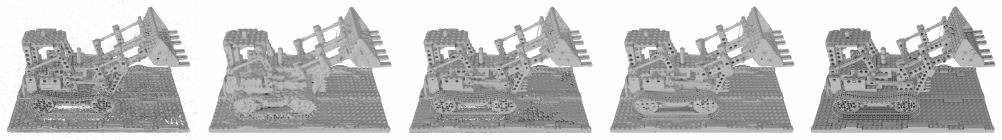}\\[1pt]
  \geomrow{Elephant}{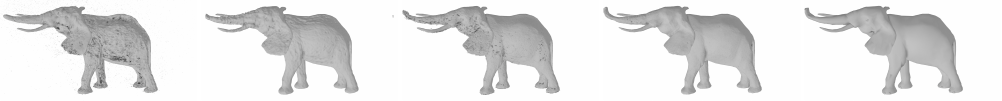}\\[1pt]
  \geomrow{Toyota}{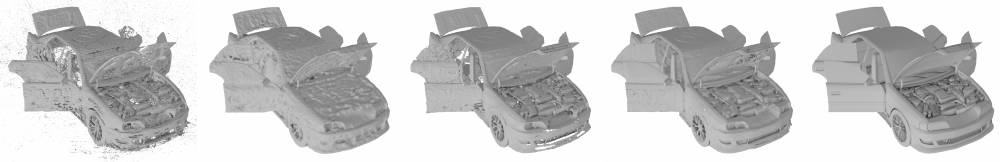}\\[1pt]
  \geomrow{Flame salmon}{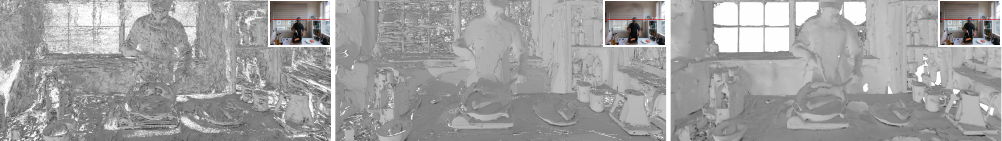}\\[1pt]
  \geomrow{Yoga}{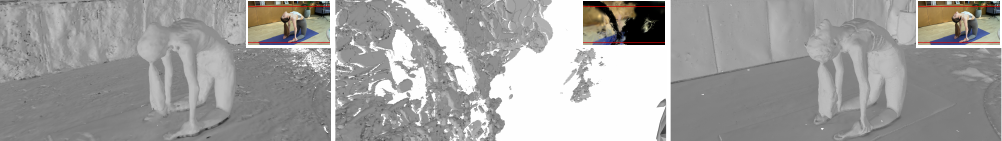}\\[1pt]
  \makebox[\linewidth][l]{\makebox[0.2in]{}\geomreal{Deformable-3DGS}{D-2DGS}{\ourmethod}}\par}
  \caption{Meshes of \ourmethod and the baselines, rendered from novel viewpoints.
    The top rows show three \benchmark scenes beside the ground truth, and the bottom rows show the real scenes Flame salmon and Yoga, with each method's color render inset and the crop outlined in red.
    Deformable-3DGS represents the appearance-only methods, as its meshes look the best among them, and the gray dots are Gaussian floaters.
    \ourmethod produces clean surfaces and preserves fine details such as facial features and Lego studs, where D-2DGS leaves noise and holes and DG-Mesh oversmooths.
    D-2DGS fails on Yoga, as its render shows.}
  \label{fig:geometry}
  \vspace{\ifx\papermode\papermodereview -8pt\else -14pt\fi}
\end{figure*}

%% file: figures/nvs.tex
\begin{figure*}[t]
  \centering
  \newlength{\nvsw}\newlength{\nvsgap}\newlength{\nvspw}
  \setlength{\nvsw}{\dimexpr\linewidth-0.2in\relax}
  \setlength{\nvsgap}{0.03in}
  \setlength{\nvspw}{\dimexpr(\nvsw-4\nvsgap)/5\relax}
  \newcommand{\nvscol}[1]{\makebox[\nvspw]{#1}}
  \newcommand{\nvsrow}[2]{\makebox[0.2in]{\rotatebox[origin=c]{90}{#1}}\raisebox{-0.5\height}{\includegraphics[width=\nvsw]{images/nvs/#2}}}
  {\footnotesize
  \makebox[\linewidth][l]{\makebox[0.2in]{}\nvscol{AST-3DGS}\hspace{\nvsgap}\nvscol{4DGS}\hspace{\nvsgap}\nvscol{FreeTimeGS}\hspace{\nvsgap}\nvscol{\ourmethod}\hspace{\nvsgap}\nvscol{Ground truth}}\\[1pt]
  \nvsrow{Spiderman}{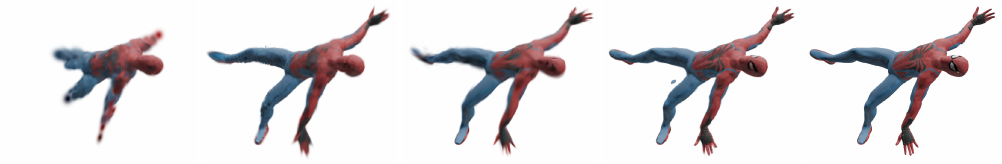}\\[1pt]
  \nvsrow{Flame steak}{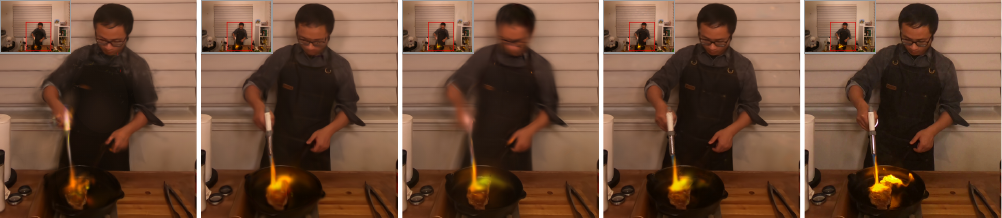}\\[1pt]
  \nvsrow{Hair2}{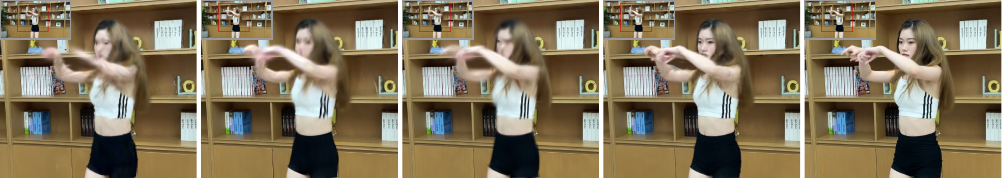}\par}
  \caption{Novel views of fast motion on \benchmark, Neural 3D Video, and SelfCap.
    On real scenes, panels crop the moving part, with the full frame inset and the crop outlined in red.
    \ourmethod keeps the limbs, hands, and face sharp, where the baselines blur or break them.}
  \label{fig:nvs}
  \vspace{-8pt}
\end{figure*}

%% file: sec/5_experiments.tex
\section{Experiments}
\label{sec:experiments}

\subsection{Experimental Setup}
\label{subsec:setup}

\paragraph{Datasets.}
\label{par:datasets}
Evaluating dynamic surface reconstruction requires ground-truth geometry.
To that end, we introduce \benchmark, a benchmark of nine animated scenes rendered in Blender (\suppref{fig:benchmark}), from articulated machinery to deforming animals and characters, each 100 to 170 frames long.
Every frame is rendered from 50 training and 10 test cameras around the scene, and a ground-truth mesh can be exported from the animation at any frame.
While DG-Mesh~\cite{liu2025dynamic} also provides ground-truth meshes, it is a single-view dataset, outside the scope of our multi-view setting.
To demonstrate the robustness of \ourmethod on real footage, we complement \benchmark with six scenes of Neural 3D Video~\cite{li2022neural} and seven scenes of SelfCap~\cite{xu2024representing}.

\paragraph{Baselines.}
\label{par:baselines}
We compare against five dynamic Gaussian splatting methods, 4DGS~\cite{wu20244d}, Deformable-3DGS~\cite{yang2024deformable}, SC-GS~\cite{huang2024scgs}, FreeTimeGS~\cite{wang2025freetimegs}, and AST-3DGS~\cite{tzathas2026adaptive}, and two geometry-based dynamic methods, D-2DGS~\cite{zhang2025dynamic} and DG-Mesh~\cite{liu2025dynamic}.
For all baselines except DG-Mesh, we extract meshes by TSDF fusion~\cite{curless1996volumetric} of the rendered depth maps.
DG-Mesh instead relies on differentiable Poisson reconstruction~\cite{peng2021shape} during both training and inference.
This restricts DG-Mesh to bounded foreground objects and rules out the real scenes, so we evaluate it on \benchmark only.

\paragraph{Metrics.}
\label{par:metrics}
We evaluate novel-view synthesis, geometry, and runtime.
For novel-view synthesis, we report PSNR, SSIM~\cite{wang2004image}, and LPIPS~\cite{zhang2018unreasonable} on the held-out views of all datasets.
For geometry, we compare meshes qualitatively on all datasets and quantitatively on \benchmark.
In each \benchmark scene, we extract ground-truth meshes at ten evenly spaced frames.
We measure position with the Chamfer distance and orientation with the normal Chamfer distance~\cite{mescheder2019occupancy}, but neither suffices alone.
Point distances barely penalize surface noise, so a noisy mesh can outscore a clean one, while normal angles can miss a significant surface offset.
Our primary metric is therefore a joint F-score, which requires matched points to agree both in position (distance within 1\% of the scene's bounding-box diagonal) and in orientation (normals within 20$^\circ$).
We compute the normal metrics both on surface samples and on normals rendered at the test views.
For runtime, we report training time.
For fairness, all methods run on a single NVIDIA A100 GPU with 80GB of VRAM.

\subsection{Results \& Comparisons}
\label{subsec:results}

We compare \ourmethod with the baselines in geometry, novel-view synthesis, and efficiency, and ablate its components.
The supplement adds per-scene results, mesh comparisons with all baselines, and the complete ablation study.

\paragraph{Geometry.}
\ourmethod reconstructs the cleanest meshes (\cref{fig:geometry}), and on \benchmark, where ground-truth geometry is available, they also score best in normal accuracy and joint F-score (\cref{tab:main_quality}).
Across all methods, the joint F-score agrees with the visual quality of the meshes, whereas the Chamfer distance barely penalizes their noise.

Both the representation and its extraction account for our lead.
Meshed by TSDF fusion, \ourmethod still surpasses every baseline in normal accuracy and joint F-score.
Marching tetrahedra on the vacancy field improves on both TSDF fusion and Poisson reconstruction.
This advantage comes from sampling the field at the Gaussians, which follow the geometry, so the mesh has an adaptive resolution that is finer where the surface has more detail.
In contrast, TSDF fusion and Poisson reconstruction solve on a fixed grid, where finer detail requires a finer grid everywhere and thus larger, more complex meshes.
Poisson reconstruction further oversmooths the surface and requires watertight, segmented objects, as our meshes reconstructed by Poisson show (\cref{tab:main_quality}).

\paragraph{Novel-view synthesis.}
Our representation of geometry does not come at the cost of appearance.
On every dataset, \ourmethod performs comparably to the dynamic splatting methods that model appearance only, and it renders the best novel views on \benchmark and the best PSNR and SSIM on SelfCap (\cref{tab:main_quality}).
It also keeps fast-moving parts sharp where other methods blur them (\cref{fig:nvs}).
Unlike our method, the other two geometry-based methods lose image quality (\cref{tab:main_quality}).
DG-Mesh~\cite{liu2025dynamic} renders far below every other method, and the quality of D-2DGS~\cite{zhang2025dynamic} does not translate from \benchmark to the real scenes.

\paragraph{Efficiency.}
\ourmethod trains much faster than the other two geometry-based methods and faster than most of the appearance-only methods (\cref{tab:main_time}).

\input{tables/ablations_main}

\paragraph{Ablations.}
\Cref{tab:ablations_main} summarizes our ablations, removing one component of \ourmethod at a time, and \suppref{sec:ablations} reports the complete study.
Without the trajectory fit, the real scenes lose their fast-moving parts, although their PSNR is nearly unchanged due to the static background (\suppref{fig:ablations_moving_parts}).
On the synthetic scenes, which have no background, the damage shows in every metric and grows as the motion gets larger and faster (\suppref{tab:ablations_anchoring}).
Halving the kernel count or removing the boundary inducing times also degrades appearance and geometry.

%% file: tables/ablations_main.tex
\begin{table}[t]
  \centering
  \footnotesize
  \setlength{\tabcolsep}{3pt}
  \begin{tabular}{@{}lcccc@{}}
    \toprule
    & \multicolumn{3}{c}{Synthetic} & Real \\
    \cmidrule(lr){2-4} \cmidrule(lr){5-5}
    Variant & PSNR & F$_\text{S}$ & F$_\text{R}$ & PSNR \\
    \midrule
    No trajectory fit & 26.88 & 0.295 & 0.436 & 28.98 \\
    Half kernel count & 32.97 & 0.470 & 0.691 & 29.32 \\
    No boundary inducing times & 33.85 & 0.497 & 0.716 & 29.34 \\
    \midrule
    \ourmethod & \textbf{34.84} & \textbf{0.515} & \textbf{0.745} & \textbf{29.40} \\
    \bottomrule
  \end{tabular}
  \caption{Ablations on three \benchmark scenes (synthetic) and four real scenes, removing one component at a time. F$_\text{S}$ and F$_\text{R}$ are the joint F-scores on surface samples and on normals rendered at the test views.}
  \label{tab:ablations_main}
  \vspace{-8pt}
\end{table}

%% file: sec/6_conclusion.tex
\section{Discussion}
\label{sec:discussion}

\paragraph{Limitations.}
Although \ourmethod recovers accurate meshes of complete dynamic scenes, it has several limitations.
First, it extracts each frame's mesh independently, so consecutive meshes are not guaranteed to share vertices or faces, and extracting temporally consistent meshes remains future work.
Second, its motion relies on the initial tracks, so thin, fast-moving parts that the tracker misses can be smeared or lost.
Third, since every Gaussian stores its own motion, its model is the largest among the compared methods.
Finally, like other methods that reconstruct geometry from images, it is affected by the ambiguity between albedo and shading and by narrow camera baselines, which can leave reliefs in the mesh or weakly constrained depth.
\Suppref{sec:limitations} discusses these limitations in depth, and \suppref{fig:limitations} illustrates its failure cases.

\paragraph{Conclusion.}
We presented \ourmethod, which jointly reconstructs the appearance and geometry of complete dynamic scenes.
Its per-Gaussian temporal motion model represents each Gaussian's motion over the whole sequence, so its geometry is defined at every frame.
A track-based initialization further lets it recover large and fast motion.
We also introduced \benchmark, a multi-view benchmark with ground-truth meshes at every frame.
\ourmethod extracts the most accurate per-frame meshes while keeping novel-view synthesis and training time on par with the strongest dynamic splatting methods.

%% file: sec/8_acknowledgements.tex
\section*{Acknowledgements}
This work was supported by the UW Reality Lab.

%% file: sec/7_llm_usage.tex
\section*{LLM Usage Statement}
\label{sec:llm_usage}

We used Claude (Anthropic), through Claude Code, as a coding and writing assistant.
In the research, it assisted with implementing and debugging the method, the evaluation infrastructure, and the figure scripts.
It also assisted with analyzing experimental results, discussing design alternatives, and surveying related work.
In the writing, it drafted and edited text, and assisted with the bibliography.
All use was directed and approved by the authors, who reviewed and verified all LLM-assisted code and text.
All research ideas, design decisions, experiments, and claims are the authors'.

%% file: sec/X_suppl.tex
This supplementary material complements the main paper.
\Cref{sec:limitations} discusses the limitations of \ourmethod.
\Cref{sec:implementation} details its implementation, including the extension of its static components to dynamic scenes, and \cref{sec:ablations} ablates its components.
\Cref{sec:benchmark} describes \benchmark, the inputs shared by all methods, and the evaluation protocol, and \cref{sec:additional_results} reports additional results, including per-scene metrics and further comparisons.
Videos of all scenes, comparing our method with every baseline, are available on \resultspage.

\section{Limitations}
\label{sec:limitations}

\input{figures/limitations}

\paragraph{Per-frame extraction.}
\ourmethod extracts the mesh of each frame independently, so consecutive meshes are not guaranteed to share vertices or faces.
This has two causes, the lack of a single tetrahedralization that stays valid over the whole sequence, and the vacancy field, which changes over time.
As the Gaussians move and rotate, their pivots shift and the structure of the tetrahedralization changes with them, especially under articulated motion and for parts hidden during part of the sequence, such as the pages of Book or the engine bay of Toyota.
Even if a stable tetrahedralization existed, the time-varying vacancy field would change which edges its 0.5 level set crosses, and with them the output of marching tetrahedra.
Extracting temporally consistent meshes is left for future work.

\paragraph{Motion recovery of thin, fast-moving parts.}
\label{par:limitation_motion}
Thin, fast-moving parts are hard to track, and the initialization of \ourmethod (\mainref{subsec:initialization}) may miss them.
Training then struggles to recover their motion, since a Gaussian receives gradients only from the pixels it covers.
Without a good initial track, a thin part that moves far between frames has either no Gaussians or static Gaussians.
In the latter case, the Gaussians match the part in a single frame, but in the others they occlude what lies behind them and increase the rendering error, so training lowers their opacity.
Either way, the part is smeared or lost.
\Cref{fig:limitations_motion} shows two instances, the rotating knob of Lego and the spinning wheel spokes of Bicycle.

\paragraph{Albedo and shading ambiguity.}
As with other methods that reconstruct geometry from appearance, \ourmethod faces the ambiguity between albedo and shading.
Sharp appearance boundaries, whether from lighting or from texture, can leave a relief in the mesh, as shown in \cref{fig:limitations_albedo}.
Using multiple viewpoints per frame keeps the ambiguity to a minimum.

\paragraph{Narrow camera baselines.}
A narrow, front-facing camera rig, as in Neural 3D Video~\cite{li2022neural}, sees surfaces parallel to its image planes with little parallax, which can leave their depth weakly constrained.
Such a surface can then be displaced toward the cameras while still rendering correctly from the training views, as shown in \cref{fig:limitations_baselines}.

\paragraph{Model size.}
\label{par:limitation_model_size}
\ourmethod has the largest models among the compared methods, on average 46\% larger than those of FreeTimeGS~\cite{wang2025freetimegs}, the second largest.
Although \ourmethod uses about 3 times fewer Gaussians than FreeTimeGS, each of its Gaussians stores about 4 times more parameters.
Fixing the inducing times and lengthscales at their initial values reduces the memory footprint by 18 to 20\% at a cost of up to 0.6\,dB in PSNR.
Adapting the number of kernels to each Gaussian's motion, as AST-3DGS~\cite{tzathas2026adaptive} does with its keyframes, would reduce it further, and we leave it to future work.

\section{Implementation Details}
\label{sec:implementation}

\subsection{Lifting static components to 4D}
\label{subsec:impl_lifting}

Three components of \ourmethod come from methods for static scenes, the importance-based densification and simplification of Mini-Splatting2~\cite{fang2026efficient}, the 3D filter of Mip-Splatting~\cite{yu2024mipsplatting}, and the normal-error densification of Gaussian Wrapping~\cite{gomez2026blobs}.
Each relies on its own sweep over all the training views, which accumulates per-Gaussian statistics.
A naive extension to dynamic scenes would be to sweep over every training view at every frame.
However, this is time-consuming and largely redundant, since neighboring Gaussians move coherently and their relative arrangement changes little between consecutive frames.
Instead, we set a budget of $N$ views per sweep (\cref{tab:impl_settings}).
If a full sweep over $V$ training views and $T$ frames exceeds it, the sweep renders the training views of every $K$-th frame only, with $K = \lfloor VT/N \rfloor$.
Below, we briefly review each component and describe how we lift it to 4D.

\paragraph{Mini-Splatting2.}
Mini-Splatting2~\cite{fang2026efficient} complements the gradient-based densification of 3DGS~\cite{kerbl20233d} with an importance-based pipeline.
The importance of a Gaussian is its accumulated blending weight across all the training views.
The pipeline densifies the scene by cloning the most important Gaussians while pruning negligible ones (aggressive cloning), and later simplifies it in two events that prune the least important Gaussians.
Since aggressive cloning densifies the scene much faster, densification ends at iteration 3K instead of 15K.
During training, Mini-Splatting2 also reinitializes the Gaussians from points sampled on the rendered depth, which removes floaters and Gaussians hidden behind surfaces.

To lift Mini-Splatting2 to 4D, we accumulate importance separately for each sampled frame.
Each aggressive cloning event selects the Gaussians to clone by their importance summed over all sampled frames.
Both the cloning and the simplification events prune a Gaussian only if it is unimportant in every sampled frame.
Gaussians visible in only a few frames, such as parts hidden during most of the sequence, therefore survive pruning.
We drop the depth reinitialization, since replacing the Gaussians would discard their trajectories.

\paragraph{3D mip filter.}
Mip-Splatting~\cite{yu2024mipsplatting} prevents aliasing by ensuring that each Gaussian is at least as large as a pixel of the training views that see it.
It convolves each Gaussian with an isotropic Gaussian kernel the size of a pixel at the closest of these views, and scales the opacity down so that the total contribution of each Gaussian is preserved.
This size is determined by the Gaussian's distance to the cameras, which varies from frame to frame in a dynamic scene.
To lift the filter to 4D, we determine each Gaussian's kernel size from its closest distance to the training views over the whole sequence, according to its position at each sampled frame.
As in Mip-Splatting, we recompute the filter whenever Gaussians are added or removed, and periodically after densification to account for the optimization updates.
Since each recomputation evaluates the motion model at every sampled frame, we increase the period from 100 to 1K iterations to reduce training time.

\paragraph{Normal-error densification.}
Normal-error densification was proposed in Gaussian Wrapping~\cite{gomez2026blobs} to close holes in the surfaces.
For each Gaussian, it accumulates the error between its oriented normal and the normal of the rendered depth across the training views.
The Gaussians with the largest error are then cloned with flipped normals, so that either orientation can be kept.
To lift normal-error densification to 4D, we accumulate each Gaussian's error over all the sampled frames.

\subsection{Mesh extraction}
\label{subsec:impl_extraction}

To extract the mesh at time $t$, we first apply the motion model and obtain each Gaussian's position, rotation, and oriented normal at $t$.
Then, we follow the two-pivot mesh extraction pipeline of Gaussian Wrapping~\cite{gomez2026blobs}.
Each Gaussian contributes two pivots, its center and the point three standard deviations from it along its oriented normal.
The pivots are tetrahedralized with a GPU Delaunay triangulation~\cite{cao2014gpu}.
Next, we evaluate the vacancy at each pivot from the training views captured at $t$.
Finally, we extract the mesh as the 0.5 level set of the vacancy with marching tetrahedra.
After extraction, we clean the mesh using the following steps:
\begin{itemize}
  \item We remove faces with a vertex farther from the center of the camera rig than five times its radius.
  \item We remove faces with an edge longer than 0.2 times the rig radius.
  \item We remove connected components of fewer than 5,000 faces.
  \item We remove faces whose barycenter vacancy exceeds the 99th percentile over all faces.
  This removes faces that marching tetrahedra places across empty space, for example between two nearby parts or across a depth edge.
\end{itemize}

By default, the vacancy is evaluated with the 3D mip filter on, which attenuates the Gaussians' opacity.
This attenuation can affect thin structures, whose Gaussians become nearly transparent, so that the vacancy around them stays just above 0.5 and removes them from the mesh.
We therefore add an opt-in thin-structures mode that evaluates the vacancy with the unattenuated opacities and keeps thin structures intact.

\input{tables/impl_settings}

\subsection{Hyperparameters}
\label{subsec:impl_hyperparameters}

To speed up training, we render color, depth, geometric and oriented normals, and alpha in a single pass of a custom CUDA rasterizer built on that of RaDe-GS~\cite{zhang2026radegs}.
We keep the densification and simplification schedule of Mini-Splatting2~\cite{fang2026efficient}, and add normal-error densification~\cite{gomez2026blobs}, which runs every 1K iterations from iteration 12K to 16K.
For the geometric losses, we set $\lambda_D = 0.02$ and $\lambda_N = 0.05$, and compute the depth normal $\mathbf{N}_d$ from a blend of 0.4 of the expected depth and 0.6 of the median depth.
$\mathcal{L}_D$ and $\mathcal{L}_N$ start at iterations 3K and 10K, after the importance-based densification and simplification events, respectively.
We set the learning rates of the motion model to $3\times10^{-5}$ for the position weights, inducing times, and lengthscales, and to $5\times10^{-3}$ for those of rotation.
We train for 30K iterations on a single NVIDIA A100 GPU.
We use the default hyperparameters of 3DGS~\cite{kerbl20233d} unless stated otherwise below.
\Cref{tab:impl_settings} lists the settings that vary across scenes, and the settings below apply to all scenes unless a dataset is named.
We decay the learning rate of the base positions over the first 18K iterations instead of 30K and, following Gaussian Wrapping~\cite{gomez2026blobs}, set the learning rates of the spherical harmonics to $1.3\times10^{-3}$ for the base color and $1.1\times10^{-4}$ for the higher bands.

\paragraph{Initialization.}
Each anchor frame starts Kalman tracks forward and backward through the sequence.
When a track finds no point near its predicted position in the next frame, it follows the motion of its nearest neighbors.
We discard tracks that follow their neighbors in over 90\% of the frames.

\section{Ablations}
\label{sec:ablations}

We ablate the components of \ourmethod, the motion model, the initialization, the geometric regularization, the densification, and the mesh extraction, changing one at a time.
Except for the mesh extraction, we evaluate each variant on seven scenes, Lego, Spiderman, and Book from \benchmark, and two scenes each from Neural 3D Video (Sear steak, Flame salmon) and SelfCap~\cite{xu2024representing} (Corgi1, Yoga).
\Cref{tab:ablations} reports the results averaged over the synthetic and the real scenes, and per scene for the initialization.

\input{tables/ablations}

\paragraph{Boundary inducing times.}
\label{par:ablation_boundary}
We remove the boundary inducing times ($B = 0$) or keep a single one on each side ($B = 1$).
Boundary inducing times significantly improve the quality of the first and last frames (\cref{tab:ablations_boundary,fig:ablations_boundary}), whose motion would otherwise be extrapolated from kernels on one side only.
The effect is less prominent on the real scenes, whose frames are dominated by static background.

\input{figures/ablations_boundary}

\paragraph{Interior inducing times.}
\label{par:ablation_interior}
We halve and quarter the number of interior inducing times, $M_{\text{pos}}$ and $M_{\text{rot}}$, in every scene.
With fewer kernels, the motion model cannot follow fast motion, and both appearance and geometry degrade (\cref{tab:ablations_interior}).
On the real scenes, the drop in PSNR is mild because static background dominates the frames, but the moving parts degrade visibly (\cref{fig:ablations_moving_parts}).

\paragraph{Trajectory fitting.}
\label{par:ablation_fit}
In the initialization (\mainref{subsec:initialization}), we fit each Gaussian's trajectory to its track.
Here, we discard the fitted trajectories, and initialize every Gaussian with zero motion (\cref{tab:ablations_anchoring}, rightmost column).
Optimization then fails to recover large motion within the training budget, and both appearance and geometry degrade severely on the synthetic scenes.
On the real scenes, PSNR drops only slightly because static background dominates the frames, but fast-moving parts are lost (\cref{fig:ablations_moving_parts}).

\input{figures/ablations_moving_parts}

\paragraph{Anchoring.}
\label{par:ablation_anchoring}
Scenes whose content is revealed over time are initialized with several anchor frames spread across the sequence.
Here, we scale each scene's default number of anchors by one half and one quarter, or use a single anchor at the first or middle frame (\cref{tab:ablations_anchoring}).
The best anchoring differs from scene to scene, and no single setting is best on all of them.
A single anchor suffices when the moving content is visible at the anchor frame, whereas content revealed over time requires anchors spread across the sequence, whose number then matters little.

\paragraph{Temporal-parameter learning.}
\label{par:ablation_temporal}
We freeze the inducing times, the lengthscales, or both, keeping only the kernel weights learned (\cref{tab:ablations_temporal}).
Frozen inducing times stay at the bin centers, uniformly spread over the sequence, and frozen lengthscales at the bin width.
Learning them slightly improves fast synthetic motion, and leaves the geometry unchanged.
On the real scenes, the effect is not measurable, since static background dominates the frames.
Frozen parameters become constants shared by all Gaussians and need not be stored, which reduces the model size by about 20\%, a mitigation of the model size limitation discussed in \cref{par:limitation_model_size}.

\paragraph{Geometric regularization.}
\label{par:ablation_regularizers}
We train without $\mathcal{L}_D$, without $\mathcal{L}_N$, or without both (\cref{tab:ablations_regularizers}).
Without $\mathcal{L}_N$, the oriented normals that guide extraction are left unconstrained, and the surface breaks into holes and fragments.
Without $\mathcal{L}_D$, broad surfaces become lumpy and perforated.
Only their combination yields clean meshes (\cref{fig:ablations}), at a small cost in PSNR.

\input{figures/ablations_regularizers}

\paragraph{Importance-based densification.}
\label{par:ablation_densification}
In the training section (\mainref{subsec:training}), we argue for the importance-based densification of Mini-Splatting2~\cite{fang2026efficient} over the slower scheme of static Gaussian Wrapping~\cite{gomez2026blobs}, which combines the vanilla densification of 3DGS~\cite{kerbl20233d} with the multi-view loss of GGGS~\cite{zhang2026geometry}.
Here, we put this choice to the test by training with the latter scheme (\cref{tab:ablations_densification}, and \cref{fig:ablations}, rightmost column).
This scheme degrades both appearance and geometry on the synthetic scenes, and improves only the PSNR of the real scenes, slightly.
On the real scenes, it can open holes or distort the geometry in textureless regions (\cref{fig:ablations}).
It also doubles the model size and the training time.

\paragraph{Mesh extraction.}
\label{par:ablation_extraction}
We replace our marching tetrahedra with the extraction routes of the baselines, TSDF fusion of the rendered depth and Poisson reconstruction from the Gaussian centers and normals, on \benchmark, the only dataset with ground-truth meshes (\mainref{tab:main_quality}).
Both routes yield worse normals and joint F-scores than marching tetrahedra on every scene, Poisson reconstruction most clearly (\cref{fig:ablations_extraction}).
Meshed by TSDF fusion, our models still lead every baseline in normal Chamfer and joint F-score, so much of the lead comes from the representation itself.

\input{figures/ablations_extraction}

\section{Datasets and Evaluation}
\label{sec:benchmark}
\input{tables/real_datasets}

\subsection{The \benchmark benchmark}
\label{subsec:benchmark_construction}
To build \benchmark, we obtained the models from Sketchfab~\cite{sketchfab}, Objaverse~\cite{deitke2023objaverse}, and the NeRF synthetic dataset~\cite{mildenhall2020nerf}.
We used or modified the existing animations of models that had them, and created animations for models that did not, yielding nine scenes of 100 to 170 frames (\cref{fig:benchmark}).
Each scene is captured by a fixed rig of 50 training and 10 test cameras, sampled uniformly at random on the upper hemisphere around the scene, and rendered in Blender~\cite{blender} Cycles at $800\times800$ pixels.
Ground-truth meshes can be exported at any frame directly from the Blender scene.
Since the meshes of consecutive frames are highly redundant, we choose to export ten per scene, at evenly spaced frames.

\input{figures/benchmark}

\subsection{Real datasets}
\label{subsec:real_datasets}
From Neural 3D Video~\cite{li2022neural}, we use all six scenes, downsampled by two to $1352\times1014$ pixels.
We hold out the central camera for testing, per the dataset convention.
From SelfCap~\cite{xu2024representing}, we follow the protocol of FreeTimeGS~\cite{wang2025freetimegs} on six clips of the Hair, Corgi, and Bike sequences, and add a clip of the Yoga sequence to evaluate a longer sequence.
Each SelfCap clip also holds out one camera for testing.
The SelfCap videos are $3840\times2160$ pixels, except for the Bike scenes, which are $1080\times1080$.
Following FreeTimeGS, we undistort every frame.
Undistortion leaves black regions along the image borders, which differ between cameras, so we crop all images of a scene by the smallest margin that removes all the black regions in every camera.
We then downsample the cropped undistorted images by two, except on Bike, where we keep the resulting resolution.
\Cref{tab:real_datasets} lists the frames, cameras, and resolution of every scene in both datasets.

\subsection{Data preprocessing}
\label{subsec:preprocessing}
To keep the comparisons fair and make sure every method trains on exactly the same data, we built a preprocessing pipeline that produces consistent camera poses and per-frame point clouds.
For each frame, we triangulate a sparse point cloud from the provided camera poses with hloc~\cite{sarlin2019coarse} and COLMAP~\cite{schonberger2016structure}.
\begin{enumerate}
  \item For each view, we use NetVLAD~\cite{arandjelovic2016netvlad} to find the five most similar views to match it against.
  \item For each view, we extract ALIKED~\cite{zhao2023aliked} features.
  \item We match the ALIKED features of each view pair from step 1 with RoMa~\cite{edstedt2024roma}.
  \item We triangulate the matches from the provided poses with COLMAP, and discard the points seen by fewer than three views or with a reprojection error above two pixels. On SelfCap, we relax these thresholds to two views and three pixels.
  \item The camera poses provided with the real datasets are slightly inaccurate. We therefore refine them by a bundle adjustment of each frame, with the scale fixed, and average each camera's refined poses over all frames, since the rig is static.
\end{enumerate}

On the real datasets, the triangulated points cover the static background only sparsely.
Every method therefore also receives a dense point cloud of the static background, from which we remove the moving subject, since the triangulated points already represent it.
On SelfCap, we take the dense point cloud provided with the dataset and remove the subject with a box around it.
Neural 3D Video does not provide dense point clouds, so we extract a dense point cloud for the middle frame of each scene with VGGT-$\Omega$~\cite{wang2026vggt} and align it to the provided camera poses.
We then remove the subject manually, along with the window panes and frames in scenes with open windows, so that the far field renders through them.

\subsection{Evaluation protocol}
\label{subsec:metrics}
We evaluate all methods with the same protocol.
We measure the image quality of the test views on all datasets, the mesh quality on \benchmark, the only dataset with ground-truth meshes, and the training time.
For image quality, we compute PSNR, SSIM, and LPIPS with a VGG backbone on every frame of every test camera, and average them over all test images.
The training time is the wall-clock time of training on a single 80GB NVIDIA A100 GPU.

For mesh quality, we compare each method's mesh with the ground-truth mesh at the ten evaluated frames of every \benchmark scene, and average the metrics over them.
We measure position accuracy with the point Chamfer distance between 200K points sampled uniformly over each mesh's surface, normalized by the diagonal of the ground-truth bounding box.
Point distances barely penalize surface noise.
A noisy surface that oscillates around the ground truth stays close to it everywhere, so it can outscore a clean surface with a small offset.
Surface noise, however, does affect the normals.
We therefore complement the traditional Chamfer distance with the normal Chamfer distance~\cite{mescheder2019occupancy}, which averages the angle between the normals of nearest samples on the two meshes.
Conversely, normal angles ignore surface offsets.
A surface displaced far from the ground truth, but parallel to it, keeps its normals and still scores perfectly.

The two Chamfer distances can rank methods in opposite orders.
One method may have a lower point Chamfer distance but a higher normal Chamfer distance than another.
Taken separately, the two scores cannot determine which of the two methods produces higher-quality meshes.
We therefore define a joint F-score, and compute its precision and recall with the following condition.
A sample and its nearest sample on the other mesh form a correct match when their distance is below 1\% of the diagonal and their normals differ by less than $20^\circ$.
Across all methods, the joint F-score agrees with the visual quality of the meshes more consistently than either the point or the normal Chamfer distance alone (\mainref{subsec:results}), and we therefore use it as our primary geometry metric.

For completeness, we also compute the joint F-score on rendered normals, defined on image pixels instead of 3D nearest neighbors, which scores only the parts of the surface visible from the test views.
At every test camera of an evaluated frame, we rasterize both meshes with nvdiffrast~\cite{laine2020modular} and apply the same match condition to the depths and normals of each pixel.

\input{tables/per_scene_benchmark_image}
\input{tables/per_scene_benchmark_mesh}
\input{tables/per_scene_real}

Lastly, on SelfCap~\cite{xu2024representing}, all methods need one post-processing step before evaluation.
The SelfCap cameras stand on a half-circle rig, so parts of the rig appear in the views of the cameras facing them.
Every method reconstructs these parts with floaters near the cameras, which can occlude novel views.
We therefore remove from every method's model, after training, the Gaussians that lie within 15\% of the scene extent from a training camera and inside its frustum.
At inference, we do the same for every novel-view camera.

\section{Additional Results}
\label{sec:additional_results}

This section reports additional results.
We give the per-scene results of every method on every dataset, compare the meshes of all methods on \benchmark and on the real scenes, including the baselines that the main paper leaves out, add novel-view comparisons on two more SelfCap scenes, and show our meshes over time on several scenes.
Videos of our results on all scenes, compared side by side with every baseline, are available on \resultspage.

\Cref{tab:per_scene_benchmark_image,tab:per_scene_benchmark_mesh,tab:per_scene_real} report the per-scene results on every dataset.
On \benchmark (\cref{tab:per_scene_benchmark_image}), \ourmethod has the best image quality on eight of the nine scenes.
In mesh quality (\cref{tab:per_scene_benchmark_mesh}), while other methods lead the point Chamfer distance on several scenes, \ourmethod has the best normal Chamfer distance on all but one scene, and the best joint F-scores across the board.
These results are consistent with our analysis of the mesh metrics in \cref{subsec:metrics}.
On Neural 3D Video (\cref{tab:per_scene_real}), no method leads on every scene.
\ourmethod has the best SSIM on most scenes, while 4DGS has the best mean PSNR and FreeTimeGS the best LPIPS.
On SelfCap, \ourmethod has the best mean PSNR and SSIM, and FreeTimeGS the best LPIPS.

\Cref{fig:mesh_comparison_benchmark} extends the mesh comparison of the main paper (\mainref{subsec:results}) to every baseline on six \benchmark scenes.
The appearance-only methods produce noisy, fragmented surfaces.
The geometry-based D-2DGS and DG-Mesh produce cleaner surfaces but degrade fine detail, such as the raised page of Book, the tip of the Elephant trunk, the brake cables of Bicycle, and the cabin of Beetle.
\ourmethod preserves both broad surfaces, such as the pages of Book, and thin structures, such as the brake cables of Bicycle and the seats and steering wheel of Beetle.

\Cref{fig:mesh_comparison_real} compares the meshes of all methods except DG-Mesh on the real scenes.
With no ground-truth meshes available, we present the test image of each scene as reference.
The appearance-only methods turn the static background into noisy, crumpled surfaces, most visibly on the walls and blinds of Coffee martini, and blur fine detail, such as the ears of the corgi in Corgi2.
D-2DGS produces smoother surfaces but distorts the geometry and loses detail, such as the wall and the person's hands and shaker in Coffee martini, the upper wheel in Bike2, and the corgi's ears and the person's shoes in Corgi2.
On Yoga and Hair1, its meshes break apart.
\ourmethod reconstructs clean walls, floors, and furniture, together with the moving subjects and their fine detail, such as the corgi's ears.

\Cref{fig:nvs_real} compares novel views of all methods except DG-Mesh on the real scenes, cropped to the moving subjects, since all methods reconstruct the static background well.
Most baselines blur or distort fast-moving parts, such as the hands and tongs in Cook spinach, the face and knife in Cut roasted beef, the corgi's head in Corgi1, and the face in Hair1.
SC-GS and D-2DGS smear the entire cook in both Neural 3D Video scenes.
Overall, \ourmethod renders the moving parts sharper and closer to the ground truth.
Specifically in Hair1, although no method fully recovers the individual strands, \ourmethod best preserves the flung hair.

\Cref{fig:meshes_over_time} shows \ourmethod's meshes over time on five additional \benchmark scenes.
Each row follows one motion: the Lego excavator lowering its bucket, the book closing, the Toyota opening its doors, hood, and trunk, Spiderman's full sequence, and the bicycle steering its front wheel.
The surfaces stay sharp through fast and large motion, and thin structures such as the bicycle's brake cables persist across frames.

\FloatBarrier
\input{figures/mesh_comparison_benchmark}
\input{figures/mesh_comparison_real}
\input{figures/nvs_real}
\input{figures/meshes_over_time}

%% file: figures/limitations.tex
\begin{figure*}[t]
  \centering
  \newlength{\limlabel}\newlength{\limgap}\newlength{\limgroup}\newlength{\limw}
  \setlength{\limlabel}{0.2in}
  \setlength{\limgap}{0.03in}
  \setlength{\limgroup}{0.12in}
  \setlength{\limw}{\dimexpr(\linewidth-\limlabel-4\limgap-\limgroup)/6\relax}
  \newcommand{\limgroupw}{\dimexpr3\limw+2\limgap\relax}
  \newcommand{\limtitles}{\makebox[\limw]{GT mesh}\hspace{\limgap}\makebox[\limw]{Rendered image}\hspace{\limgap}\makebox[\limw]{Extracted mesh}}
  \newcommand{\limrow}[2]{\makebox[\limlabel]{\rotatebox[origin=c]{90}{#1}}\raisebox{-0.5\height}{\includegraphics[width=\dimexpr\linewidth-\limlabel\relax]{images/limitations/#2}}}
  \begin{subfigure}{\linewidth}
    \centering
    {\footnotesize
    \makebox[\linewidth][l]{\hspace{\limlabel}\limtitles\hspace{\limgroup}\limtitles}\\[1pt]
    \limrow{Lego}{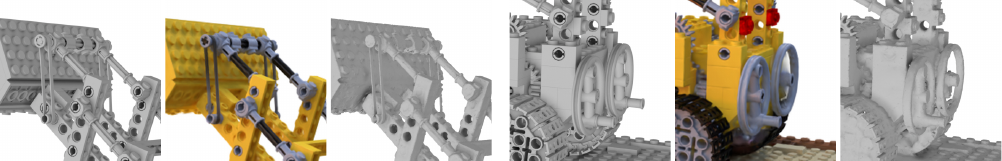}\\[2pt]
    \limrow{Bicycle}{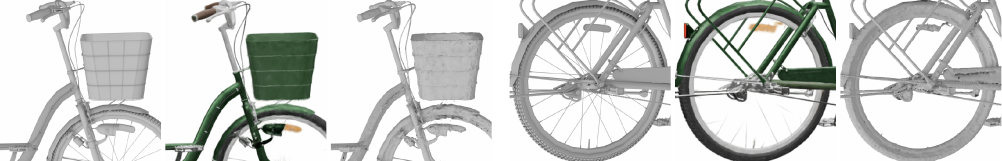}\\[1pt]
    \makebox[\linewidth][l]{\hspace{\limlabel}\makebox[\limgroupw]{Static thin structures}\hspace{\limgroup}\makebox[\limgroupw]{Fast-moving thin structures}}\par}
    \caption{Motion recovery of thin, fast-moving parts.
      Left: static thin structures, such as the bucket cables of Lego and the brake cables of Bicycle, are cleanly reconstructed.
      Right: fast-moving thin structures are smeared or lost, since the initialization may miss their tracks (\cref{par:limitation_motion}).
      On Lego, the rotating knob (right) is lost, while the static knob beside it is recovered, and on Bicycle the spinning wheel spokes are lost.}
    \label{fig:limitations_motion}
  \end{subfigure}\\[6pt]
  \newlength{\limwb}
  \setlength{\limwb}{\dimexpr(\linewidth-2\limgap-\limgroup)/4\relax}
  \newcommand{\limgroupwb}{\dimexpr2\limwb+\limgap\relax}
  \newcommand{\limtitlesb}{\makebox[\limwb]{Rendered image}\hspace{\limgap}\makebox[\limwb]{Extracted mesh}}
  \begin{subfigure}{\linewidth}
    \centering
    {\footnotesize
    \makebox[\linewidth][l]{\limtitlesb\hspace{\limgroup}\limtitlesb}\\[1pt]
    \includegraphics[width=\linewidth]{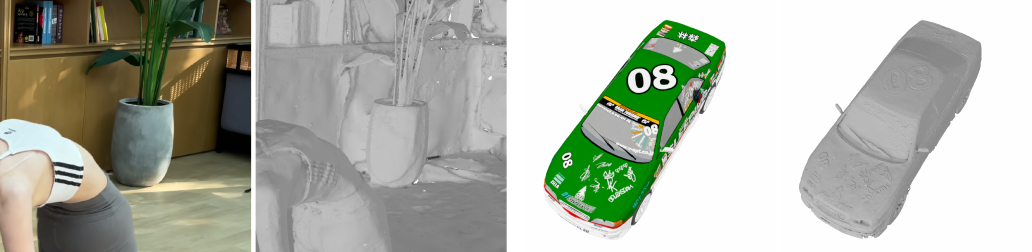}\\[1pt]
    \makebox[\linewidth][l]{\makebox[\limgroupwb]{Lighting (Yoga)}\hspace{\limgroup}\makebox[\limgroupwb]{Texture (Toyota)}}\par}
    \caption{Albedo and shading ambiguity.
      Left: the edge of the sunlit patch on the vase in Yoga leaves a ridge in the mesh.
      Right: the painted decals on Toyota leave a relief in the mesh.
      In both, a sharp appearance boundary, from lighting in Yoga and from texture in Toyota, is partly explained as geometry.}
    \label{fig:limitations_albedo}
  \end{subfigure}\\[6pt]
  \begin{subfigure}{\linewidth}
    \centering
    {\footnotesize
    \makebox[\linewidth][l]{\limtitlesb\hspace{\limgroup}\limtitlesb}\\[1pt]
    \includegraphics[width=\linewidth]{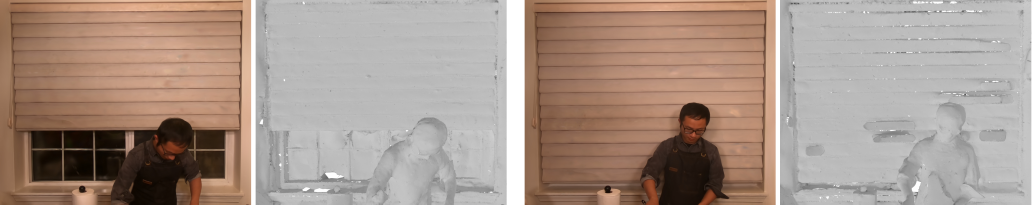}\\[1pt]
    \makebox[\linewidth][l]{\makebox[\limgroupwb]{Cut roasted beef}\hspace{\limgroup}\makebox[\limgroupwb]{Sear steak}}\par}
    \caption{Narrow camera baselines.
      Cut roasted beef and Sear steak share the kitchen and the front-facing rig of Neural 3D Video.
      Left: in Cut roasted beef, the window blind is reconstructed flat.
      Right: in Sear steak, some of its slats bulge toward the cameras, which see them with little parallax.}
    \label{fig:limitations_baselines}
  \end{subfigure}
  \caption{Limitations of \ourmethod: \subref{fig:limitations_motion} motion recovery of thin, fast-moving parts, \subref{fig:limitations_albedo} albedo and shading ambiguity, and \subref{fig:limitations_baselines} narrow camera baselines.}
  \label{fig:limitations}
\end{figure*}

%% file: tables/impl_settings.tex
\begin{table}[t]
  \centering
  \footnotesize
  \setlength{\tabcolsep}{4pt}
  \begin{tabular}{@{}lccccc@{}}
    \toprule
    \textbf{Scenes} & \textbf{Frames} & \textbf{\boldmath$M_{\text{pos}}$} & \textbf{\boldmath$M_{\text{rot}}$} & \textbf{Anchors} & \textbf{Sweep} \\
    \midrule
    \multicolumn{6}{@{}l}{\textbf{\benchmark}} \\
    \cmidrule{1-6}
    Spiderman & 120 & 30 & 30 & 20 & 1,200 \\
    Beetle, Toyota & 150 & 20 & 20 & 20 & 1,500 \\
    Book & 170 & 20 & 20 & 20 & 1,700 \\
    Bicycle$^\dagger$, Ficus$^\dagger$ & 100 & 20 & 20 & 20 & 1,000 \\
    Alligator, Elephant, Lego & 100 & 20 & 20 & 20 & 1,000 \\
    \midrule
    \multicolumn{6}{@{}l}{\textbf{Neural 3D Video}} \\
    \cmidrule{1-6}
    All scenes & 300 & 50 & 30 & 1 & 1,200 \\
    \midrule
    \multicolumn{6}{@{}l}{\textbf{SelfCap}} \\
    \cmidrule{1-6}
    Hair1, Hair2 & 60 & 20 & 20 & 3 & 1,440 \\
    Corgi1, Corgi2 & 60 & 15 & 15 & 3 & 1,440 \\
    Bike1$^\dagger$, Bike2$^\dagger$ & 60 & 15 & 15 & 3 & 1,440 \\
    Yoga & 300 & 50 & 30 & 1 & 1,440 \\
    \bottomrule
  \end{tabular}
  \caption{Per-scene settings, the number of frames, the interior inducing times of the position ($M_{\text{pos}}$) and rotation ($M_{\text{rot}}$) motion, the anchor frames of the initialization, and the view budget for each importance sweep. $^\dagger$Thin-structures mode.}
  \label{tab:impl_settings}
\end{table}

%% file: tables/ablations.tex
\begin{table*}[t]
  \centering
  \begin{subtable}[t]{0.32\linewidth}
    \centering
    \footnotesize
    \setlength{\tabcolsep}{3pt}
    \begin{tabular}[t]{@{}lcccc@{}}
      \toprule
      & \multicolumn{2}{c}{Synthetic} & \multicolumn{2}{c}{Real} \\
      \cmidrule(lr){2-3} \cmidrule(lr){4-5}
      $B$ & Ends & All & Ends & All \\
      \midrule
      0 & 31.45 & 33.85 & 28.68 & 29.34 \\
      1 & 36.00 & 34.80 & \textbf{29.37} & \textbf{29.42} \\
      \midrule
      2 (default) & \textbf{36.32} & \textbf{34.84} & 29.35 & 29.40 \\
      \bottomrule
    \end{tabular}
    \caption{Boundary inducing times $B$ per side. Ends is the PSNR over the first and last five frames.}
    \label{tab:ablations_boundary}
  \end{subtable}
  \hfill
  \begin{subtable}[t]{0.32\linewidth}
    \centering
    \footnotesize
    \setlength{\tabcolsep}{3pt}
    \begin{tabular}[t]{@{}lcccc@{}}
      \toprule
      & \multicolumn{3}{c}{Synthetic} & Real \\
      \cmidrule(lr){2-4} \cmidrule(lr){5-5}
      $M_{\text{pos}}, M_{\text{rot}}$ & PSNR & F$_\text{S}$ & F$_\text{R}$ & PSNR \\
      \midrule
      $\times 1/4$ & 29.59 & 0.381 & 0.555 & 29.25 \\
      $\times 1/2$ & 32.97 & 0.470 & 0.691 & 29.32 \\
      \midrule
      $\times 1$ (default) & \textbf{34.84} & \textbf{0.515} & \textbf{0.745} & \textbf{29.40} \\
      \bottomrule
    \end{tabular}
    \caption{Interior inducing times $M_{\text{pos}}$ and $M_{\text{rot}}$, scaled from each scene's default counts.}
    \label{tab:ablations_interior}
  \end{subtable}
  \hfill
  \begin{subtable}[t]{0.32\linewidth}
    \centering
    \footnotesize
    \setlength{\tabcolsep}{3pt}
    \begin{tabular}[t]{@{}lcccc@{}}
      \toprule
      & \multicolumn{3}{c}{Synthetic} & Real \\
      \cmidrule(lr){2-4} \cmidrule(lr){5-5}
      Regularizers & PSNR & F$_\text{S}$ & F$_\text{R}$ & PSNR \\
      \midrule
      None & \textbf{35.07} & 0.424 & 0.586 & \textbf{29.78} \\
      $\mathcal{L}_D$ only & 35.06 & 0.479 & 0.680 & 29.35 \\
      $\mathcal{L}_N$ only & 34.90 & 0.513 & 0.741 & 29.66 \\
      \midrule
      Both (default) & 34.84 & \textbf{0.515} & \textbf{0.745} & 29.40 \\
      \bottomrule
    \end{tabular}
    \caption{Geometric regularization.}
    \label{tab:ablations_regularizers}
  \end{subtable}

  \vspace{0.8em}

  \begin{subtable}[t]{\linewidth}
    \centering
    \footnotesize
    \setlength{\tabcolsep}{4pt}
    \begin{tabular}[t]{@{}lcccccccc@{}}
      \toprule
      Scene & Anchors & Default & $\times 1/2$ & $\times 1/4$ & Middle & First & Many & No fit \\
      \midrule
      Lego & 20 & 32.83 & 33.05 & 33.26 & 33.27 & \textbf{33.28} & -- & 29.93 \\
      Spiderman & 20 & \textbf{33.88} & 33.66 & 33.41 & 30.54 & 30.76 & -- & 21.46 \\
      Book & 20 & 37.82 & 37.87 & 37.77 & \textbf{37.98} & 37.06 & -- & 29.25 \\
      Sear steak & First & 32.95 & -- & -- & 32.81 & -- & \textbf{33.01} & 32.81 \\
      Flame salmon & First & 27.16 & -- & -- & 27.27 & -- & \textbf{27.67} & 26.98 \\
      Corgi1 & 3 & 27.17 & 26.98 & \textbf{27.72}$^\dagger$ & \textbf{27.72}$^\dagger$ & 27.63 & -- & 26.87 \\
      Yoga & Middle & 30.30 & -- & -- & -- & \textbf{30.41} & 30.20 & 29.26 \\
      \bottomrule
    \end{tabular}
    \caption{Initialization, PSNR per scene. Anchors is each scene's default, $\times 1/2$ and $\times 1/4$ scale the default anchor count, Middle and First use a single anchor at that frame, Many uses 20 anchors on Neural 3D Video and 3 on Yoga, and No fit discards the fitted trajectories. $^\dagger$Same run.}
    \label{tab:ablations_anchoring}
  \end{subtable}

  \vspace{0.8em}

  \begin{subtable}[t]{0.48\linewidth}
    \centering
    \footnotesize
    \setlength{\tabcolsep}{3pt}
    \begin{tabular}[t]{@{}lccccc@{}}
      \toprule
      & \multicolumn{3}{c}{Synthetic} & Real & \\
      \cmidrule(lr){2-4} \cmidrule(lr){5-5}
      Frozen & PSNR & F$_\text{S}$ & F$_\text{R}$ & PSNR & Size \\
      \midrule
      Inducing times & 34.75 & 0.512 & \textbf{0.745} & 29.39 & $-18\%$ \\
      Lengthscales & 34.58 & 0.507 & 0.739 & 29.46 & $-0.6\%$ \\
      Both & 34.52 & 0.506 & 0.739 & \textbf{29.50} & $\mathbf{-19\%}$ \\
      \midrule
      None (default) & \textbf{34.84} & \textbf{0.515} & \textbf{0.745} & 29.40 & -- \\
      \bottomrule
    \end{tabular}
    \caption{Temporal-parameter learning. Size is the change in stored parameters per Gaussian.}
    \label{tab:ablations_temporal}
  \end{subtable}
  \hfill
  \begin{subtable}[t]{0.48\linewidth}
    \centering
    \footnotesize
    \setlength{\tabcolsep}{3pt}
    \begin{tabular}[t]{@{}lcccccc@{}}
      \toprule
      & \multicolumn{3}{c}{Synthetic} & Real & & \\
      \cmidrule(lr){2-4} \cmidrule(lr){5-5}
      Densification & PSNR & F$_\text{S}$ & F$_\text{R}$ & PSNR & $N$ & Time \\
      \midrule
      Multi-view loss & 31.99 & 0.334 & 0.482 & \textbf{29.53} & 766k & 71m \\
      \midrule
      Importance-based (default) & \textbf{34.84} & \textbf{0.515} & \textbf{0.745} & 29.40 & \textbf{386k} & \textbf{35m} \\
      \bottomrule
    \end{tabular}
    \caption{Importance-based densification against the scheme of static Gaussian Wrapping. $N$ is the mean number of Gaussians and Time the mean training time.}
    \label{tab:ablations_densification}
  \end{subtable}
  \caption{Ablations of \ourmethod on three \benchmark scenes (synthetic) and four real scenes, with the default settings marked. F$_\text{S}$ and F$_\text{R}$ are the joint F-scores on surface samples and on rendered normals.}
  \label{tab:ablations}
\end{table*}

%% file: figures/ablations_boundary.tex
\begin{figure}[t]
  \centering
  \begin{subfigure}{\linewidth}
    \centering
    \includegraphics[width=\linewidth]{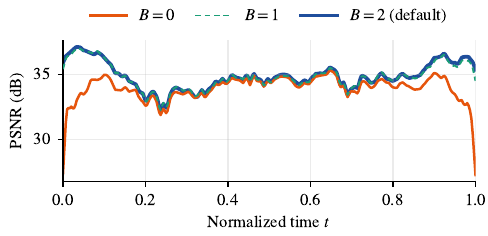}
    \caption{Synthetic scenes. Without boundary points, quality degrades toward both ends of the sequence and drops sharply at the first and last frames, whose motion is extrapolated from kernels on one side only.}
    \label{fig:ablations_boundary_synthetic}
  \end{subfigure}

  \vspace{0.5em}

  \begin{subfigure}{\linewidth}
    \centering
    \includegraphics[width=\linewidth]{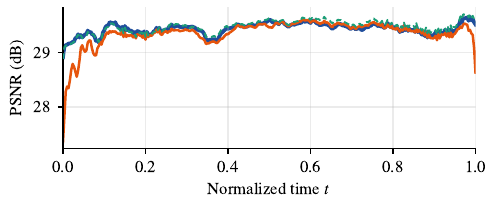}
    \caption{Real scenes. The drop at the ends is weaker, since static background dominates the frames.}
    \label{fig:ablations_boundary_real}
  \end{subfigure}
  \caption{Per-frame PSNR over normalized time with $B$ boundary inducing times per side, averaged over the synthetic ablation scenes \subref{fig:ablations_boundary_synthetic}, and the real ablation scenes \subref{fig:ablations_boundary_real}.
    We compare $B = 0$ (orange) and $B = 1$ (green) with the default $B = 2$ (dark blue) at the two ends of the sequence.}
  \label{fig:ablations_boundary}
\end{figure}

%% file: figures/ablations_moving_parts.tex
\begin{figure*}[t]
  \centering
  \setlength{\tabcolsep}{1pt}
  \renewcommand{\arraystretch}{0.6}
  \newlength{\mpcol}\setlength{\mpcol}{\dimexpr(\linewidth-1.2em-10\tabcolsep)/5\relax}
  \begin{tabular}{@{}m{1.2em}m{\mpcol}m{\mpcol}m{\mpcol}m{\mpcol}m{\mpcol}@{}}
    & \multicolumn{1}{c}{\footnotesize Ground truth} & \multicolumn{1}{c}{\footnotesize $M$ (default)} & \multicolumn{1}{c}{\footnotesize $M/2$} & \multicolumn{1}{c}{\footnotesize $M/4$} & \multicolumn{1}{c}{\footnotesize $M$, no fit} \\
    \rotatebox{90}{\footnotesize Spiderman} & \includegraphics[width=\mpcol]{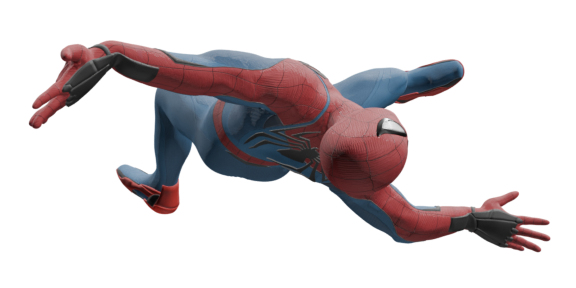} & \includegraphics[width=\mpcol]{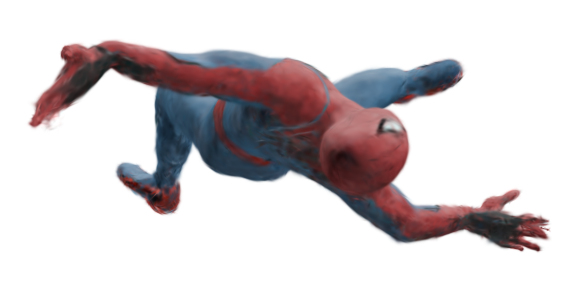} & \includegraphics[width=\mpcol]{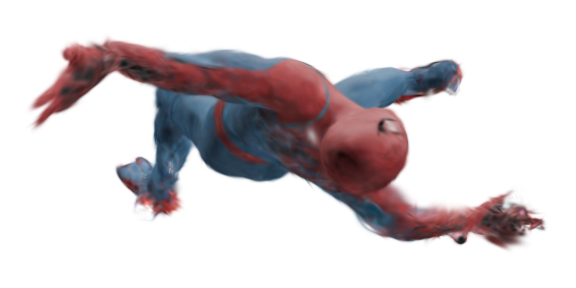} & \includegraphics[width=\mpcol]{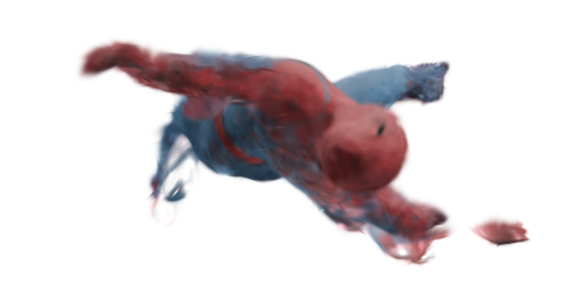} & \includegraphics[width=\mpcol]{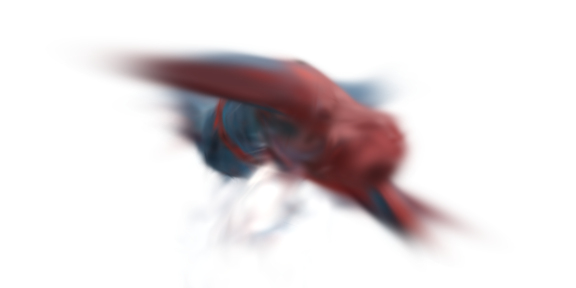} \\
    \rotatebox{90}{\footnotesize Flame salmon} & \includegraphics[width=\mpcol]{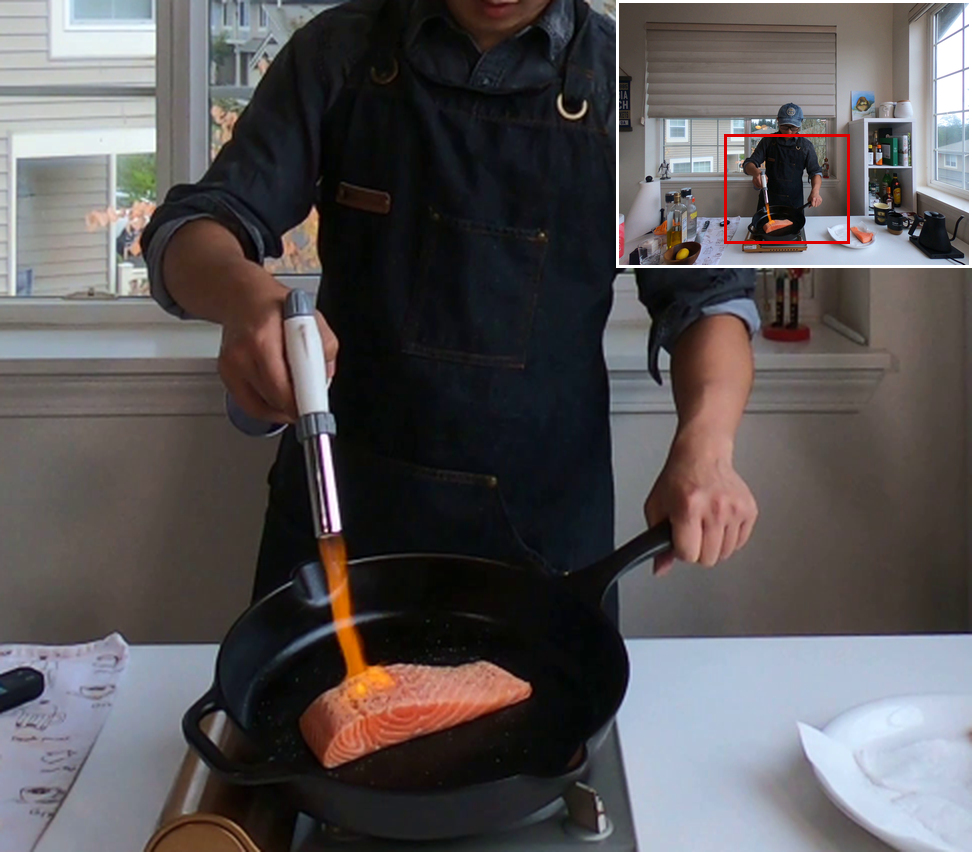} & \includegraphics[width=\mpcol]{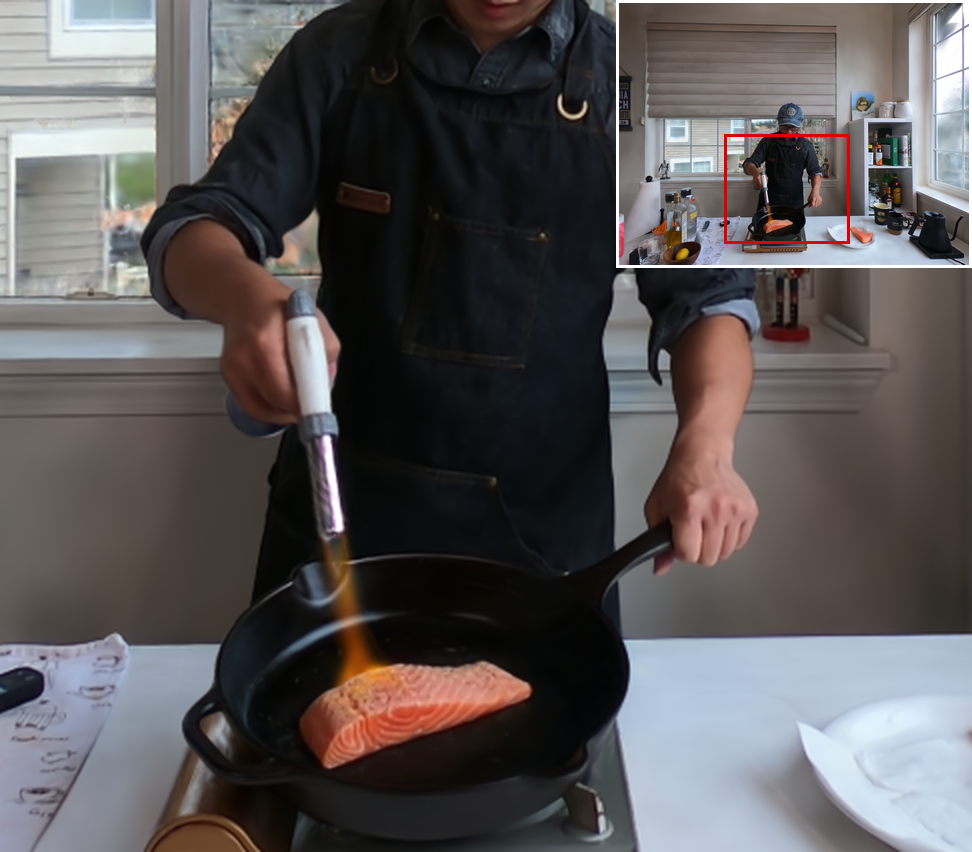} & \includegraphics[width=\mpcol]{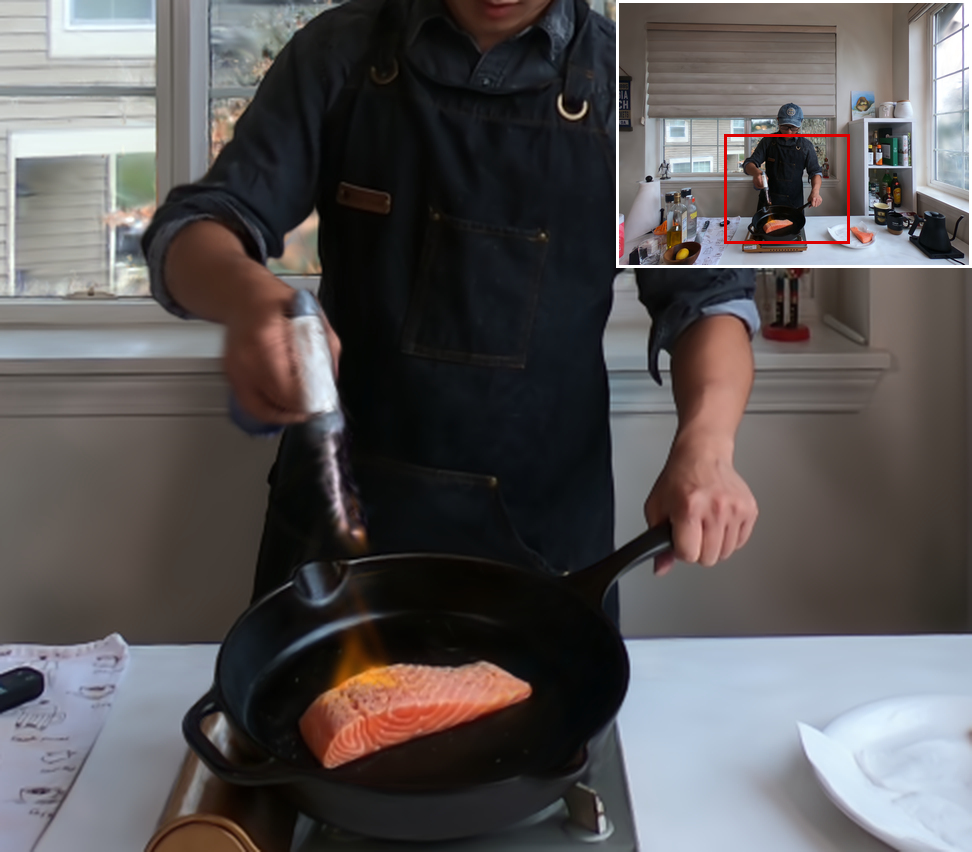} & \includegraphics[width=\mpcol]{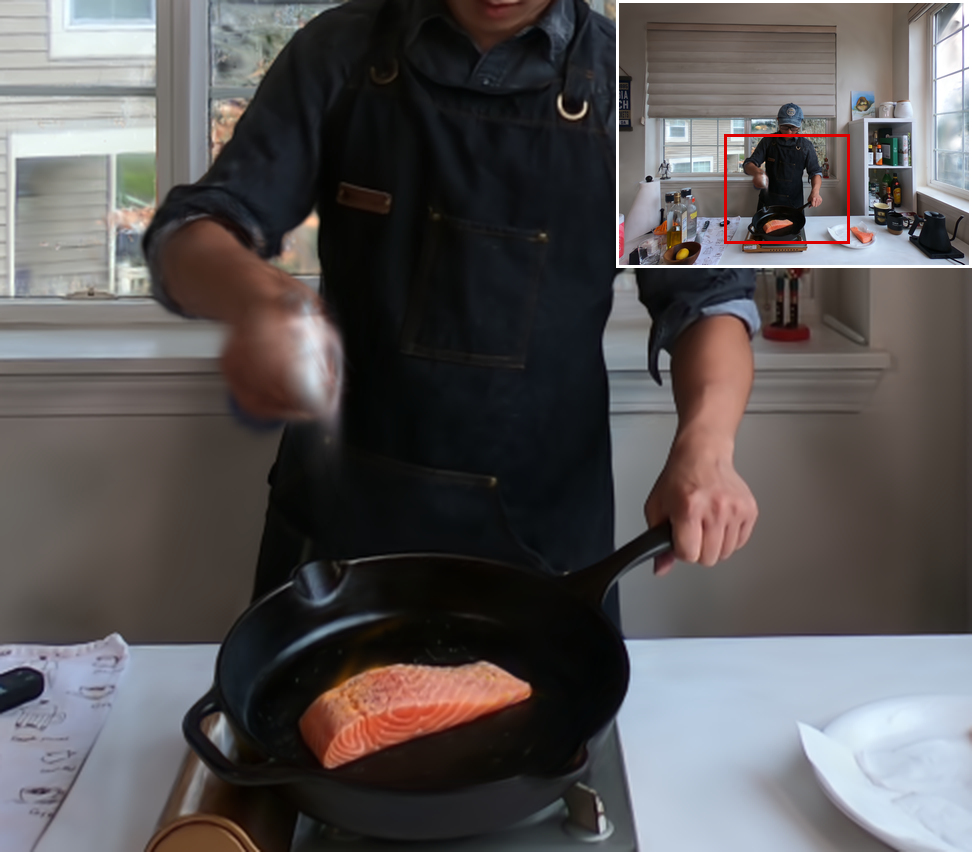} & \includegraphics[width=\mpcol]{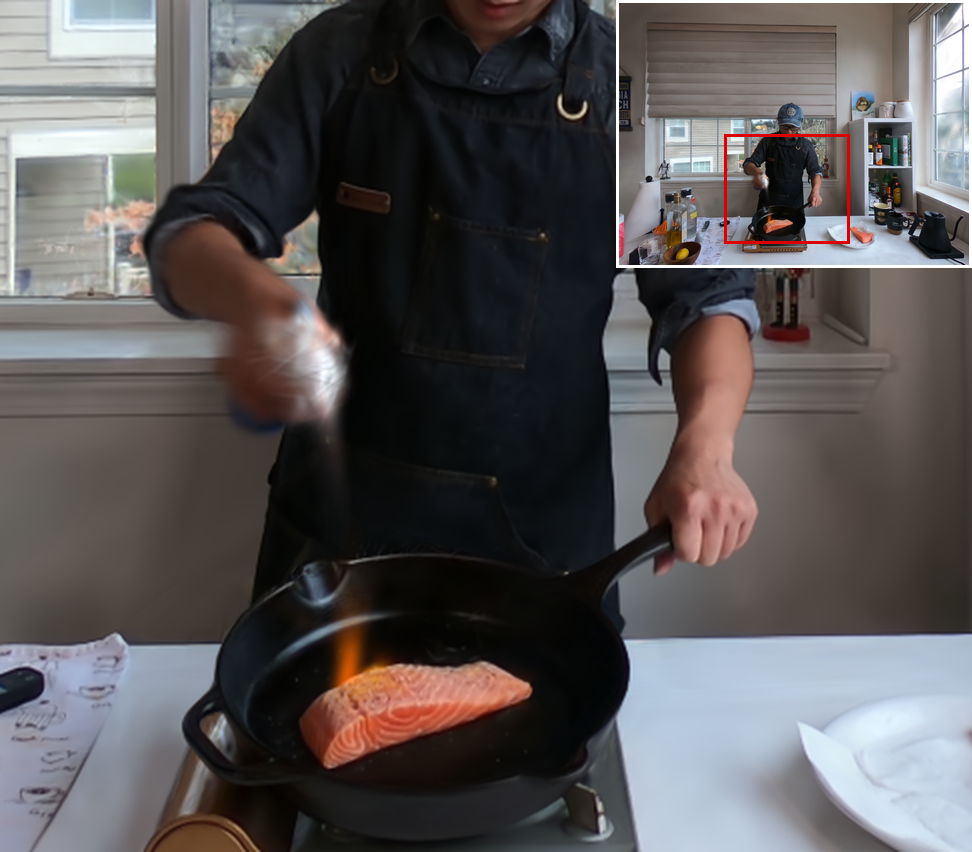} \\
  \end{tabular}
  \caption{Ground truth and test renders of Spiderman and Flame salmon with the default number of interior inducing times $M$, with $M/2$ and $M/4$, and without the trajectory fit.
    Insets show the full Flame salmon frames, with the crop outlined in red.
    Spiderman's hands and Flame salmon's torch smear at $M/2$ and break up at $M/4$, while the default $M$ keeps them sharp.
    Without the trajectory fit, both scenes lose their moving parts, although the PSNR of the real scene barely changes.}
  \label{fig:ablations_moving_parts}
\end{figure*}

%% file: figures/ablations_regularizers.tex
\begin{figure*}[t]
  \centering
  \setlength{\tabcolsep}{1pt}
  \renewcommand{\arraystretch}{0.6}
  \newlength{\rgcol}\setlength{\rgcol}{\dimexpr(\linewidth-1.2em-10\tabcolsep)/5\relax}
  \begin{tabular}{@{}m{1.2em}m{\rgcol}m{\rgcol}m{\rgcol}m{\rgcol}m{\rgcol}@{}}
    & \multicolumn{1}{c}{\footnotesize No reg.} & \multicolumn{1}{c}{\footnotesize $\mathcal{L}_D$ only} & \multicolumn{1}{c}{\footnotesize $\mathcal{L}_N$ only} & \multicolumn{1}{c}{\footnotesize Both (default)} & \multicolumn{1}{c}{\footnotesize Multi-view} \\
    \rotatebox{90}{\footnotesize Flame salmon} & \includegraphics[width=\rgcol]{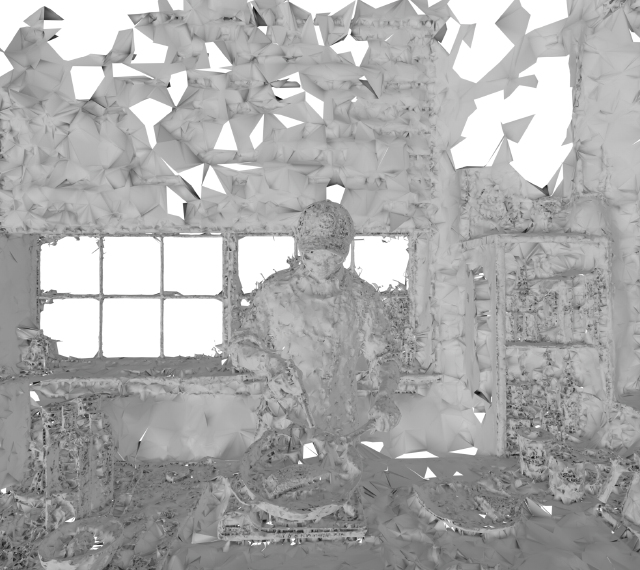} & \includegraphics[width=\rgcol]{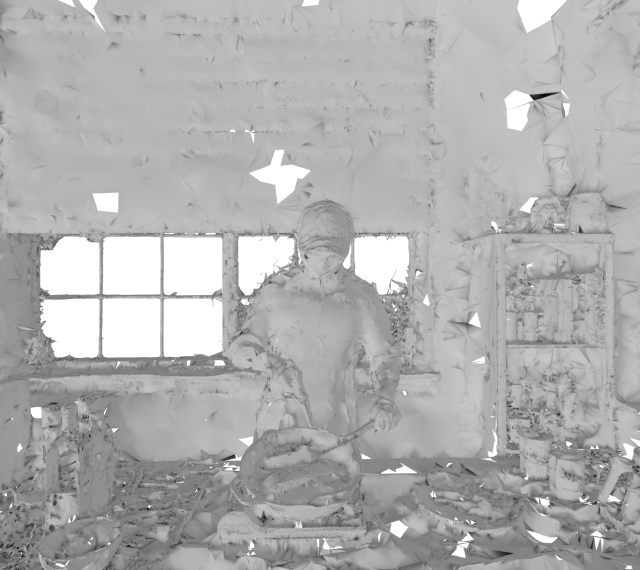} & \includegraphics[width=\rgcol]{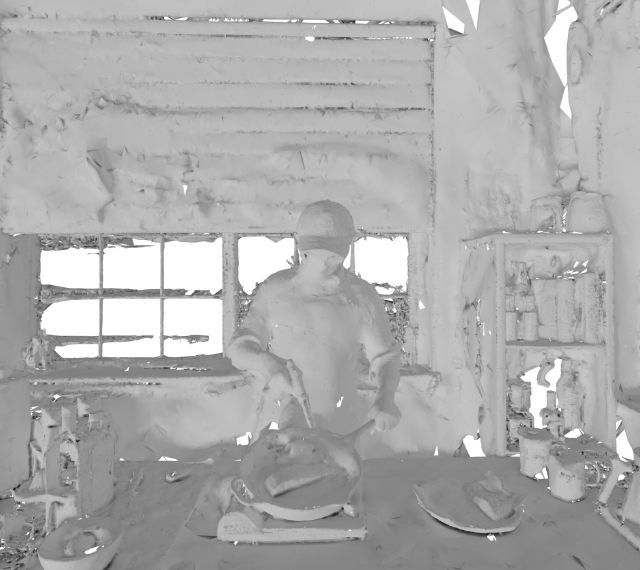} & \includegraphics[width=\rgcol]{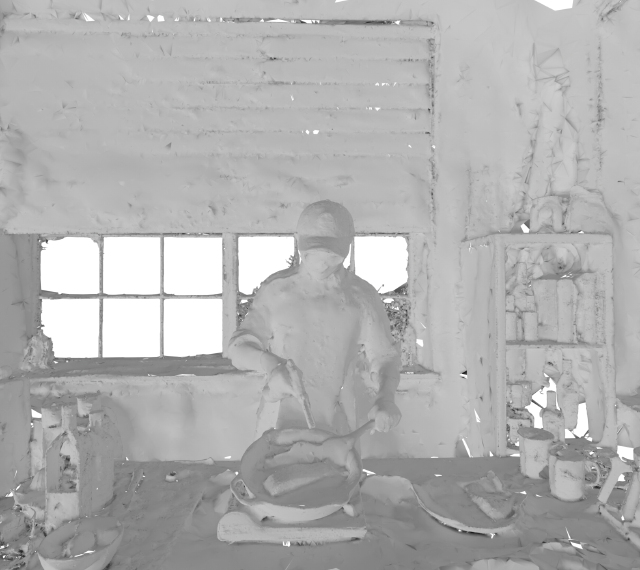} & \includegraphics[width=\rgcol]{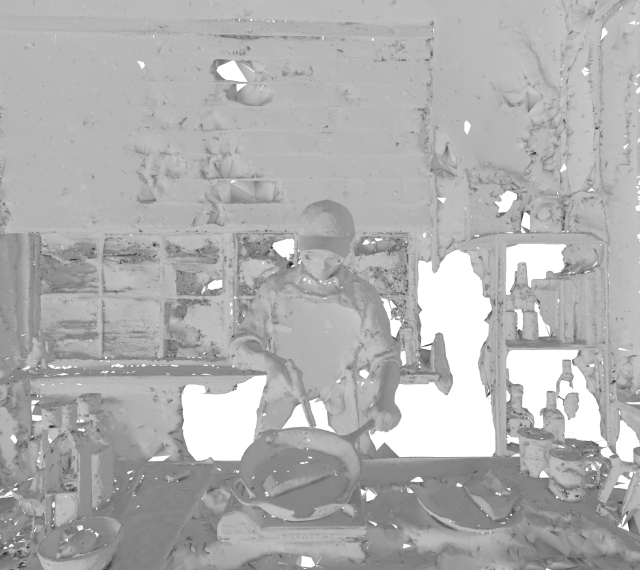} \\
    \rotatebox{90}{\footnotesize Sear steak} & \includegraphics[width=\rgcol]{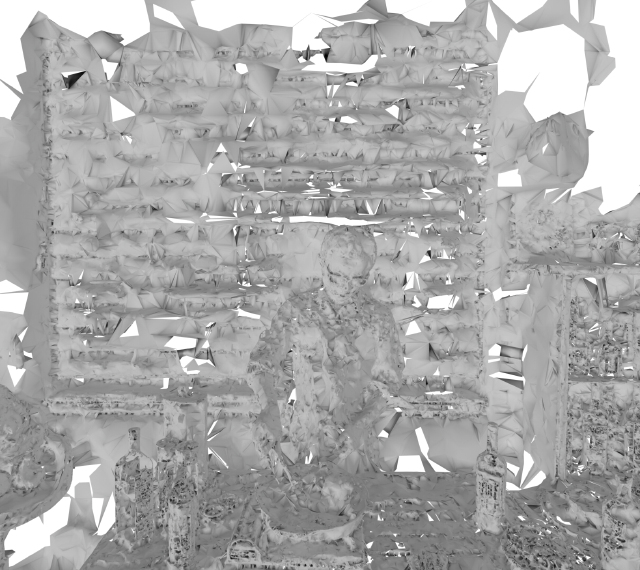} & \includegraphics[width=\rgcol]{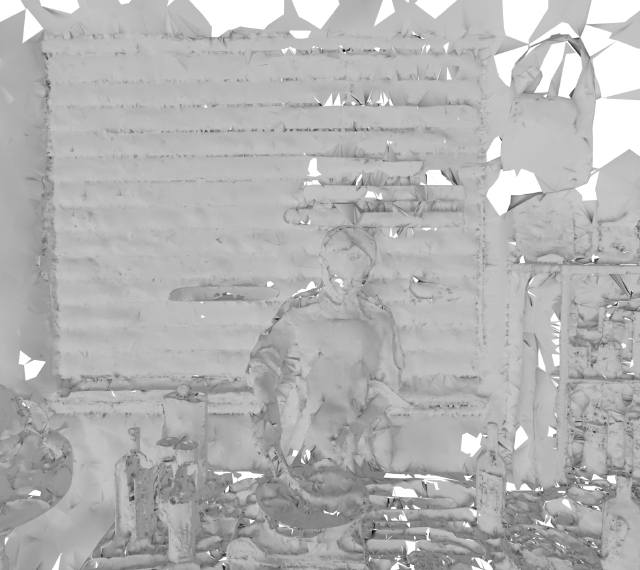} & \includegraphics[width=\rgcol]{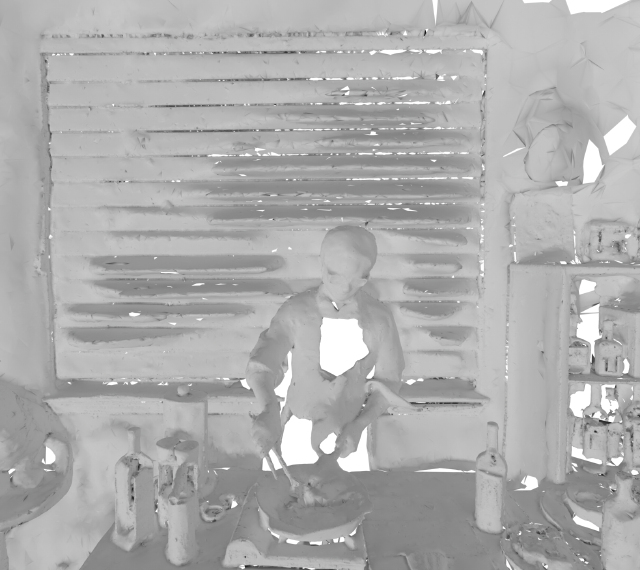} & \includegraphics[width=\rgcol]{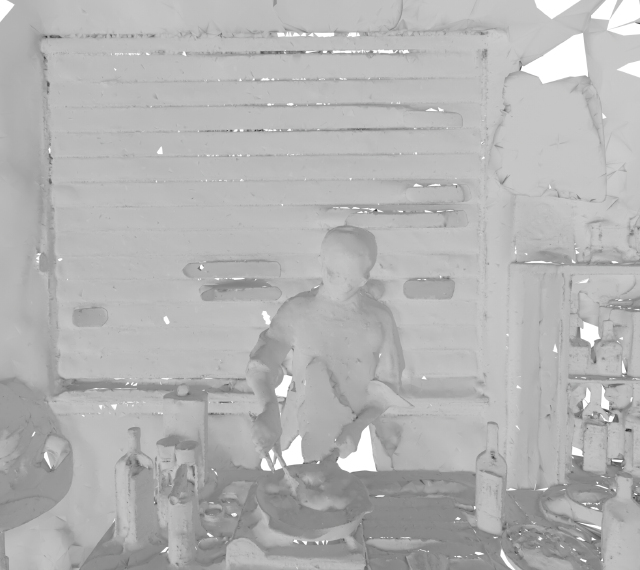} & \includegraphics[width=\rgcol]{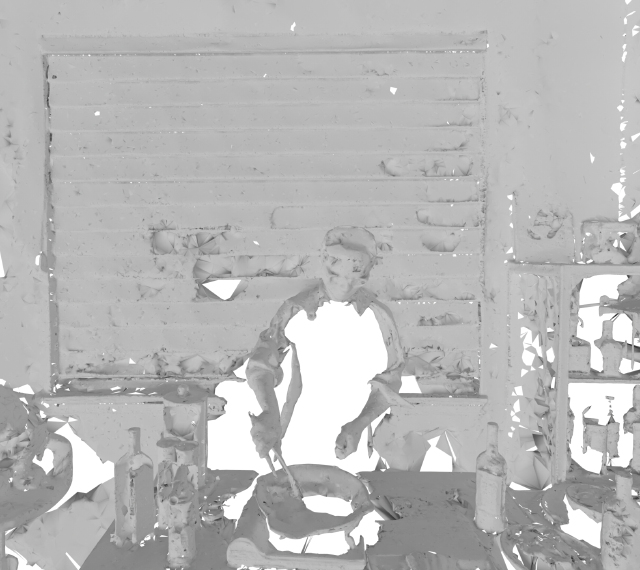} \\
    \rotatebox{90}{\footnotesize Corgi1} & \includegraphics[width=\rgcol]{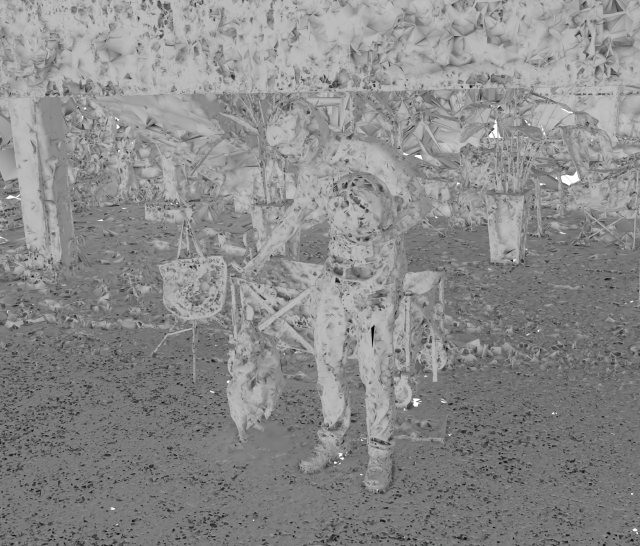} & \includegraphics[width=\rgcol]{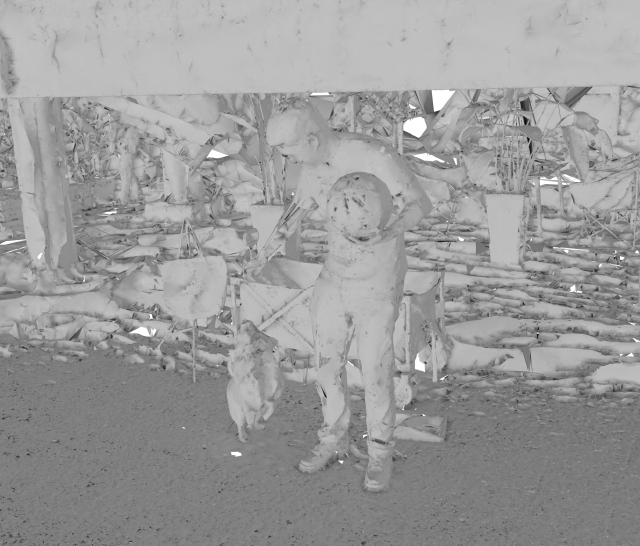} & \includegraphics[width=\rgcol]{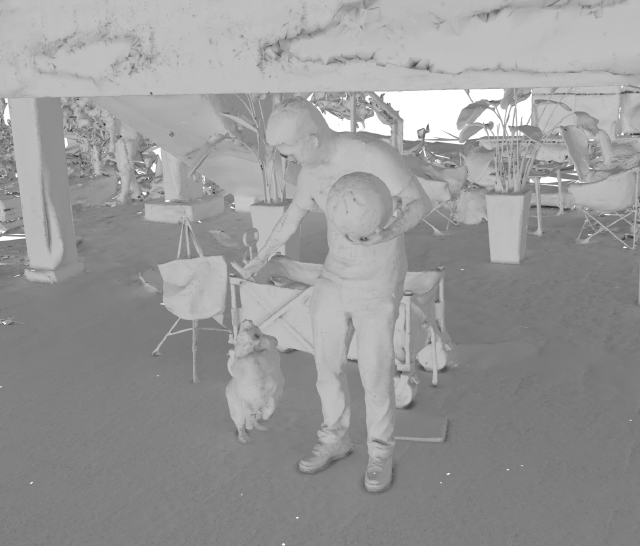} & \includegraphics[width=\rgcol]{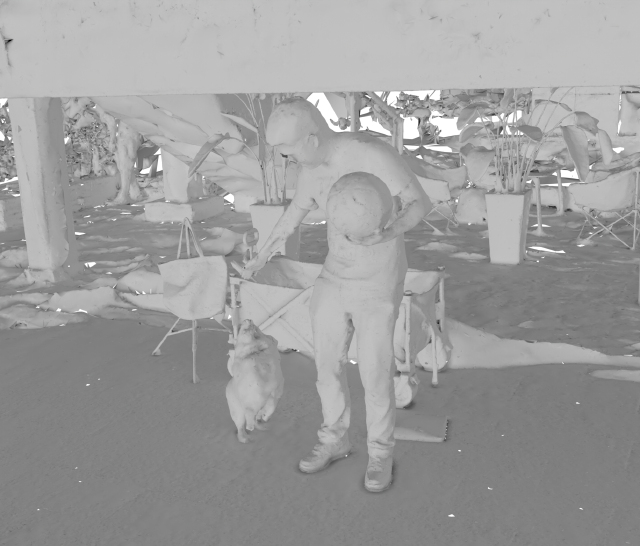} & \includegraphics[width=\rgcol]{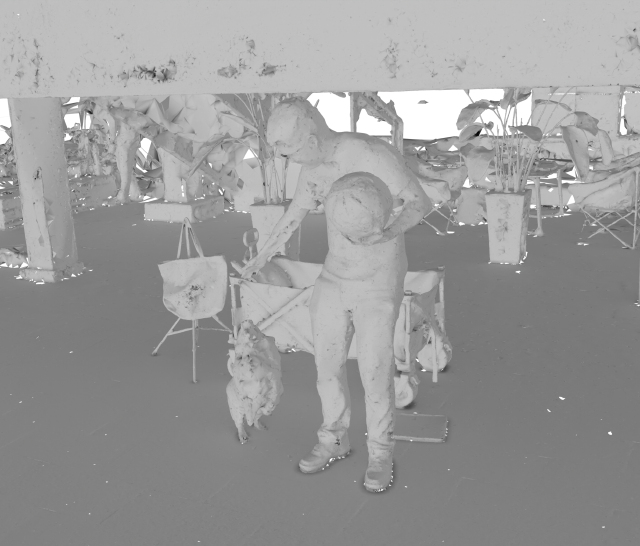} \\
  \end{tabular}
  \caption{Meshes of three real ablation scenes with no regularizers, only with $\mathcal{L}_D$, only with $\mathcal{L}_N$, with both (the default), and with the multi-view densification scheme of static Gaussian Wrapping.
    All meshes were rendered from the test camera.
    Without $\mathcal{L}_D$, broad surfaces such as the blinds break into ridges and detached slats. Without $\mathcal{L}_N$, surfaces are noisy and fragmented.
    The multi-view scheme opens holes or distorts the geometry in textureless regions.}
  \label{fig:ablations}
  \vspace{-1pt}
\end{figure*}

%% file: figures/ablations_extraction.tex
\begin{figure*}[t]
  \centering
  \setlength{\tabcolsep}{1pt}
  \renewcommand{\arraystretch}{0.6}
  \newlength{\excol}\setlength{\excol}{\dimexpr(\linewidth-1.2em-8\tabcolsep)/4\relax}
  \begin{tabular}{@{}m{1.2em}m{\excol}m{\excol}m{\excol}m{\excol}@{}}
    & \multicolumn{1}{c}{\footnotesize Ground truth} & \multicolumn{1}{c}{\footnotesize Marching tetrahedra} & \multicolumn{1}{c}{\footnotesize TSDF fusion} & \multicolumn{1}{c}{\footnotesize Poisson} \\
    \rotatebox{90}{\footnotesize Lego} & \includegraphics[width=\excol]{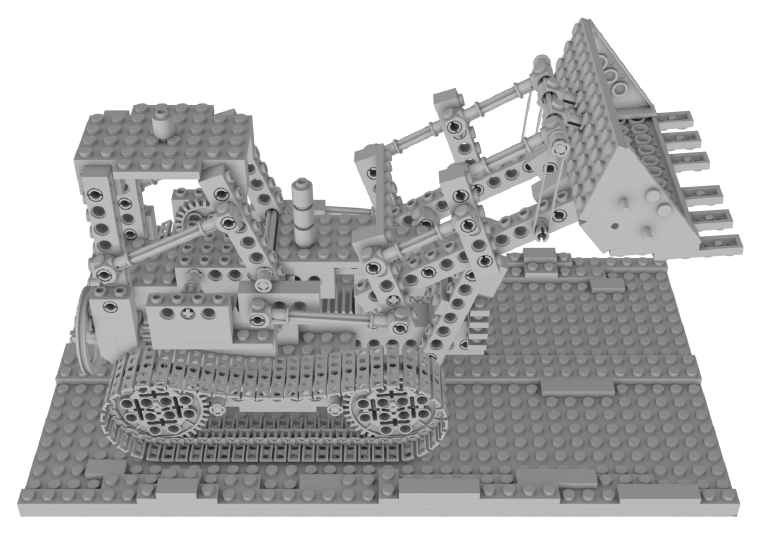} & \includegraphics[width=\excol]{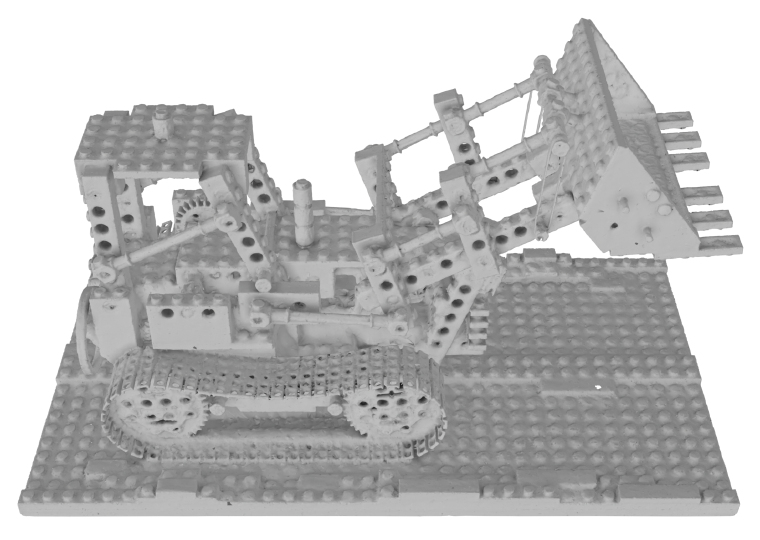} & \includegraphics[width=\excol]{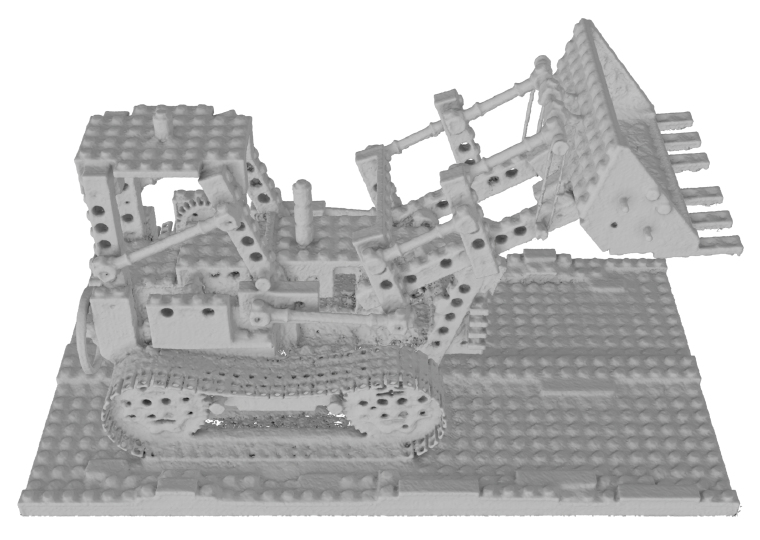} & \includegraphics[width=\excol]{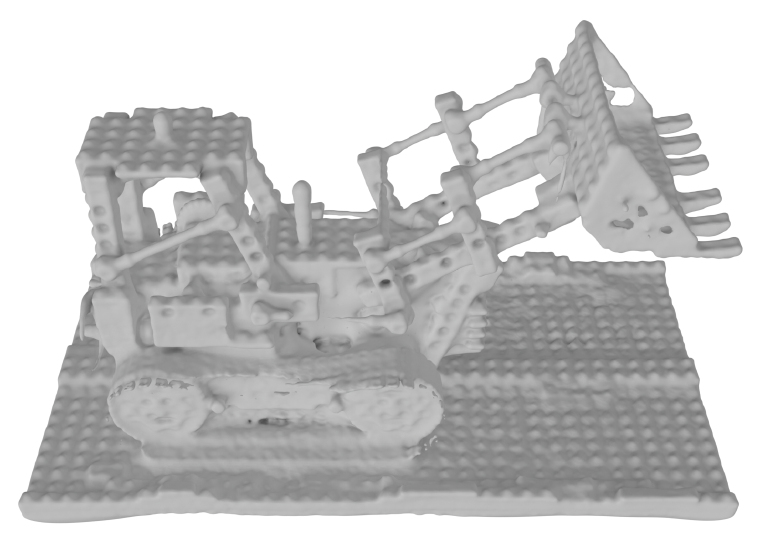} \\
    \rotatebox{90}{\footnotesize Bicycle} & \includegraphics[width=\excol]{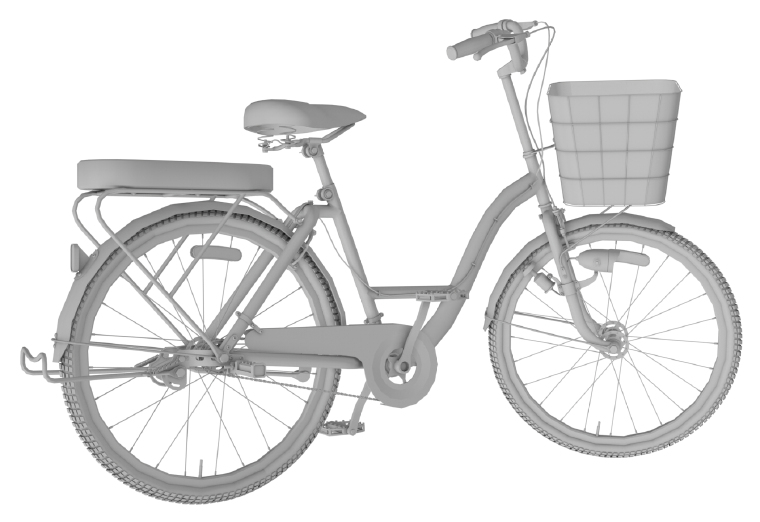} & \includegraphics[width=\excol]{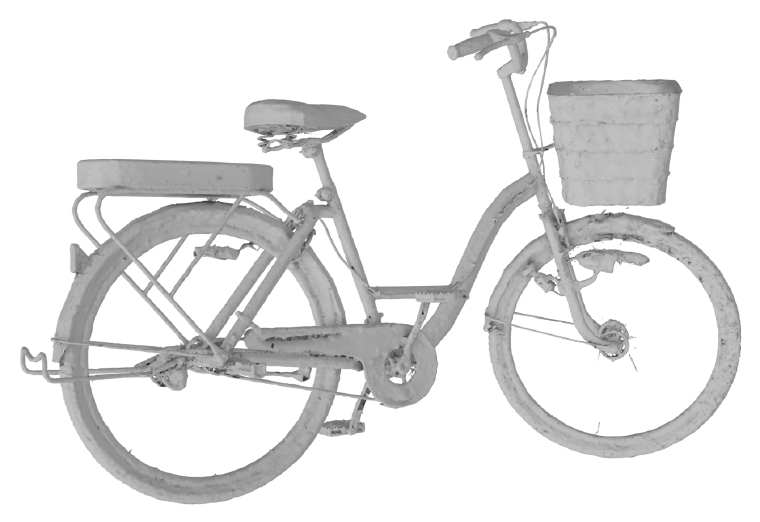} & \includegraphics[width=\excol]{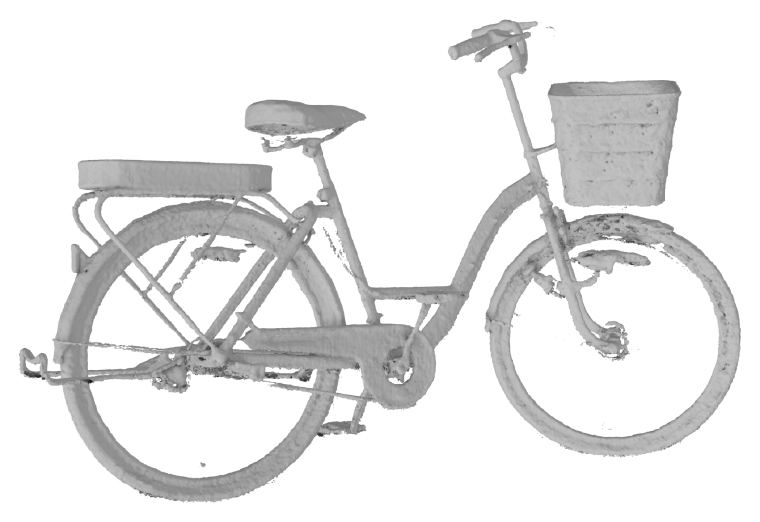} & \includegraphics[width=\excol]{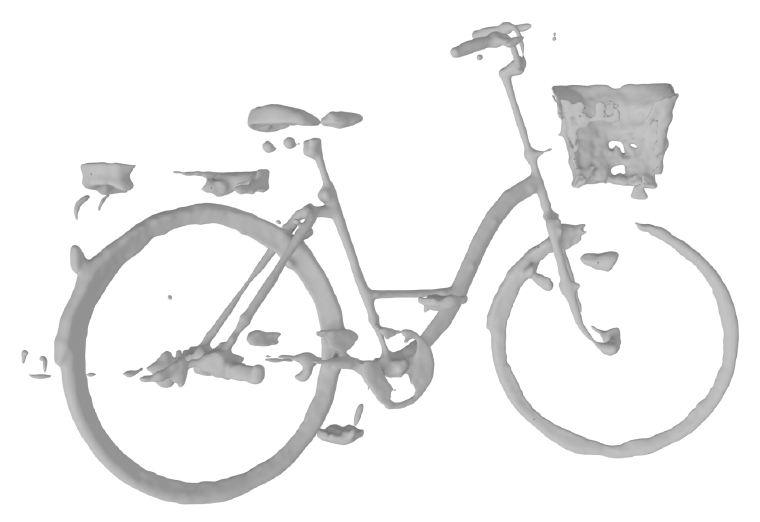} \\
    \rotatebox{90}{\footnotesize Ficus} & \includegraphics[width=\excol]{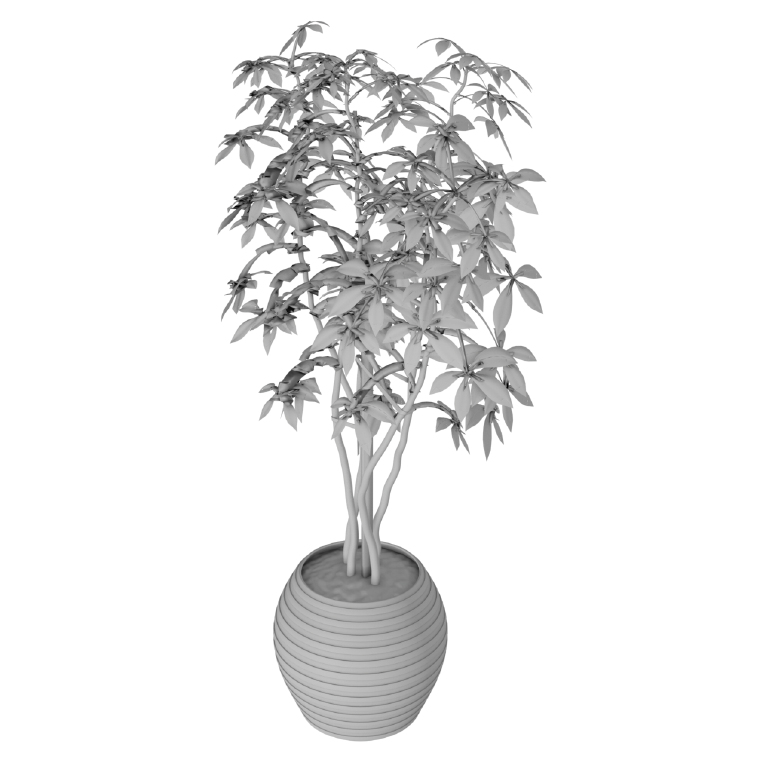} & \includegraphics[width=\excol]{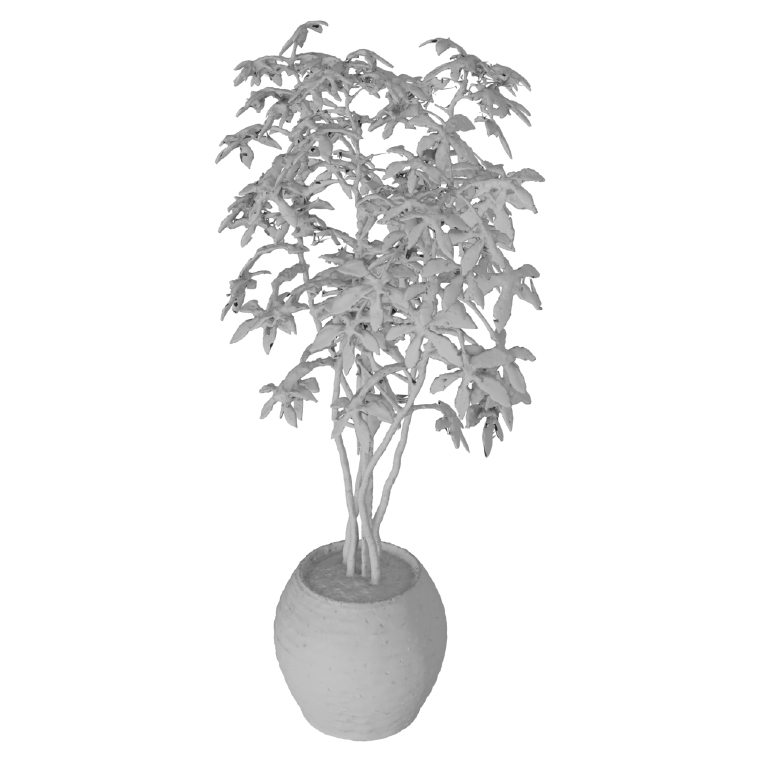} & \includegraphics[width=\excol]{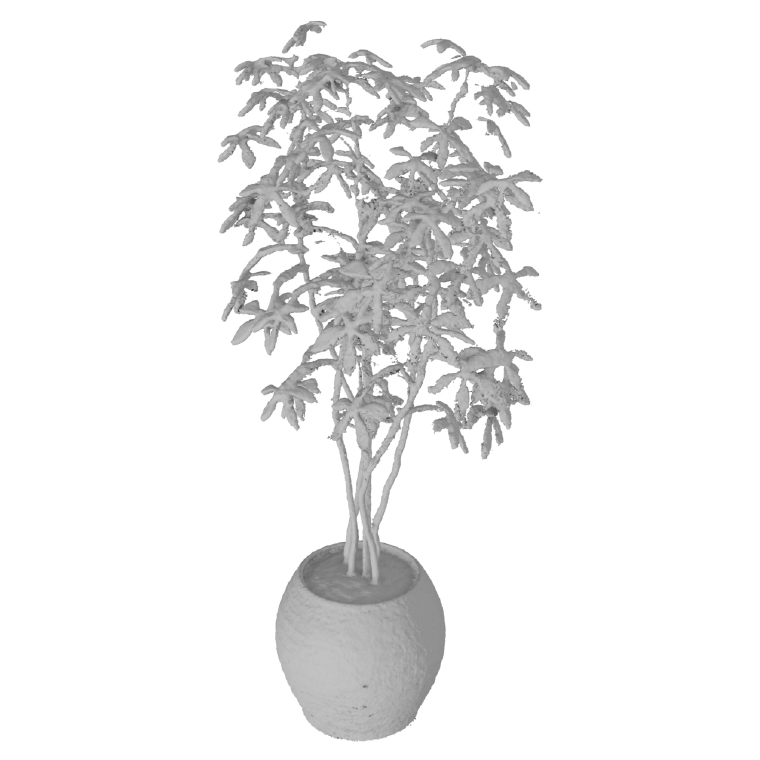} & \includegraphics[width=\excol]{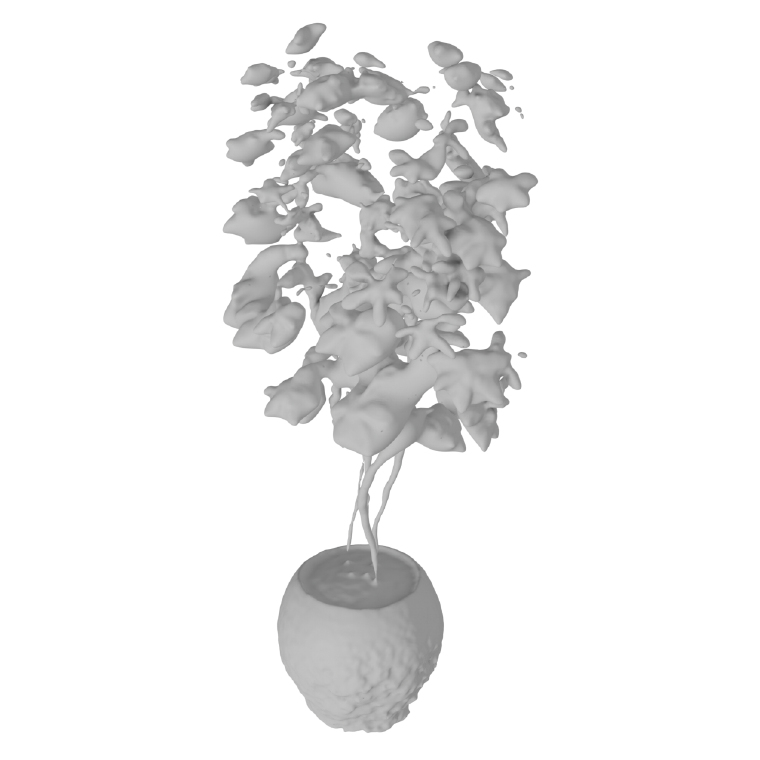} \\
    \rotatebox{90}{\footnotesize Spiderman} & \includegraphics[width=\excol]{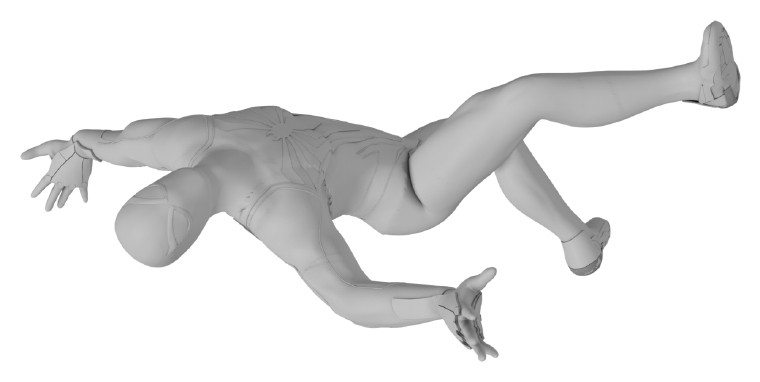} & \includegraphics[width=\excol]{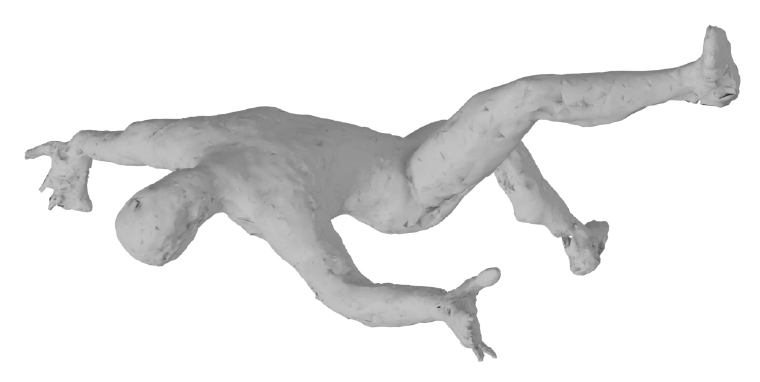} & \includegraphics[width=\excol]{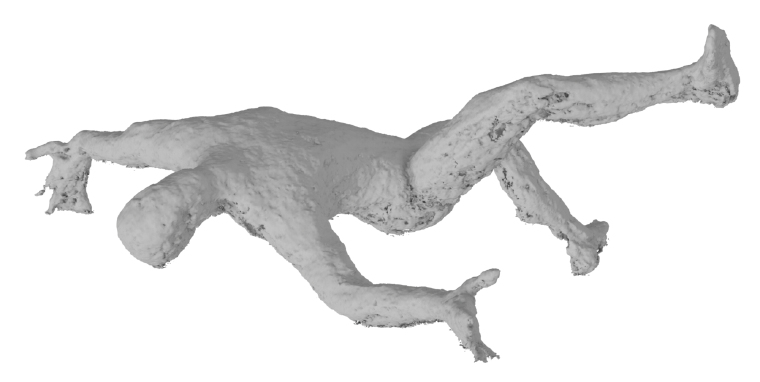} & \includegraphics[width=\excol]{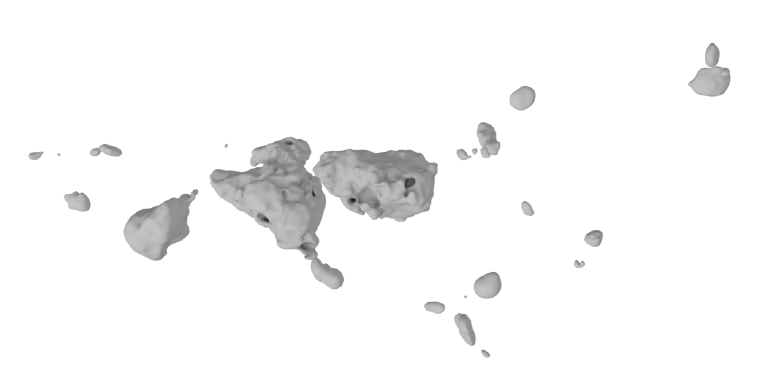} \\
  \end{tabular}
  \caption{\benchmark meshes of \ourmethod extracted by our marching tetrahedra and by the extraction routes of the baselines, TSDF fusion of the rendered depth and Poisson reconstruction from the Gaussian centers and normals, beside the ground truth.
    Marching tetrahedra preserves thin and detailed structures, TSDF fusion roughens and breaks them, and Poisson reconstruction melts them into smooth blobs or, on Spiderman, shatters the mesh.}
  \label{fig:ablations_extraction}
  \vspace{-2pt}
\end{figure*}

%% file: tables/real_datasets.tex
\begin{table}[t]
  \centering
  \footnotesize
  \setlength{\tabcolsep}{4pt}
  \begin{tabular}{@{}lccc@{}}
    \toprule
    \textbf{Scenes} & \textbf{Frames} & \textbf{Cams} & \textbf{Res.} \\
    \midrule
    \multicolumn{4}{@{}l}{\textbf{Neural 3D Video}} \\
    \cmidrule{1-4}
    Coffee martini & 300 & 17 & $1352\times1014$ \\
    Cut roasted beef & 300 & 19 & $1352\times1014$ \\
    Flame salmon & 300 & 18 & $1352\times1014$ \\
    Cook spinach & 300 & 20 & $1352\times1014$ \\
    Flame steak & 300 & 20 & $1352\times1014$ \\
    Sear steak & 300 & 20 & $1352\times1014$ \\
    \midrule
    \multicolumn{4}{@{}l}{\textbf{SelfCap}} \\
    \cmidrule{1-4}
    Hair1 & 60 & 23 & $1872\times1034$ \\
    Hair2 & 60 & 23 & $1872\times1034$ \\
    Corgi1 & 60 & 23 & $1879\times1042$ \\
    Corgi2 & 60 & 23 & $1879\times1042$ \\
    Bike1 & 60 & 21 & $1040\times1042$ \\
    Bike2 & 60 & 21 & $1040\times1042$ \\
    Yoga & 300 & 23 & $1876\times1038$ \\
    \bottomrule
  \end{tabular}
  \caption{The real scenes. Cams is the number of training cameras, besides the one held-out test camera of every scene, and Res.\ the image resolution in pixels.}
  \label{tab:real_datasets}
\end{table}

%% file: figures/benchmark.tex
\begin{figure*}[t]
  \centering
  \setlength{\tabcolsep}{2pt}
  \renewcommand{\arraystretch}{0.6}
  \newlength{\bmcol}\setlength{\bmcol}{\dimexpr(\linewidth-18\tabcolsep)/6\relax}
  \begin{tabular}{@{}cccccc@{}}
    \includegraphics[width=\bmcol]{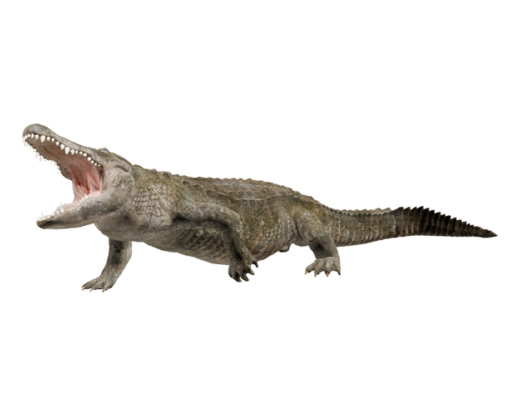} & \includegraphics[width=\bmcol]{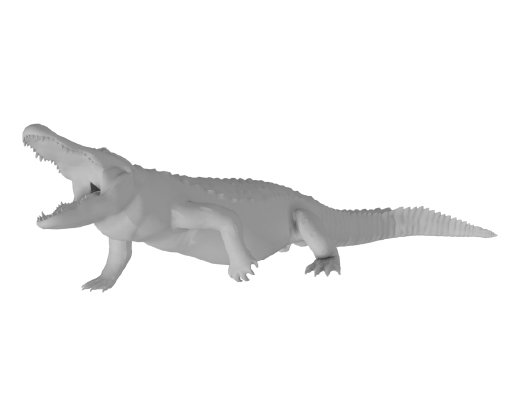} & \includegraphics[width=\bmcol]{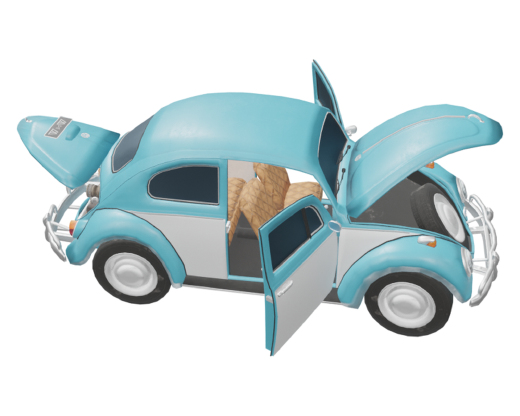} & \includegraphics[width=\bmcol]{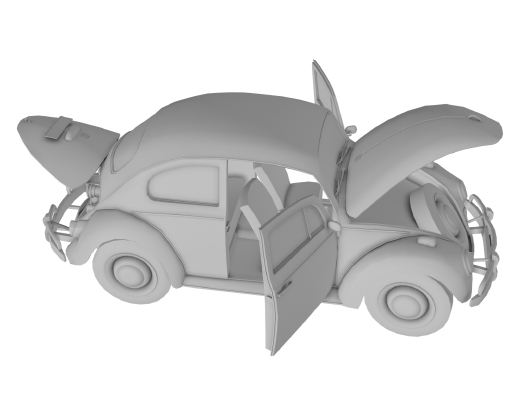} & \includegraphics[width=\bmcol]{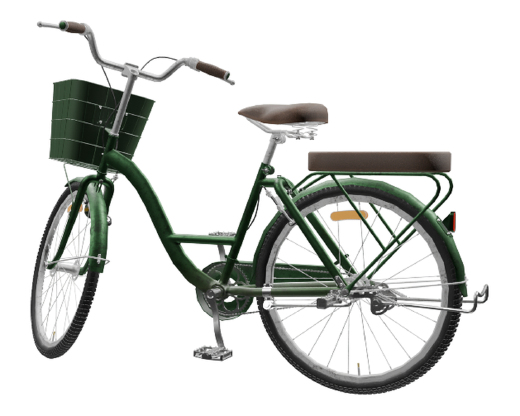} & \includegraphics[width=\bmcol]{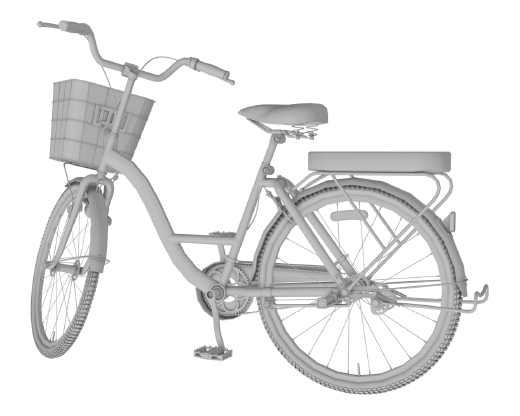} \\
    \multicolumn{2}{c}{\footnotesize Alligator, deforming body} & \multicolumn{2}{c}{\footnotesize Beetle, articulated doors and hood} & \multicolumn{2}{c}{\footnotesize Bicycle, dynamic thin structures} \\[4pt]
    \includegraphics[width=\bmcol]{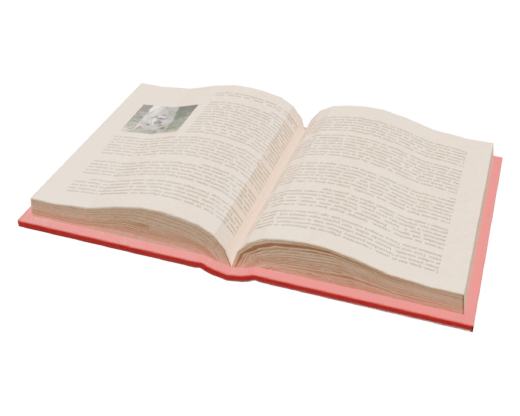} & \includegraphics[width=\bmcol]{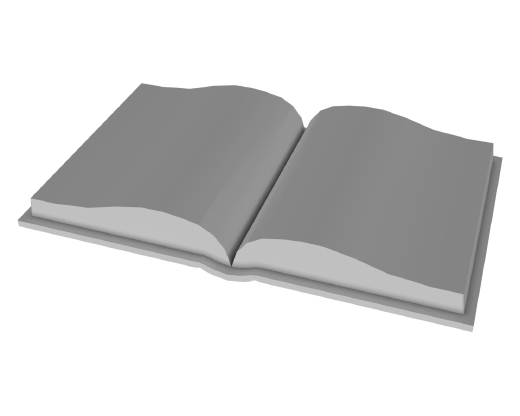} & \includegraphics[width=\bmcol]{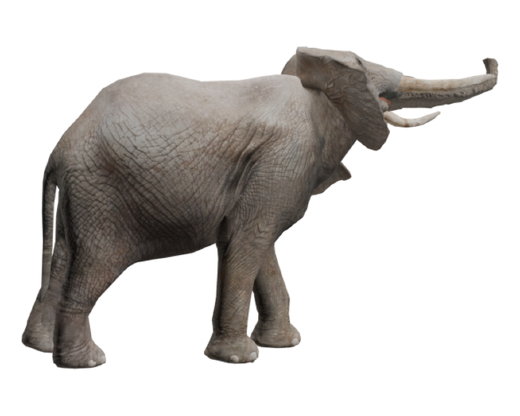} & \includegraphics[width=\bmcol]{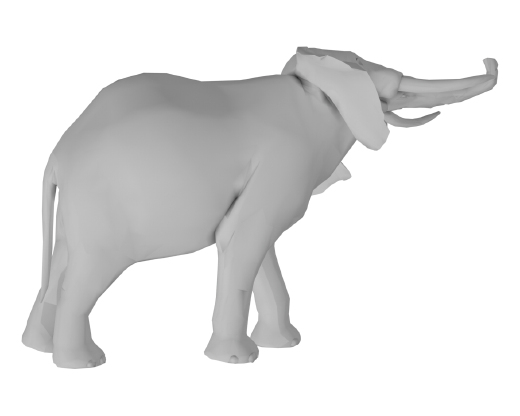} & \includegraphics[width=\bmcol]{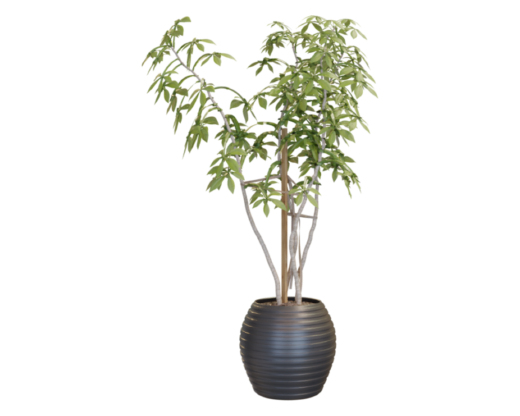} & \includegraphics[width=\bmcol]{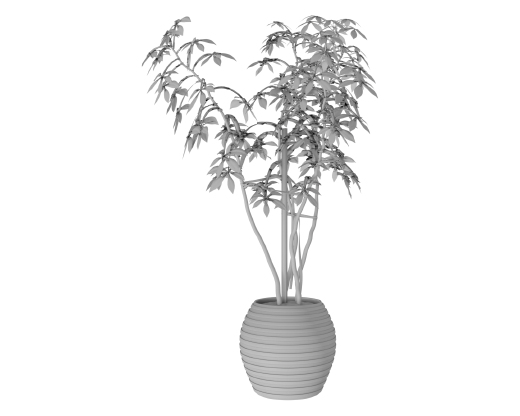} \\
    \multicolumn{2}{c}{\footnotesize Book, opening and closing} & \multicolumn{2}{c}{\footnotesize Elephant, deforming body} & \multicolumn{2}{c}{\footnotesize Ficus, swaying foliage} \\[4pt]
    \includegraphics[width=\bmcol]{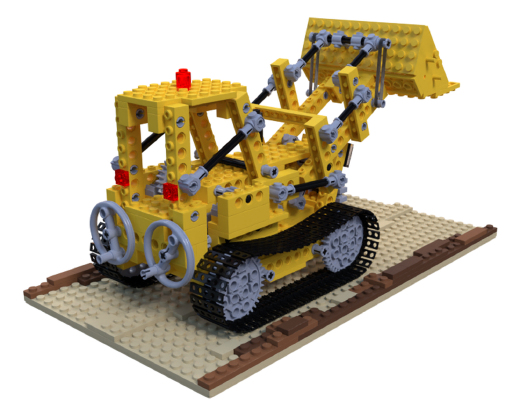} & \includegraphics[width=\bmcol]{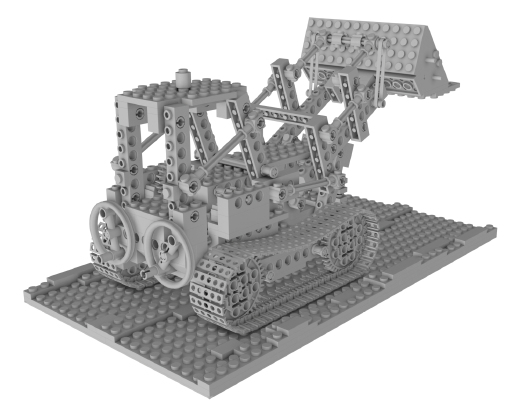} & \includegraphics[width=\bmcol]{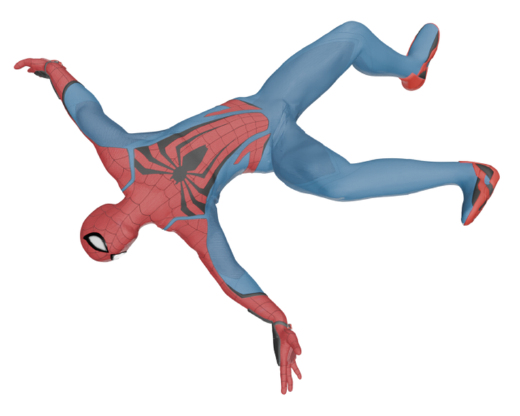} & \includegraphics[width=\bmcol]{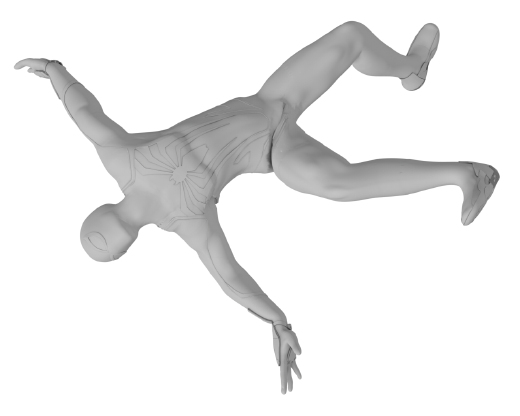} & \includegraphics[width=\bmcol]{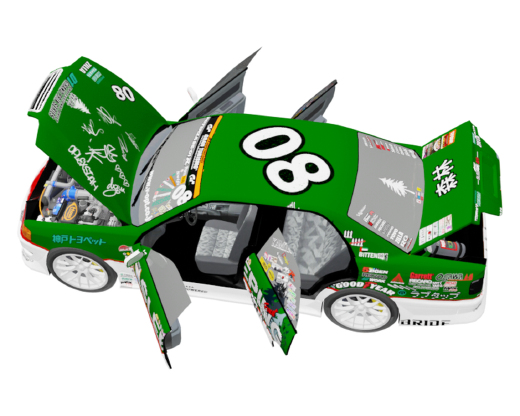} & \includegraphics[width=\bmcol]{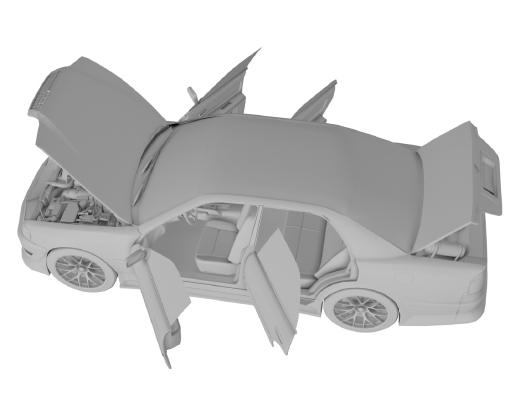} \\
    \multicolumn{2}{c}{\footnotesize Lego, articulated machinery} & \multicolumn{2}{c}{\footnotesize Spiderman, fast articulated character} & \multicolumn{2}{c}{\footnotesize Toyota, articulated doors and hood} \\[4pt]
  \end{tabular}
  \caption{The nine \benchmark scenes, each shown as a training image beside its ground-truth mesh at one frame.}
  \label{fig:benchmark}
\end{figure*}

%% file: tables/per_scene_benchmark_image.tex
\begin{table*}[tp]
  \centering
  \footnotesize
  \setlength{\tabcolsep}{3.5pt}
  \renewcommand{\arraystretch}{1.12}
  \begin{subtable}{\linewidth}
    \centering
    \begin{tabular}{@{}lcccccccccc@{}}
      \toprule
      Method & Alligator & Beetle & Bicycle & Book & Elephant & Ficus & Lego & Spiderman & Toyota & Mean \\
      \midrule
      4DGS~\cite{wu20244d} & 37.77 & \silver{35.30} & \bronze{30.99} & 35.66 & 36.76 & \bronze{30.73} & 31.84 & \bronze{29.08} & \bronze{31.13} & 33.25 \\
      Deformable-3DGS~\cite{yang2024deformable} & 38.68 & 34.25 & 30.45 & \silver{36.57} & 35.95 & 30.07 & \silver{32.69} & 24.67 & 30.01 & 32.59 \\
      SC-GS~\cite{huang2024scgs} & \silver{39.77} & 34.61 & 30.20 & 36.09 & \silver{37.88} & 30.56 & 32.02 & 25.06 & 30.03 & 32.91 \\
      FreeTimeGS~\cite{wang2025freetimegs} & 37.43 & 33.81 & \silver{32.00} & \bronze{36.32} & 37.21 & \gold{31.89} & 31.98 & \silver{31.42} & 29.51 & \silver{33.51} \\
      AST-3DGS~\cite{tzathas2026adaptive} & 34.30 & 33.45 & 29.19 & 29.06 & 32.66 & 27.21 & 31.01 & 20.77 & 26.69 & 29.37 \\
      D-2DGS~\cite{zhang2025dynamic} & \bronze{39.61} & \gold{35.54} & 30.58 & 35.77 & \bronze{37.49} & \silver{31.28} & \bronze{32.20} & 25.74 & \silver{32.26} & \bronze{33.38} \\
      DG-Mesh~\cite{liu2025dynamic} & 30.91 & 25.36 & 24.87 & 27.68 & 26.85 & 24.33 & 22.17 & 19.21 & 22.70 & 24.90 \\
      \midrule
      \ourmethod & \gold{41.09} & \bronze{35.15} & \gold{32.18} & \gold{37.82} & \gold{39.77} & 29.58 & \gold{32.83} & \gold{33.88} & \gold{32.36} & \gold{34.96} \\
      \bottomrule
    \end{tabular}
    \caption{PSNR$\uparrow$}
    \label{tab:per_scene_benchmark_image_psnr}
  \end{subtable}
  \par\vspace{0.6em}
  \begin{subtable}{\linewidth}
    \centering
    \begin{tabular}{@{}lcccccccccc@{}}
      \toprule
      Method & Alligator & Beetle & Bicycle & Book & Elephant & Ficus & Lego & Spiderman & Toyota & Mean \\
      \midrule
      4DGS~\cite{wu20244d} & 0.984 & 0.983 & 0.975 & 0.955 & 0.974 & 0.970 & 0.954 & \bronze{0.967} & 0.978 & 0.971 \\
      Deformable-3DGS~\cite{yang2024deformable} & \bronze{0.991} & 0.983 & \bronze{0.978} & \silver{0.971} & \bronze{0.983} & 0.969 & \silver{0.969} & 0.959 & \bronze{0.980} & \bronze{0.976} \\
      SC-GS~\cite{huang2024scgs} & \silver{0.992} & \bronze{0.984} & 0.977 & 0.968 & \silver{0.986} & \bronze{0.972} & \silver{0.969} & 0.961 & \bronze{0.980} & \silver{0.977} \\
      FreeTimeGS~\cite{wang2025freetimegs} & 0.986 & 0.983 & \bronze{0.978} & \bronze{0.969} & 0.979 & \gold{0.975} & 0.962 & \silver{0.976} & 0.974 & \bronze{0.976} \\
      AST-3DGS~\cite{tzathas2026adaptive} & 0.985 & 0.980 & 0.970 & 0.952 & 0.976 & 0.954 & 0.956 & 0.953 & 0.961 & 0.965 \\
      D-2DGS~\cite{zhang2025dynamic} & \silver{0.992} & \silver{0.986} & \silver{0.979} & 0.967 & \silver{0.986} & \silver{0.974} & \bronze{0.968} & 0.961 & \gold{0.985} & \silver{0.977} \\
      DG-Mesh~\cite{liu2025dynamic} & 0.976 & 0.953 & 0.958 & 0.934 & 0.960 & 0.939 & 0.922 & 0.940 & 0.921 & 0.945 \\
      \midrule
      \ourmethod & \gold{0.995} & \gold{0.987} & \gold{0.980} & \gold{0.978} & \gold{0.991} & 0.967 & \gold{0.972} & \gold{0.985} & \silver{0.984} & \gold{0.982} \\
      \bottomrule
    \end{tabular}
    \caption{SSIM$\uparrow$}
    \label{tab:per_scene_benchmark_image_ssim}
  \end{subtable}
  \par\vspace{0.6em}
  \begin{subtable}{\linewidth}
    \centering
    \begin{tabular}{@{}lcccccccccc@{}}
      \toprule
      Method & Alligator & Beetle & Bicycle & Book & Elephant & Ficus & Lego & Spiderman & Toyota & Mean \\
      \midrule
      4DGS~\cite{wu20244d} & 0.022 & 0.031 & 0.035 & 0.081 & 0.047 & \bronze{0.027} & 0.064 & \bronze{0.053} & 0.026 & 0.043 \\
      Deformable-3DGS~\cite{yang2024deformable} & \bronze{0.012} & \bronze{0.030} & \bronze{0.031} & \silver{0.059} & 0.028 & \bronze{0.027} & \bronze{0.034} & 0.065 & \bronze{0.021} & \bronze{0.034} \\
      SC-GS~\cite{huang2024scgs} & \bronze{0.012} & \silver{0.028} & 0.032 & 0.063 & \silver{0.025} & \silver{0.025} & \silver{0.032} & 0.062 & 0.022 & \silver{0.033} \\
      FreeTimeGS~\cite{wang2025freetimegs} & 0.023 & \bronze{0.030} & \bronze{0.031} & \bronze{0.060} & 0.039 & \silver{0.025} & 0.046 & \silver{0.043} & 0.031 & 0.036 \\
      AST-3DGS~\cite{tzathas2026adaptive} & 0.024 & 0.035 & 0.041 & 0.092 & 0.043 & 0.043 & 0.053 & 0.074 & 0.044 & 0.050 \\
      D-2DGS~\cite{zhang2025dynamic} & \silver{0.011} & \silver{0.028} & \silver{0.028} & 0.066 & \bronze{0.026} & \gold{0.024} & 0.035 & 0.062 & \silver{0.018} & \silver{0.033} \\
      DG-Mesh~\cite{liu2025dynamic} & 0.029 & 0.073 & 0.048 & 0.121 & 0.057 & 0.049 & 0.066 & 0.096 & 0.055 & 0.066 \\
      \midrule
      \ourmethod & \gold{0.008} & \gold{0.024} & \gold{0.025} & \gold{0.046} & \gold{0.016} & 0.030 & \gold{0.030} & \gold{0.027} & \gold{0.017} & \gold{0.025} \\
      \bottomrule
    \end{tabular}
    \caption{LPIPS$\downarrow$}
    \label{tab:per_scene_benchmark_image_lpips}
  \end{subtable}
  \caption{Per-scene image quality on \benchmark. Mean is the mean over the scenes. \swatch{rankgold}\,=\,best, \swatch{ranksilver}\,=\,second best, \swatch{rankbronze}\,=\,third best.}
  \label{tab:per_scene_benchmark_image}
  \vspace{-3pt}
\end{table*}

%% file: tables/per_scene_benchmark_mesh.tex
\begin{table*}[tp]
  \centering
  \footnotesize
  \setlength{\tabcolsep}{3.5pt}
  \renewcommand{\arraystretch}{1.12}
  \begin{subtable}{\linewidth}
    \centering
    \begin{tabular}{@{}lcccccccccc@{}}
      \toprule
      Method & Alligator & Beetle & Bicycle & Book & Elephant & Ficus & Lego & Spiderman & Toyota & Mean \\
      \midrule
      4DGS~\cite{wu20244d} & 0.65 & 0.96 & 0.78 & 1.37 & 0.58 & 2.49 & 0.76 & 1.67 & 1.02 & 1.14 \\
      Deformable-3DGS~\cite{yang2024deformable} & 0.69 & 1.09 & 0.78 & 1.21 & 0.62 & 2.73 & 0.57 & 2.45 & 0.98 & 1.24 \\
      SC-GS~\cite{huang2024scgs} & 0.58 & \bronze{0.93} & 0.76 & 1.33 & \bronze{0.53} & 2.66 & \bronze{0.56} & 2.06 & 1.04 & 1.16 \\
      FreeTimeGS~\cite{wang2025freetimegs} & 0.69 & 4.26 & 0.37 & \bronze{1.05} & 0.61 & 0.47 & \gold{0.41} & \silver{0.42} & 6.17 & 1.61 \\
      AST-3DGS~\cite{tzathas2026adaptive} & 1.11 & 1.04 & 1.27 & 4.67 & 0.71 & 2.68 & 0.66 & 3.33 & 1.35 & 1.87 \\
      D-2DGS~\cite{zhang2025dynamic} & \bronze{0.20} & \gold{0.67} & \silver{0.26} & \silver{0.88} & \silver{0.23} & \gold{0.26} & \silver{0.46} & \bronze{0.56} & \gold{0.66} & \gold{0.46} \\
      DG-Mesh~\cite{liu2025dynamic} & \silver{0.15} & \silver{0.78} & \bronze{0.33} & \gold{0.77} & \silver{0.23} & \silver{0.29} & 1.00 & 2.81 & \silver{0.91} & \bronze{0.81} \\
      \midrule
      \ourmethod & \gold{0.10} & 0.96 & \gold{0.24} & 1.07 & \gold{0.17} & \bronze{0.31} & 0.68 & \gold{0.13} & \bronze{0.95} & \silver{0.51} \\
      \bottomrule
    \end{tabular}
    \caption{Point Chamfer distance ($\times 100$)$\downarrow$}
    \label{tab:per_scene_benchmark_mesh_cd}
  \end{subtable}
  \par\vspace{0.6em}
  \begin{subtable}{\linewidth}
    \centering
    \begin{tabular}{@{}lcccccccccc@{}}
      \toprule
      Method & Alligator & Beetle & Bicycle & Book & Elephant & Ficus & Lego & Spiderman & Toyota & Mean \\
      \midrule
      4DGS~\cite{wu20244d} & 27.0 & 40.0 & 33.6 & 36.8 & 30.0 & 47.6 & 49.3 & \silver{36.1} & 41.6 & 38.0 \\
      Deformable-3DGS~\cite{yang2024deformable} & 24.6 & 40.5 & 33.3 & \bronze{35.6} & 29.7 & 46.8 & 48.8 & 40.8 & 41.3 & 37.9 \\
      SC-GS~\cite{huang2024scgs} & \bronze{21.4} & 38.3 & \bronze{32.6} & 37.0 & 25.3 & 46.4 & 48.6 & \bronze{38.9} & 39.4 & 36.4 \\
      FreeTimeGS~\cite{wang2025freetimegs} & 40.9 & 48.3 & 42.4 & 46.6 & 41.4 & 46.2 & 51.8 & 39.8 & 48.6 & 45.1 \\
      AST-3DGS~\cite{tzathas2026adaptive} & 29.4 & 38.6 & 38.3 & 40.9 & 27.0 & 47.7 & 49.2 & 47.4 & 42.6 & 40.1 \\
      D-2DGS~\cite{zhang2025dynamic} & 23.5 & \silver{33.1} & 33.0 & 36.9 & \bronze{23.9} & \bronze{36.1} & \bronze{47.4} & 42.5 & \bronze{35.2} & \bronze{34.6} \\
      DG-Mesh~\cite{liu2025dynamic} & \silver{15.2} & \bronze{34.4} & \silver{28.8} & \silver{33.4} & \silver{18.0} & \silver{32.9} & \gold{42.6} & 52.3 & \silver{32.6} & \silver{32.3} \\
      \midrule
      \ourmethod & \gold{15.0} & \gold{27.1} & \gold{26.9} & \gold{15.2} & \gold{13.1} & \gold{32.3} & \silver{43.1} & \gold{18.2} & \gold{30.9} & \gold{24.6} \\
      \bottomrule
    \end{tabular}
    \caption{Normal Chamfer distance in degrees$\downarrow$}
    \label{tab:per_scene_benchmark_mesh_nc}
  \end{subtable}
  \par\vspace{0.6em}
  \begin{subtable}{\linewidth}
    \centering
    \begin{tabular}{@{}lcccccccccc@{}}
      \toprule
      Method & Alligator & Beetle & Bicycle & Book & Elephant & Ficus & Lego & Spiderman & Toyota & Mean \\
      \midrule
      4DGS~\cite{wu20244d} & 0.463 & 0.240 & 0.374 & 0.240 & 0.407 & 0.149 & 0.131 & \silver{0.312} & 0.230 & 0.283 \\
      Deformable-3DGS~\cite{yang2024deformable} & 0.513 & 0.232 & 0.374 & \silver{0.280} & 0.404 & 0.163 & 0.141 & 0.226 & 0.234 & 0.285 \\
      SC-GS~\cite{huang2024scgs} & \bronze{0.600} & 0.272 & 0.395 & 0.243 & 0.515 & 0.175 & 0.147 & \bronze{0.273} & 0.269 & 0.321 \\
      FreeTimeGS~\cite{wang2025freetimegs} & 0.196 & 0.118 & 0.209 & 0.104 & 0.187 & 0.165 & 0.108 & 0.240 & 0.119 & 0.161 \\
      AST-3DGS~\cite{tzathas2026adaptive} & 0.438 & 0.268 & 0.288 & 0.180 & 0.481 & 0.153 & 0.130 & 0.157 & 0.212 & 0.256 \\
      D-2DGS~\cite{zhang2025dynamic} & 0.563 & \silver{0.382} & \bronze{0.400} & 0.266 & \bronze{0.570} & \bronze{0.357} & \bronze{0.195} & 0.211 & \bronze{0.341} & \bronze{0.365} \\
      DG-Mesh~\cite{liu2025dynamic} & \silver{0.773} & \bronze{0.331} & \silver{0.519} & \bronze{0.276} & \silver{0.688} & \silver{0.436} & \silver{0.225} & 0.071 & \silver{0.370} & \silver{0.410} \\
      \midrule
      \ourmethod & \gold{0.787} & \gold{0.485} & \gold{0.533} & \gold{0.566} & \gold{0.820} & \gold{0.444} & \gold{0.259} & \gold{0.719} & \gold{0.392} & \gold{0.556} \\
      \bottomrule
    \end{tabular}
    \caption{Joint F-score on surface samples$\uparrow$}
    \label{tab:per_scene_benchmark_mesh_fs}
  \end{subtable}
  \par\vspace{0.6em}
  \begin{subtable}{\linewidth}
    \centering
    \begin{tabular}{@{}lcccccccccc@{}}
      \toprule
      Method & Alligator & Beetle & Bicycle & Book & Elephant & Ficus & Lego & Spiderman & Toyota & Mean \\
      \midrule
      4DGS~\cite{wu20244d} & 0.467 & 0.375 & 0.436 & 0.297 & 0.451 & 0.156 & 0.271 & \silver{0.312} & 0.466 & 0.359 \\
      Deformable-3DGS~\cite{yang2024deformable} & 0.531 & 0.373 & 0.438 & \bronze{0.380} & 0.447 & 0.161 & 0.300 & 0.181 & 0.478 & 0.365 \\
      SC-GS~\cite{huang2024scgs} & 0.620 & 0.434 & 0.453 & 0.353 & 0.572 & 0.165 & 0.311 & 0.258 & 0.551 & 0.413 \\
      FreeTimeGS~\cite{wang2025freetimegs} & 0.224 & 0.228 & 0.241 & 0.172 & 0.254 & 0.202 & 0.204 & \bronze{0.309} & 0.306 & 0.238 \\
      AST-3DGS~\cite{tzathas2026adaptive} & 0.478 & 0.426 & 0.299 & 0.204 & 0.554 & 0.129 & 0.254 & 0.148 & 0.425 & 0.324 \\
      D-2DGS~\cite{zhang2025dynamic} & \bronze{0.664} & \silver{0.707} & \bronze{0.489} & \silver{0.462} & \bronze{0.724} & \bronze{0.395} & \silver{0.441} & 0.279 & \silver{0.709} & \silver{0.541} \\
      DG-Mesh~\cite{liu2025dynamic} & \silver{0.765} & \bronze{0.485} & \silver{0.537} & 0.292 & \silver{0.748} & \silver{0.420} & \bronze{0.356} & 0.084 & \bronze{0.607} & \bronze{0.477} \\
      \midrule
      \ourmethod & \gold{0.835} & \gold{0.837} & \gold{0.628} & \gold{0.877} & \gold{0.913} & \gold{0.490} & \gold{0.555} & \gold{0.802} & \gold{0.778} & \gold{0.746} \\
      \bottomrule
    \end{tabular}
    \caption{Joint F-score on rendered normals$\uparrow$}
    \label{tab:per_scene_benchmark_mesh_fr}
  \end{subtable}
  \caption{Per-scene mesh quality on \benchmark. Mean is the mean over the scenes. \swatch{rankgold}\,=\,best, \swatch{ranksilver}\,=\,second best, \swatch{rankbronze}\,=\,third best.}
  \label{tab:per_scene_benchmark_mesh}
  \vspace{-3pt}
\end{table*}

%% file: tables/per_scene_real.tex
\begin{table*}[tp]
  \centering
  \footnotesize
  \setlength{\tabcolsep}{2pt}
  \renewcommand{\arraystretch}{1.2}
  \begin{subtable}{\linewidth}
    \centering
    \begin{tabular}{@{}lccccccc@{\hspace{\dimexpr 2\tabcolsep+6pt\relax}}cccccccc@{}}
      \toprule
      & \multicolumn{7}{c}{Neural 3D Video} & \multicolumn{8}{c}{SelfCap} \\
      \cmidrule(r{10pt}){2-8}\cmidrule{9-16}
      Method & \shortstack{Coffee\\martini} & \shortstack{Cut\\roasted\\beef} & \shortstack{Flame\\salmon} & \shortstack{Cook\\spinach} & \shortstack{Flame\\steak} & \shortstack{Sear\\steak} & Mean & Hair1 & Hair2 & Corgi1 & Corgi2 & Bike1 & Bike2 & Yoga & Mean \\
      \cmidrule[\lightrulewidth](r{10pt}){1-8}\cmidrule[\lightrulewidth]{9-16}
      4DGS~\cite{wu20244d} & \silver{27.97} & \gold{33.35} & \gold{28.74} & \gold{32.49} & \silver{32.54} & \silver{32.69} & \gold{31.29} & \bronze{26.24} & 25.84 & \gold{27.91} & 27.20 & \bronze{25.85} & \gold{24.60} & \silver{29.84} & \silver{26.78} \\
      Deformable-3DGS~\cite{yang2024deformable} & \bronze{26.91} & 32.15 & 26.98 & \bronze{31.88} & 31.27 & \bronze{32.49} & 30.28 & \bronze{26.24} & \bronze{25.95} & \silver{27.60} & \bronze{27.28} & 25.68 & 23.39 & 29.41 & 26.51 \\
      SC-GS~\cite{huang2024scgs} & 24.83 & 27.41 & 24.84 & 28.08 & 27.91 & 28.08 & 26.86 & 25.11 & 24.71 & 25.93 & 25.96 & 25.09 & 19.71 & 24.02 & 24.36 \\
      FreeTimeGS~\cite{wang2025freetimegs} & 26.73 & \silver{32.49} & \bronze{27.27} & 31.78 & 31.43 & 29.87 & 29.93 & \gold{27.44} & \silver{26.82} & \bronze{27.39} & \gold{28.04} & 23.68 & \bronze{24.29} & \bronze{29.76} & \bronze{26.77} \\
      AST-3DGS~\cite{tzathas2026adaptive} & \gold{28.13} & 32.18 & \silver{28.04} & 31.85 & \bronze{32.15} & 32.43 & \silver{30.80} & 25.86 & 25.70 & 27.36 & 27.07 & 25.46 & 22.85 & 29.52 & 26.26 \\
      D-2DGS~\cite{zhang2025dynamic} & 24.02 & 26.79 & 24.58 & 26.76 & 26.33 & 27.47 & 25.99 & 15.27 & 25.06 & 26.78 & 26.40 & \silver{25.88} & 22.18 & 12.17 & 21.96 \\
      \cmidrule[\lightrulewidth](r{10pt}){1-8}\cmidrule[\lightrulewidth]{9-16}
      \ourmethod & 26.77 & \bronze{32.20} & 27.16 & \silver{32.30} & \gold{33.07} & \gold{32.95} & \bronze{30.74} & \silver{27.27} & \gold{27.08} & 27.17 & \silver{27.70} & \gold{26.03} & \silver{24.46} & \gold{30.30} & \gold{27.14} \\
      \bottomrule
    \end{tabular}
    \caption{PSNR$\uparrow$}
    \label{tab:per_scene_real_psnr}
  \end{subtable}
  \par\vspace{1.5em}
  \begin{subtable}{\linewidth}
    \centering
    \begin{tabular}{@{}lccccccc@{\hspace{\dimexpr 2\tabcolsep+6pt\relax}}cccccccc@{}}
      \toprule
      & \multicolumn{7}{c}{Neural 3D Video} & \multicolumn{8}{c}{SelfCap} \\
      \cmidrule(r{10pt}){2-8}\cmidrule{9-16}
      Method & \shortstack{Coffee\\martini} & \shortstack{Cut\\roasted\\beef} & \shortstack{Flame\\salmon} & \shortstack{Cook\\spinach} & \shortstack{Flame\\steak} & \shortstack{Sear\\steak} & Mean & Hair1 & Hair2 & Corgi1 & Corgi2 & Bike1 & Bike2 & Yoga & Mean \\
      \cmidrule[\lightrulewidth](r{10pt}){1-8}\cmidrule[\lightrulewidth]{9-16}
      4DGS~\cite{wu20244d} & \silver{0.910} & \silver{0.951} & \bronze{0.915} & \bronze{0.947} & \silver{0.955} & \bronze{0.955} & \bronze{0.939} & 0.903 & 0.900 & 0.816 & 0.814 & \bronze{0.905} & \bronze{0.886} & 0.947 & 0.881 \\
      Deformable-3DGS~\cite{yang2024deformable} & 0.906 & 0.949 & 0.909 & \silver{0.948} & \bronze{0.954} & 0.954 & 0.937 & \bronze{0.915} & \silver{0.914} & \bronze{0.867} & 0.857 & \silver{0.907} & 0.872 & \bronze{0.949} & \bronze{0.897} \\
      SC-GS~\cite{huang2024scgs} & 0.875 & 0.929 & 0.880 & 0.929 & 0.939 & 0.933 & 0.914 & 0.905 & \bronze{0.907} & 0.839 & 0.857 & 0.890 & 0.807 & 0.925 & 0.876 \\
      FreeTimeGS~\cite{wang2025freetimegs} & 0.896 & \bronze{0.950} & 0.898 & \bronze{0.947} & 0.953 & 0.948 & 0.932 & \silver{0.921} & \silver{0.914} & \silver{0.884} & \gold{0.882} & 0.891 & \silver{0.893} & \silver{0.950} & \silver{0.905} \\
      AST-3DGS~\cite{tzathas2026adaptive} & \gold{0.914} & \gold{0.952} & \silver{0.916} & \bronze{0.947} & 0.953 & \silver{0.956} & \silver{0.940} & 0.903 & 0.899 & 0.845 & 0.841 & 0.900 & 0.868 & 0.946 & 0.886 \\
      D-2DGS~\cite{zhang2025dynamic} & 0.857 & 0.920 & 0.868 & 0.914 & 0.923 & 0.916 & 0.900 & 0.713 & 0.897 & 0.855 & \bronze{0.859} & 0.904 & 0.871 & 0.654 & 0.822 \\
      \cmidrule[\lightrulewidth](r{10pt}){1-8}\cmidrule[\lightrulewidth]{9-16}
      \ourmethod & \bronze{0.909} & \gold{0.952} & \gold{0.917} & \gold{0.952} & \gold{0.959} & \gold{0.959} & \gold{0.941} & \gold{0.923} & \gold{0.922} & \gold{0.887} & \silver{0.880} & \gold{0.919} & \gold{0.905} & \gold{0.957} & \gold{0.913} \\
      \bottomrule
    \end{tabular}
    \caption{SSIM$\uparrow$}
    \label{tab:per_scene_real_ssim}
  \end{subtable}
  \par\vspace{1.5em}
  \begin{subtable}{\linewidth}
    \centering
    \begin{tabular}{@{}lccccccc@{\hspace{\dimexpr 2\tabcolsep+6pt\relax}}cccccccc@{}}
      \toprule
      & \multicolumn{7}{c}{Neural 3D Video} & \multicolumn{8}{c}{SelfCap} \\
      \cmidrule(r{10pt}){2-8}\cmidrule{9-16}
      Method & \shortstack{Coffee\\martini} & \shortstack{Cut\\roasted\\beef} & \shortstack{Flame\\salmon} & \shortstack{Cook\\spinach} & \shortstack{Flame\\steak} & \shortstack{Sear\\steak} & Mean & Hair1 & Hair2 & Corgi1 & Corgi2 & Bike1 & Bike2 & Yoga & Mean \\
      \cmidrule[\lightrulewidth](r{10pt}){1-8}\cmidrule[\lightrulewidth]{9-16}
      4DGS~\cite{wu20244d} & 0.227 & 0.228 & 0.218 & 0.228 & 0.213 & 0.218 & 0.222 & 0.224 & 0.219 & 0.320 & 0.334 & 0.231 & \bronze{0.239} & 0.218 & 0.255 \\
      Deformable-3DGS~\cite{yang2024deformable} & \bronze{0.225} & 0.227 & 0.215 & 0.223 & 0.214 & 0.214 & 0.220 & \silver{0.199} & \silver{0.199} & \bronze{0.197} & \bronze{0.226} & \silver{0.214} & 0.246 & \bronze{0.202} & \bronze{0.212} \\
      SC-GS~\cite{huang2024scgs} & 0.263 & 0.256 & 0.249 & 0.247 & 0.237 & 0.243 & 0.249 & 0.212 & \bronze{0.205} & 0.267 & 0.246 & 0.239 & 0.313 & 0.244 & 0.247 \\
      FreeTimeGS~\cite{wang2025freetimegs} & \silver{0.219} & \gold{0.206} & \bronze{0.213} & \gold{0.210} & \gold{0.200} & \gold{0.204} & \gold{0.209} & \gold{0.184} & \gold{0.187} & \gold{0.185} & \gold{0.195} & \bronze{0.218} & \gold{0.213} & \gold{0.184} & \gold{0.195} \\
      AST-3DGS~\cite{tzathas2026adaptive} & \gold{0.216} & \bronze{0.218} & \gold{0.209} & \bronze{0.220} & \bronze{0.212} & \bronze{0.210} & \bronze{0.214} & 0.222 & 0.225 & 0.270 & 0.278 & 0.231 & 0.260 & 0.219 & 0.244 \\
      D-2DGS~\cite{zhang2025dynamic} & 0.300 & 0.274 & 0.284 & 0.278 & 0.273 & 0.290 & 0.283 & 0.482 & 0.244 & 0.248 & 0.246 & 0.228 & 0.258 & 0.615 & 0.331 \\
      \cmidrule[\lightrulewidth](r{10pt}){1-8}\cmidrule[\lightrulewidth]{9-16}
      \ourmethod & \bronze{0.225} & \silver{0.212} & \silver{0.211} & \silver{0.215} & \silver{0.201} & \silver{0.206} & \silver{0.212} & \bronze{0.200} & \silver{0.199} & \silver{0.195} & \silver{0.204} & \gold{0.202} & \silver{0.214} & \silver{0.194} & \silver{0.201} \\
      \bottomrule
    \end{tabular}
    \caption{LPIPS$\downarrow$}
    \label{tab:per_scene_real_lpips}
  \end{subtable}
  \caption{Per-scene image quality on Neural 3D Video and SelfCap. Mean is the mean over the scenes. \swatch{rankgold}\,=\,best, \swatch{ranksilver}\,=\,second best, \swatch{rankbronze}\,=\,third best.}
  \label{tab:per_scene_real}
\end{table*}

%% file: figures/mesh_comparison_benchmark.tex
\begin{figure*}[p]
  \centering
  \setlength{\tabcolsep}{1pt}
  \renewcommand{\arraystretch}{0.6}
  \newlength{\mcrow}\setlength{\mcrow}{\dimexpr(\linewidth-1.2em-12\tabcolsep)*1000/7056\relax}
  \ifdim\mcrow>0.95in\setlength{\mcrow}{0.95in}\fi
  \begin{tabular}{@{}m{1.2em}cccccc@{}}
    & \multicolumn{1}{c}{\footnotesize Book} & \multicolumn{1}{c}{\footnotesize Elephant} & \multicolumn{1}{c}{\footnotesize Lego} & \multicolumn{1}{c}{\footnotesize Ficus} & \multicolumn{1}{c}{\footnotesize Bicycle} & \multicolumn{1}{c}{\footnotesize Beetle} \\
    \rotatebox{90}{\footnotesize 4DGS} & \raisebox{-0.5\height}{\includegraphics[height=\mcrow]{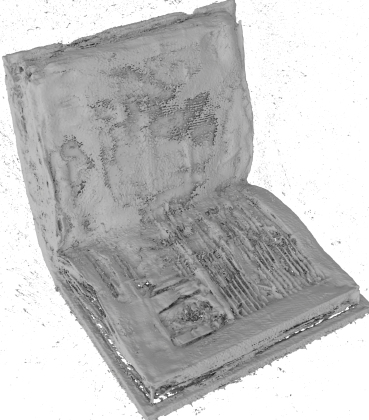}} & \raisebox{-0.5\height}{\includegraphics[height=\mcrow]{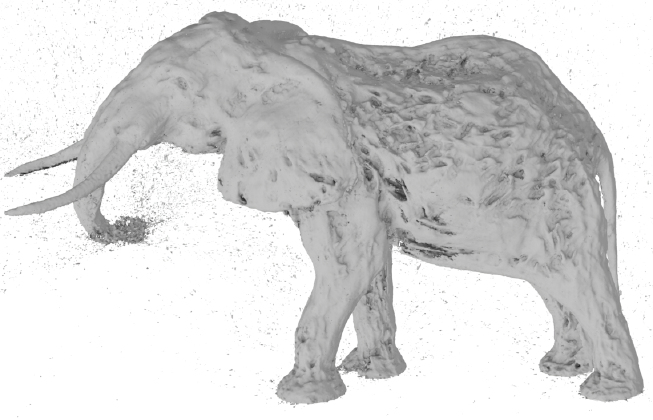}} & \raisebox{-0.5\height}{\includegraphics[height=\mcrow]{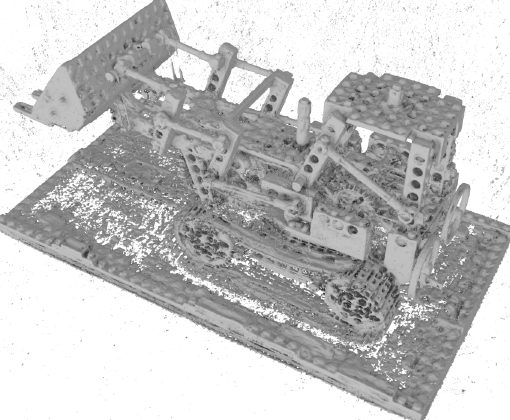}} & \raisebox{-0.5\height}{\includegraphics[height=\mcrow]{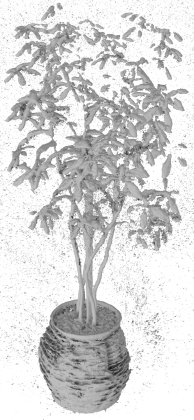}} & \raisebox{-0.5\height}{\includegraphics[height=\mcrow]{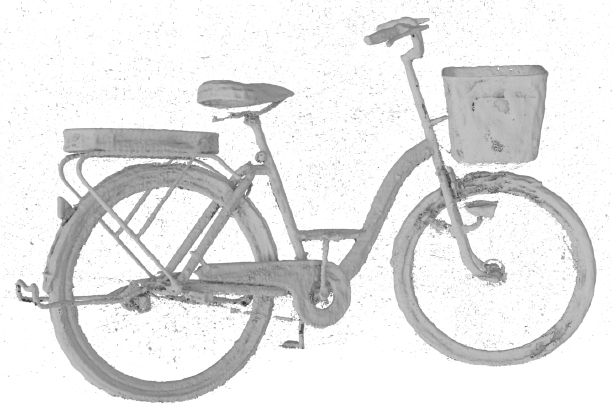}} & \raisebox{-0.5\height}{\includegraphics[height=\mcrow]{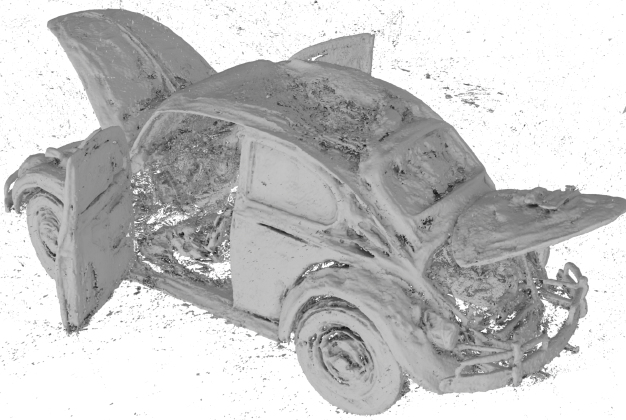}} \\
    \rotatebox{90}{\footnotesize Deformable-3DGS} & \raisebox{-0.5\height}{\includegraphics[height=\mcrow]{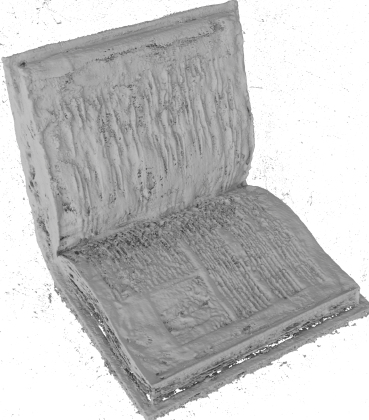}} & \raisebox{-0.5\height}{\includegraphics[height=\mcrow]{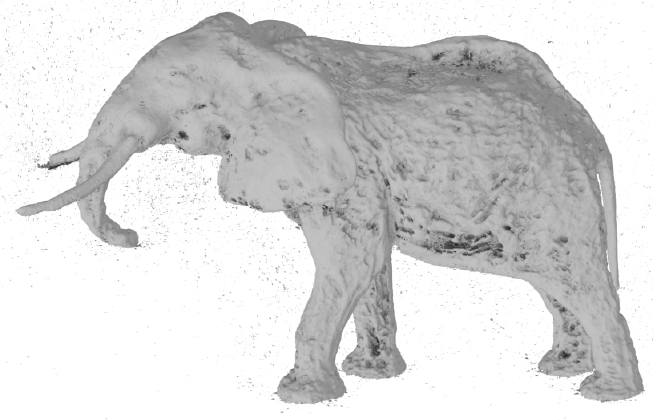}} & \raisebox{-0.5\height}{\includegraphics[height=\mcrow]{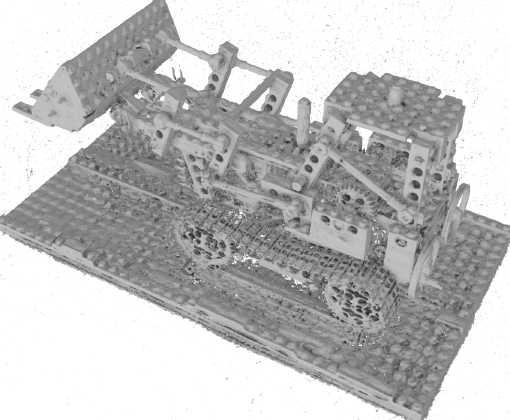}} & \raisebox{-0.5\height}{\includegraphics[height=\mcrow]{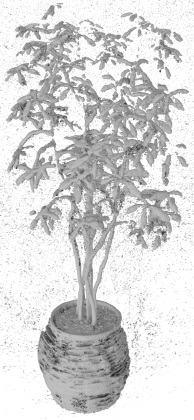}} & \raisebox{-0.5\height}{\includegraphics[height=\mcrow]{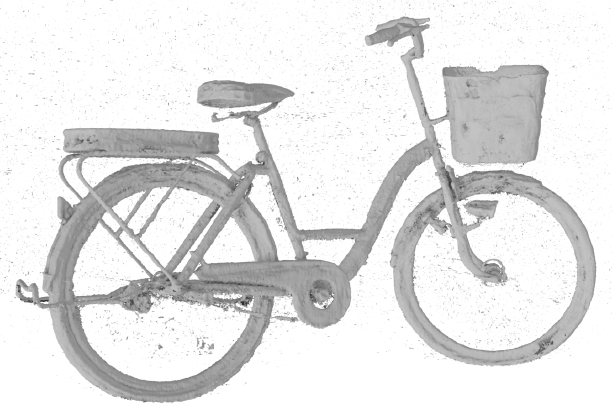}} & \raisebox{-0.5\height}{\includegraphics[height=\mcrow]{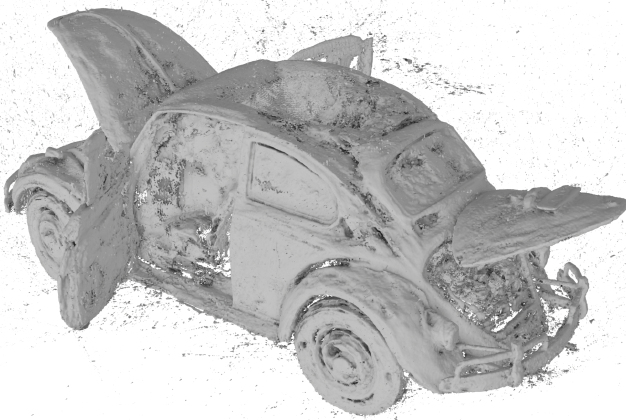}} \\
    \rotatebox{90}{\footnotesize SC-GS} & \raisebox{-0.5\height}{\includegraphics[height=\mcrow]{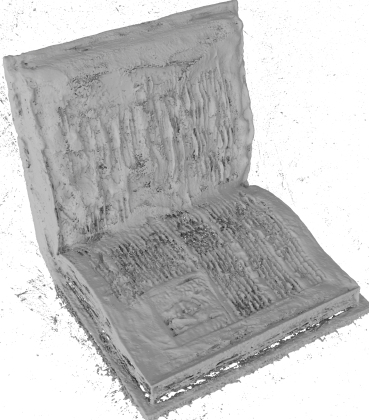}} & \raisebox{-0.5\height}{\includegraphics[height=\mcrow]{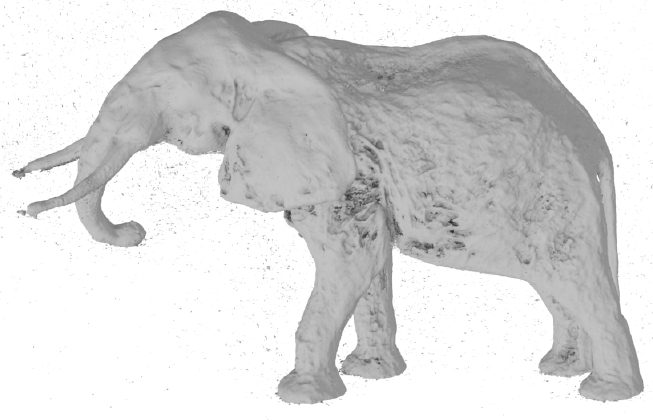}} & \raisebox{-0.5\height}{\includegraphics[height=\mcrow]{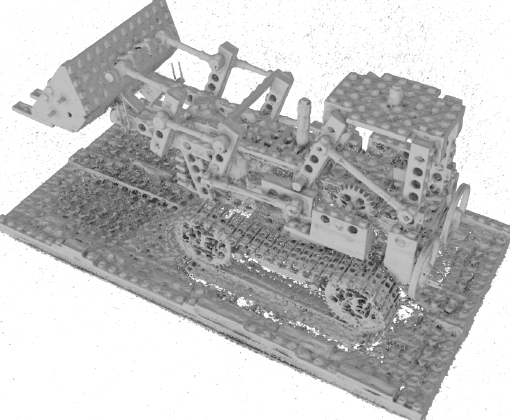}} & \raisebox{-0.5\height}{\includegraphics[height=\mcrow]{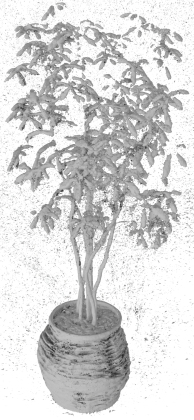}} & \raisebox{-0.5\height}{\includegraphics[height=\mcrow]{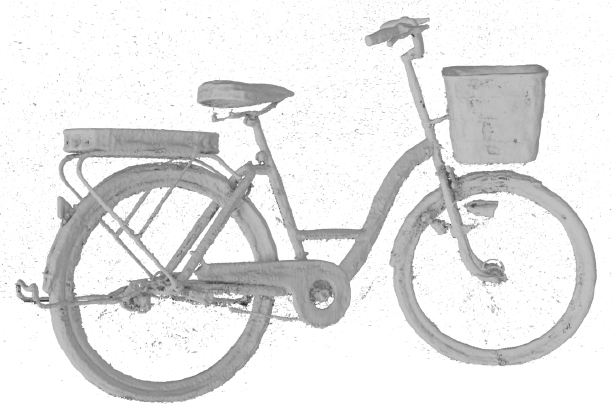}} & \raisebox{-0.5\height}{\includegraphics[height=\mcrow]{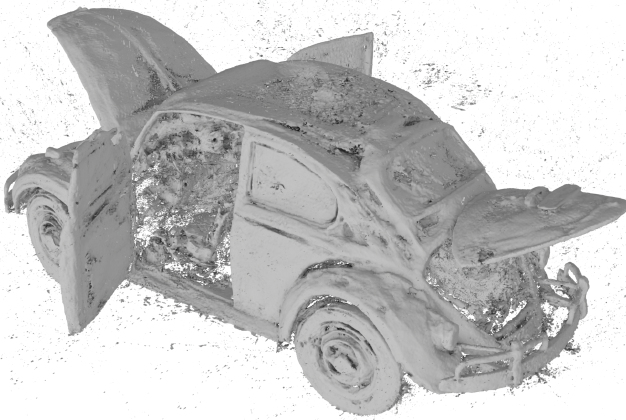}} \\
    \rotatebox{90}{\footnotesize FreeTimeGS} & \raisebox{-0.5\height}{\includegraphics[height=\mcrow]{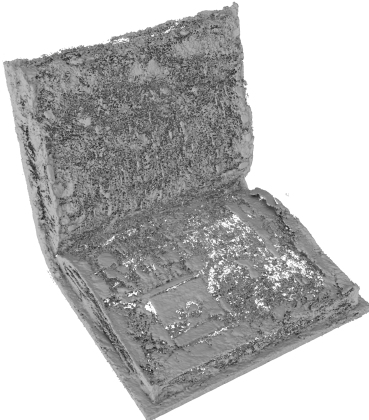}} & \raisebox{-0.5\height}{\includegraphics[height=\mcrow]{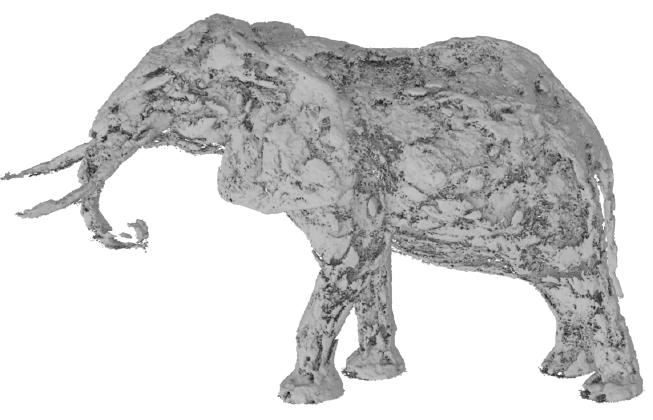}} & \raisebox{-0.5\height}{\includegraphics[height=\mcrow]{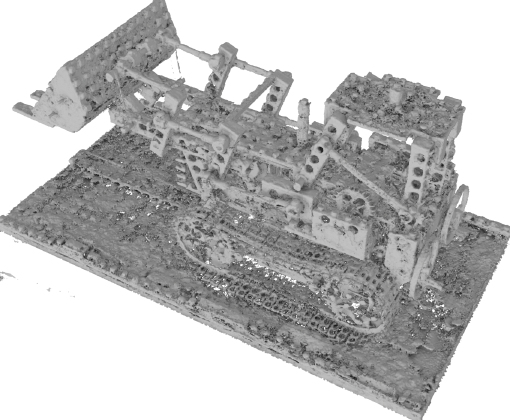}} & \raisebox{-0.5\height}{\includegraphics[height=\mcrow]{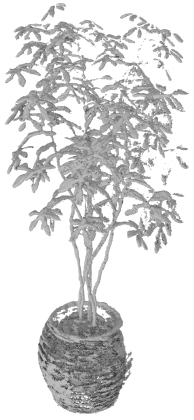}} & \raisebox{-0.5\height}{\includegraphics[height=\mcrow]{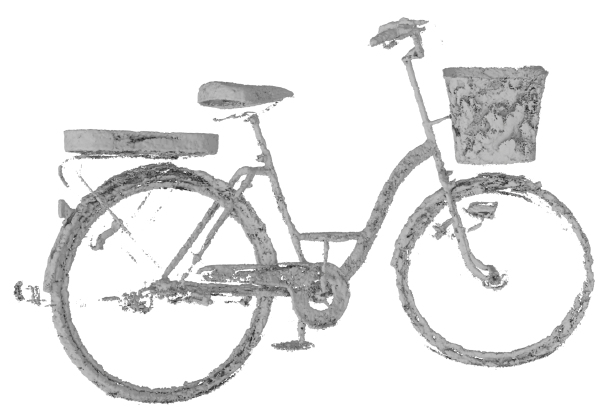}} & \raisebox{-0.5\height}{\includegraphics[height=\mcrow]{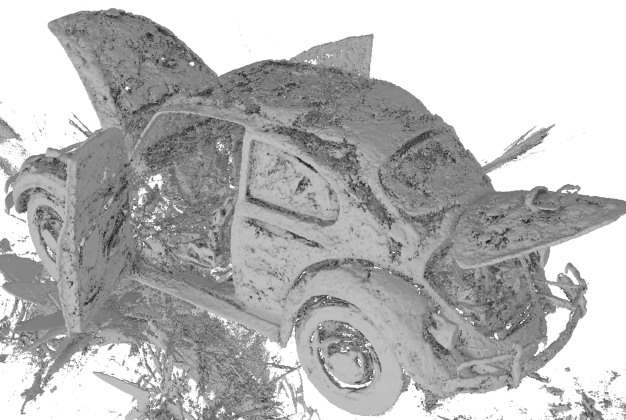}} \\
    \rotatebox{90}{\footnotesize AST-3DGS} & \raisebox{-0.5\height}{\includegraphics[height=\mcrow]{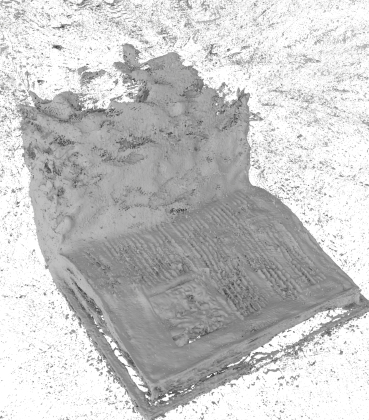}} & \raisebox{-0.5\height}{\includegraphics[height=\mcrow]{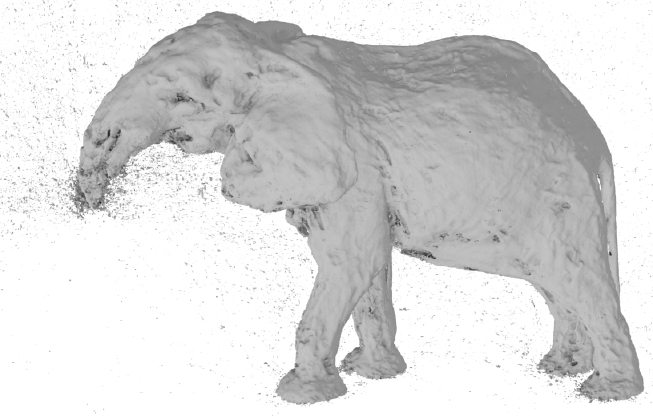}} & \raisebox{-0.5\height}{\includegraphics[height=\mcrow]{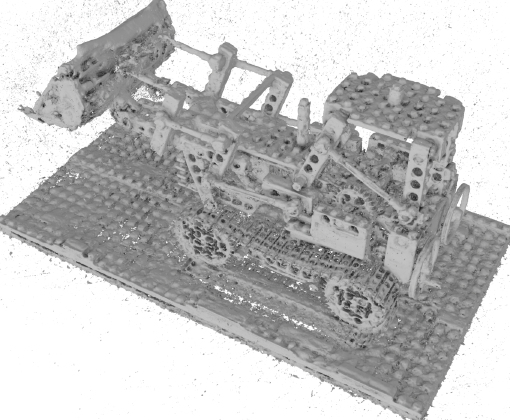}} & \raisebox{-0.5\height}{\includegraphics[height=\mcrow]{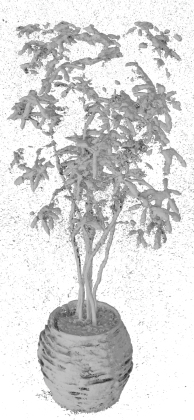}} & \raisebox{-0.5\height}{\includegraphics[height=\mcrow]{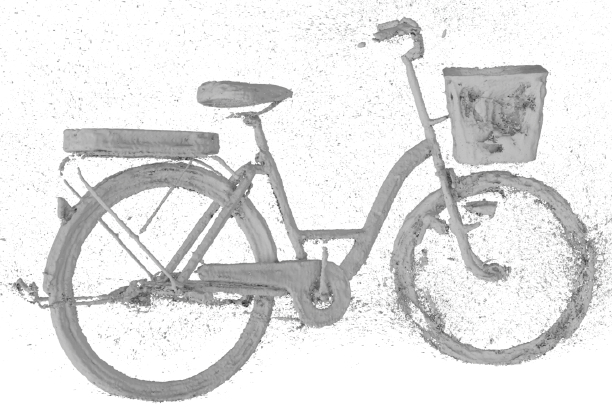}} & \raisebox{-0.5\height}{\includegraphics[height=\mcrow]{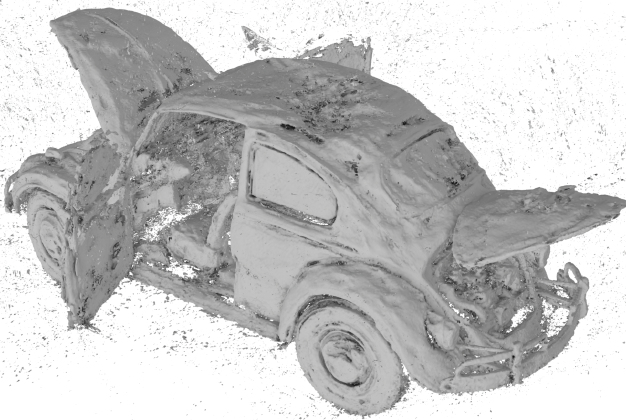}} \\
    \rotatebox{90}{\footnotesize D-2DGS} & \raisebox{-0.5\height}{\includegraphics[height=\mcrow]{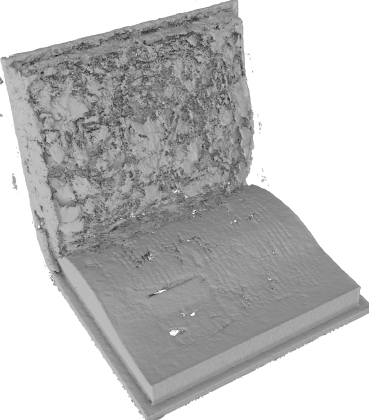}} & \raisebox{-0.5\height}{\includegraphics[height=\mcrow]{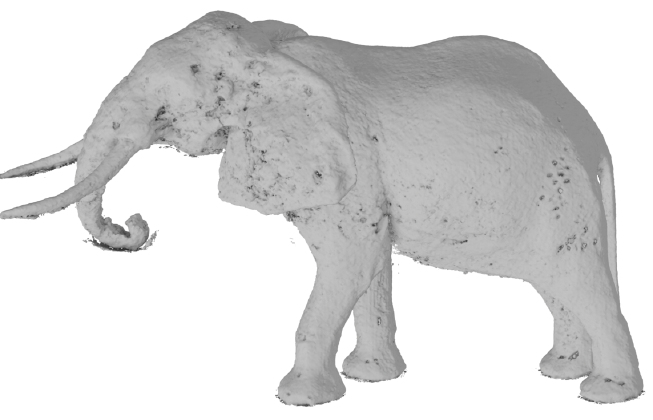}} & \raisebox{-0.5\height}{\includegraphics[height=\mcrow]{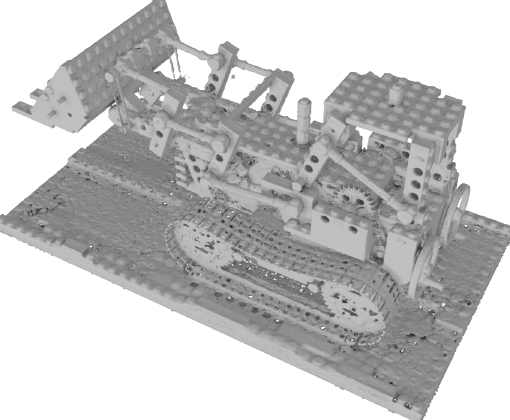}} & \raisebox{-0.5\height}{\includegraphics[height=\mcrow]{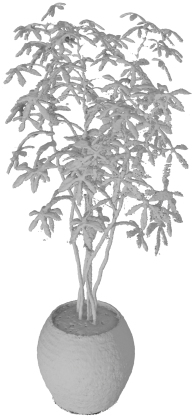}} & \raisebox{-0.5\height}{\includegraphics[height=\mcrow]{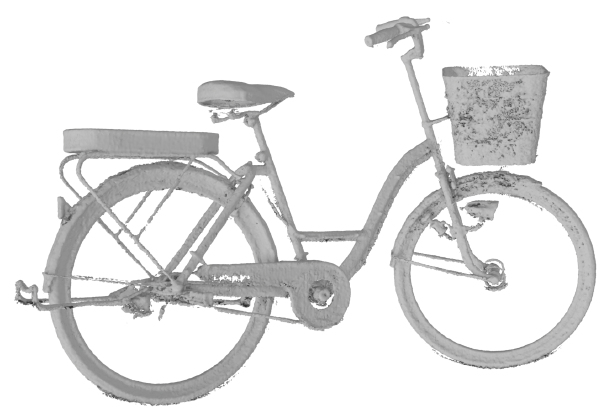}} & \raisebox{-0.5\height}{\includegraphics[height=\mcrow]{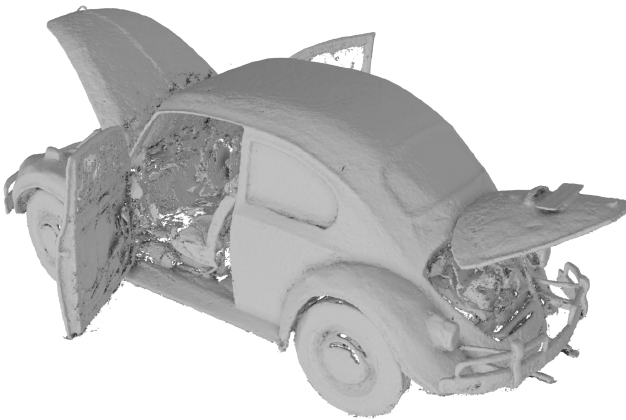}} \\
    \rotatebox{90}{\footnotesize DG-Mesh} & \raisebox{-0.5\height}{\includegraphics[height=\mcrow]{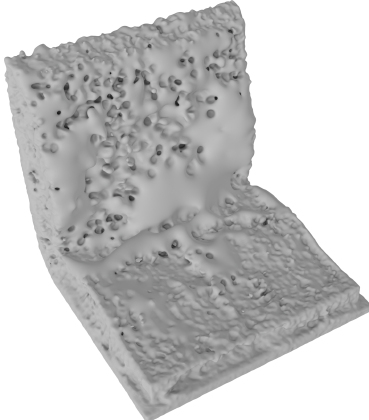}} & \raisebox{-0.5\height}{\includegraphics[height=\mcrow]{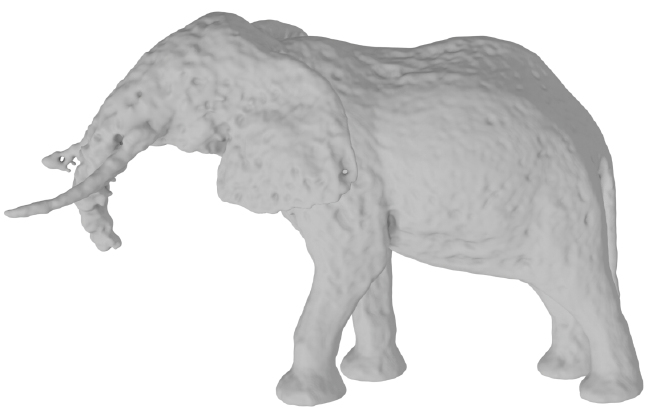}} & \raisebox{-0.5\height}{\includegraphics[height=\mcrow]{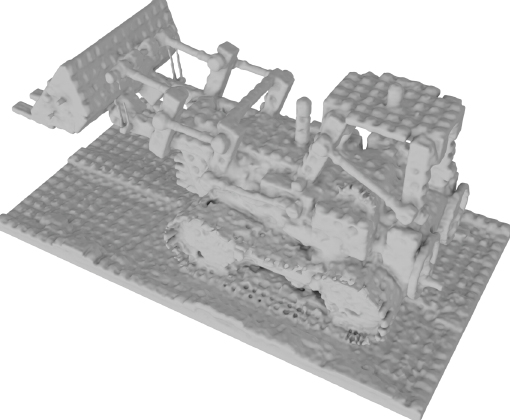}} & \raisebox{-0.5\height}{\includegraphics[height=\mcrow]{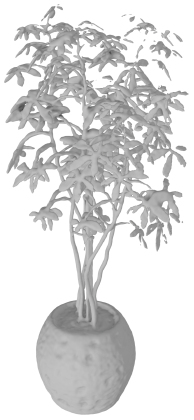}} & \raisebox{-0.5\height}{\includegraphics[height=\mcrow]{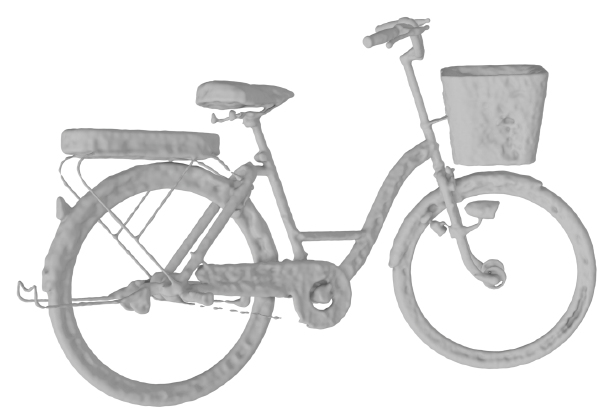}} & \raisebox{-0.5\height}{\includegraphics[height=\mcrow]{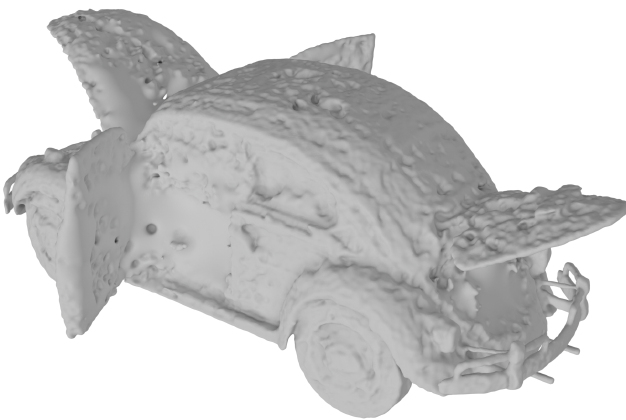}} \\
    \rotatebox{90}{\footnotesize \ourmethod} & \raisebox{-0.5\height}{\includegraphics[height=\mcrow]{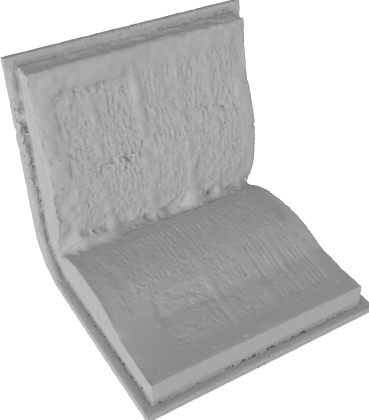}} & \raisebox{-0.5\height}{\includegraphics[height=\mcrow]{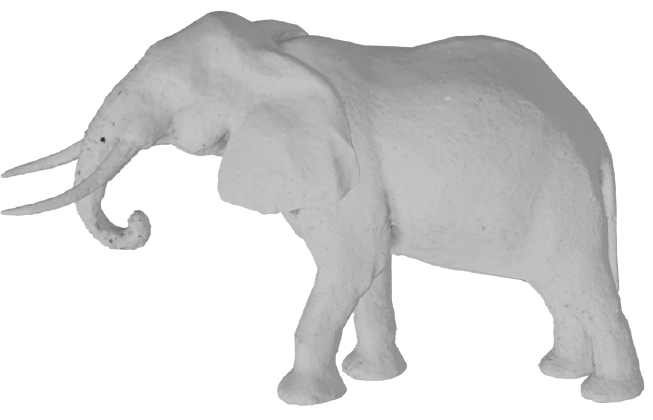}} & \raisebox{-0.5\height}{\includegraphics[height=\mcrow]{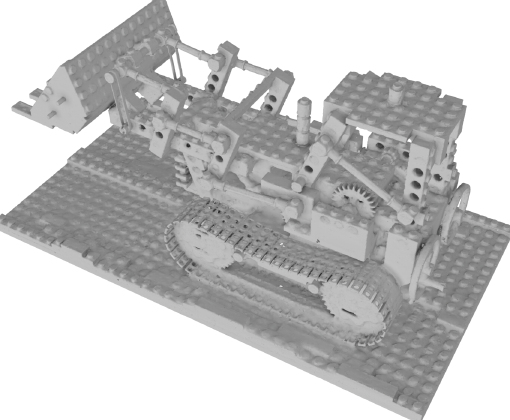}} & \raisebox{-0.5\height}{\includegraphics[height=\mcrow]{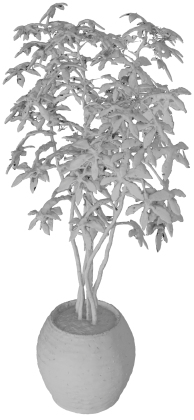}} & \raisebox{-0.5\height}{\includegraphics[height=\mcrow]{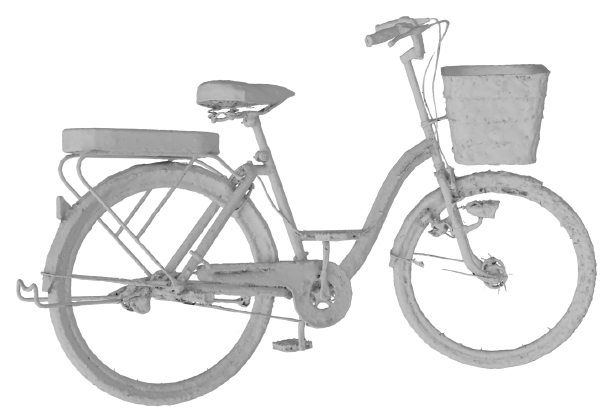}} & \raisebox{-0.5\height}{\includegraphics[height=\mcrow]{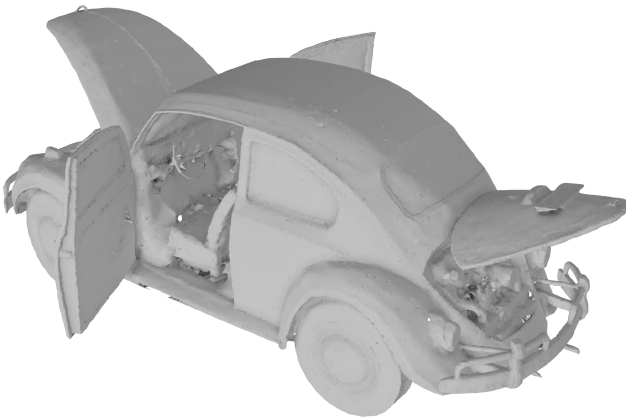}} \\
    \rotatebox{90}{\footnotesize Ground truth} & \raisebox{-0.5\height}{\includegraphics[height=\mcrow]{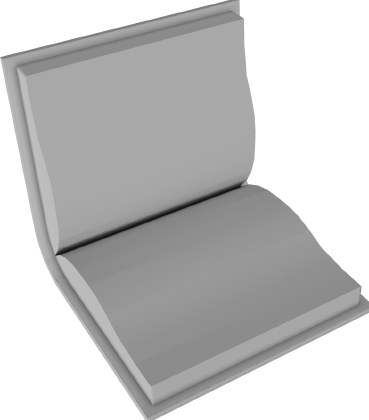}} & \raisebox{-0.5\height}{\includegraphics[height=\mcrow]{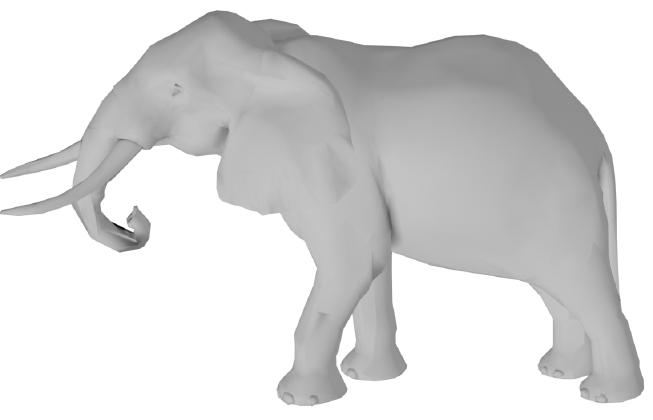}} & \raisebox{-0.5\height}{\includegraphics[height=\mcrow]{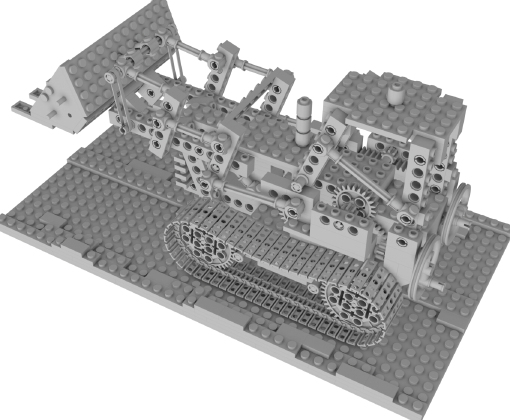}} & \raisebox{-0.5\height}{\includegraphics[height=\mcrow]{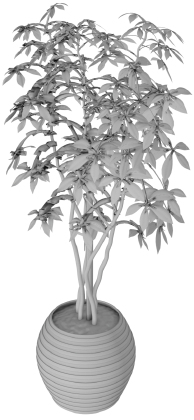}} & \raisebox{-0.5\height}{\includegraphics[height=\mcrow]{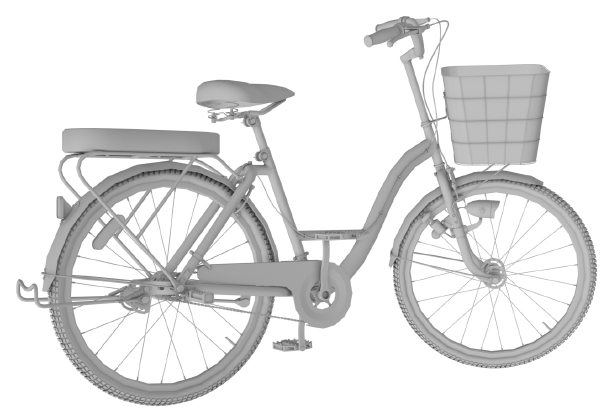}} & \raisebox{-0.5\height}{\includegraphics[height=\mcrow]{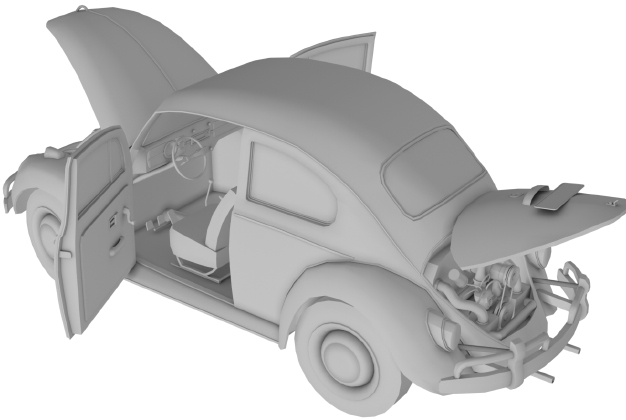}} \\
  \end{tabular}
  \caption{\benchmark meshes extracted by all methods. The gray spots in the background of some methods are Gaussian floaters. Please zoom in for details.}
  \label{fig:mesh_comparison_benchmark}
\end{figure*}

%% file: figures/mesh_comparison_real.tex
\begin{figure*}[p]
  \centering
  \setlength{\tabcolsep}{1pt}
  \renewcommand{\arraystretch}{0.6}
  \newlength{\mcrealrow}\setlength{\mcrealrow}{\dimexpr(\linewidth-1.2em-10\tabcolsep)*1000/7752\relax}
  \ifdim\mcrealrow>0.95in\setlength{\mcrealrow}{0.95in}\fi
  \begin{tabular}{@{}m{1.2em}ccccc@{}}
    & \multicolumn{1}{c}{\footnotesize Coffee martini} & \multicolumn{1}{c}{\footnotesize Yoga} & \multicolumn{1}{c}{\footnotesize Bike2} & \multicolumn{1}{c}{\footnotesize Corgi2} & \multicolumn{1}{c}{\footnotesize Hair1} \\
    \rotatebox{90}{\makebox[\mcrealrow]{\footnotesize 4DGS}} & \raisebox{-0.5\height}{\includegraphics[height=\mcrealrow]{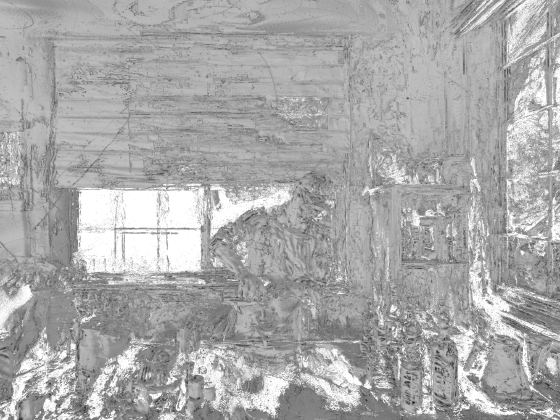}} & \raisebox{-0.5\height}{\includegraphics[height=\mcrealrow]{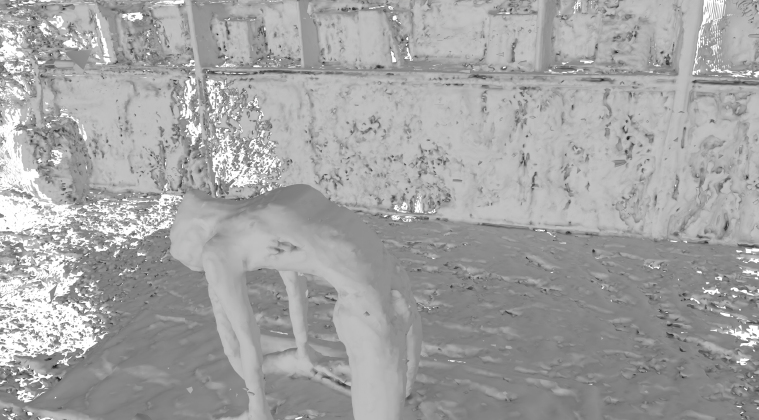}} & \raisebox{-0.5\height}{\includegraphics[height=\mcrealrow]{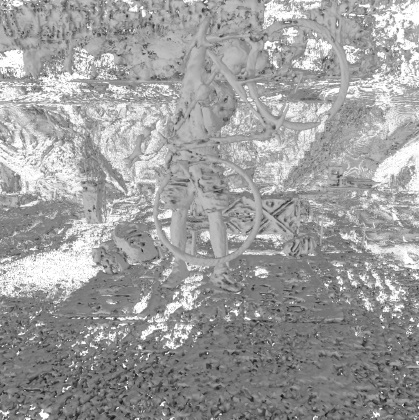}} & \raisebox{-0.5\height}{\includegraphics[height=\mcrealrow]{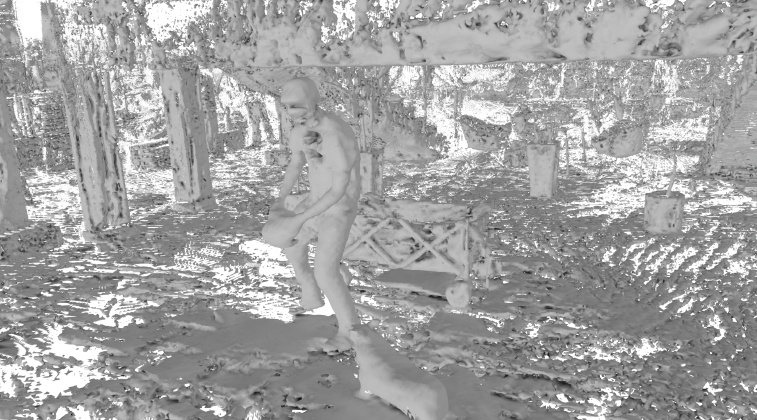}} & \raisebox{-0.5\height}{\includegraphics[height=\mcrealrow]{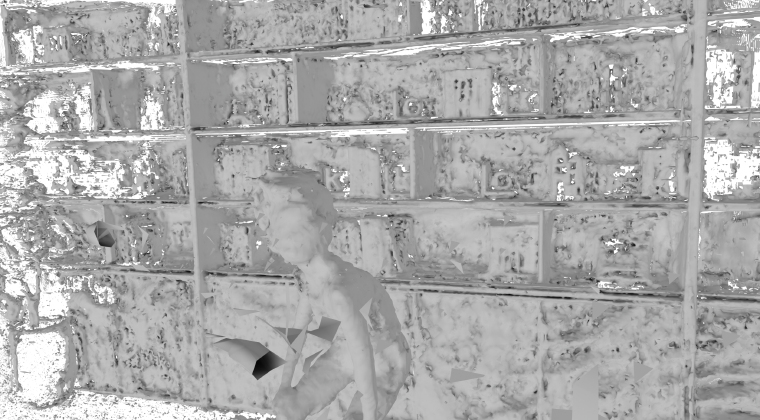}} \\[2pt]
    \rotatebox{90}{\makebox[\mcrealrow]{\footnotesize Deformable-3DGS}} & \raisebox{-0.5\height}{\includegraphics[height=\mcrealrow]{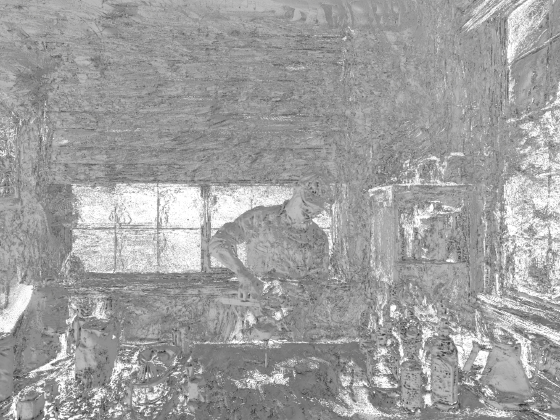}} & \raisebox{-0.5\height}{\includegraphics[height=\mcrealrow]{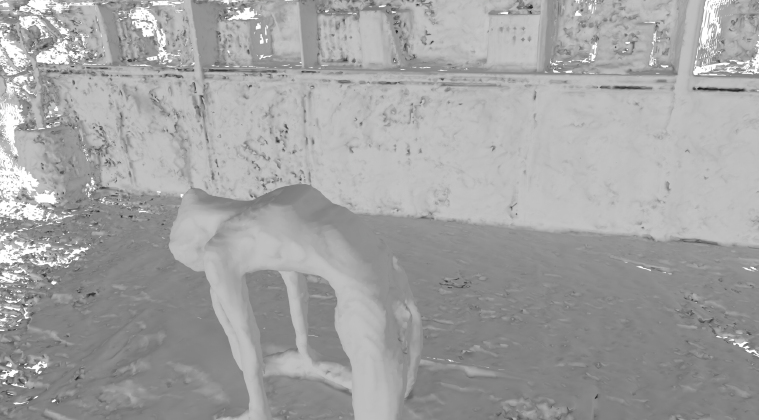}} & \raisebox{-0.5\height}{\includegraphics[height=\mcrealrow]{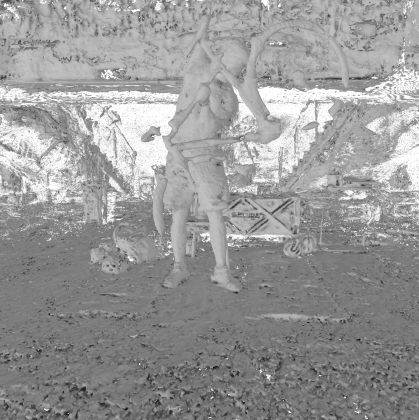}} & \raisebox{-0.5\height}{\includegraphics[height=\mcrealrow]{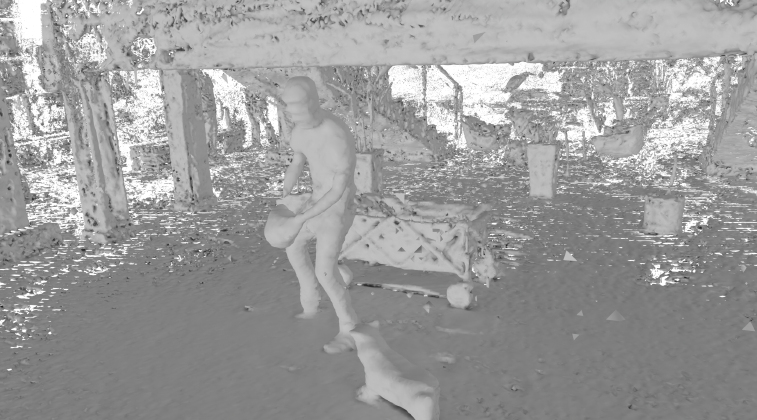}} & \raisebox{-0.5\height}{\includegraphics[height=\mcrealrow]{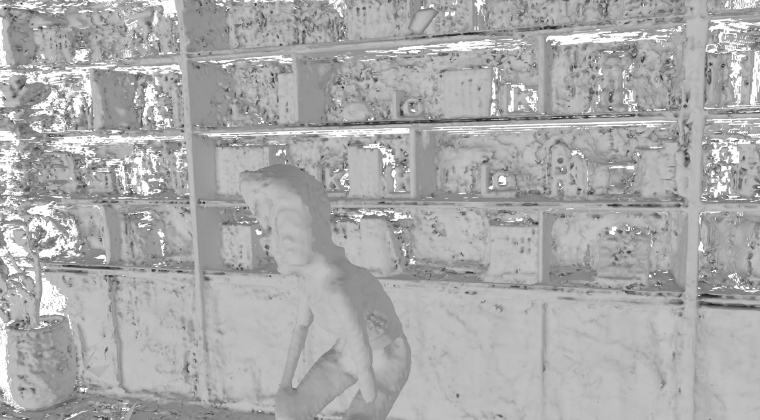}} \\[2pt]
    \rotatebox{90}{\makebox[\mcrealrow]{\footnotesize SC-GS}} & \raisebox{-0.5\height}{\includegraphics[height=\mcrealrow]{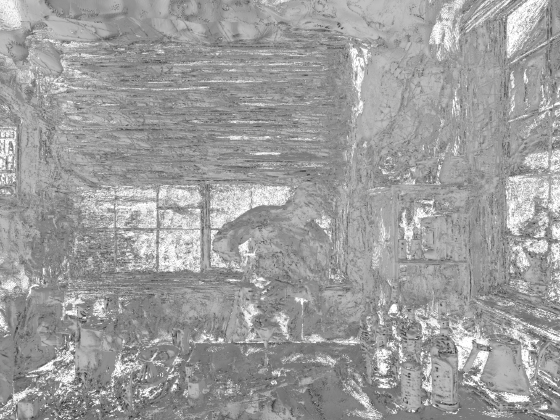}} & \raisebox{-0.5\height}{\includegraphics[height=\mcrealrow]{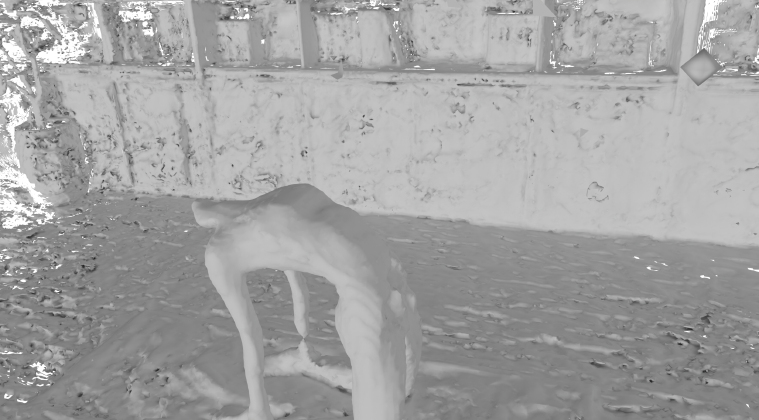}} & \raisebox{-0.5\height}{\includegraphics[height=\mcrealrow]{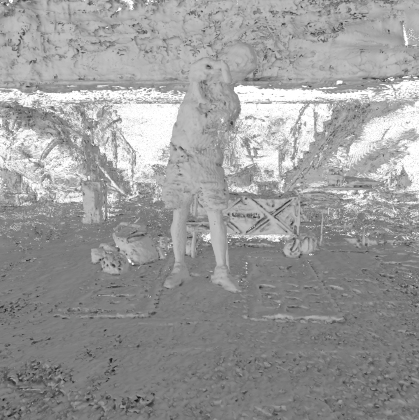}} & \raisebox{-0.5\height}{\includegraphics[height=\mcrealrow]{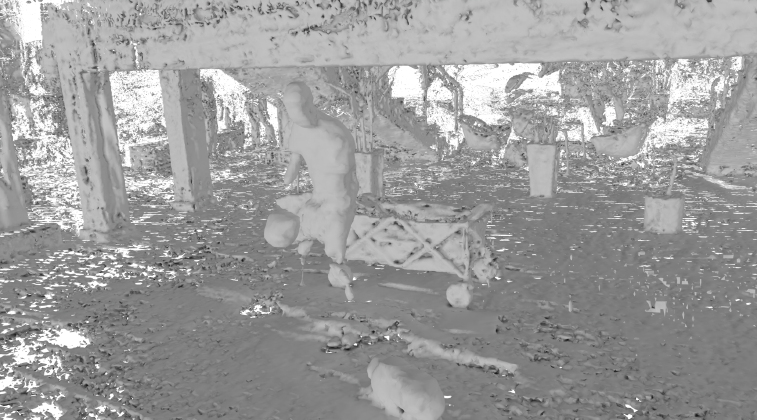}} & \raisebox{-0.5\height}{\includegraphics[height=\mcrealrow]{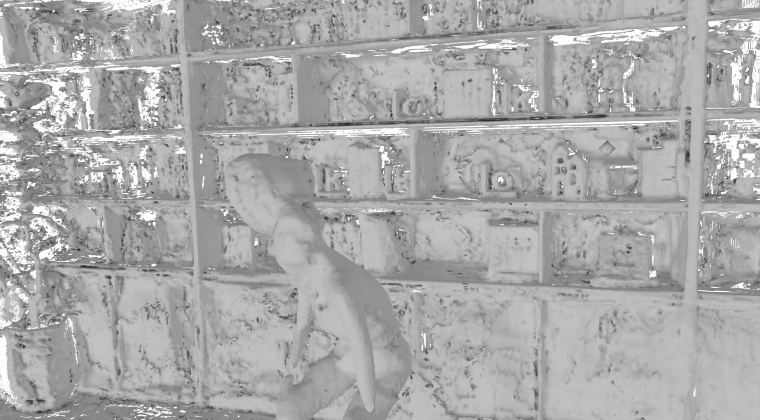}} \\[2pt]
    \rotatebox{90}{\makebox[\mcrealrow]{\footnotesize FreeTimeGS}} & \raisebox{-0.5\height}{\includegraphics[height=\mcrealrow]{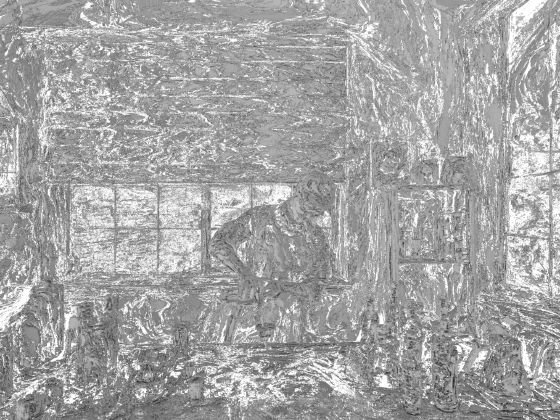}} & \raisebox{-0.5\height}{\includegraphics[height=\mcrealrow]{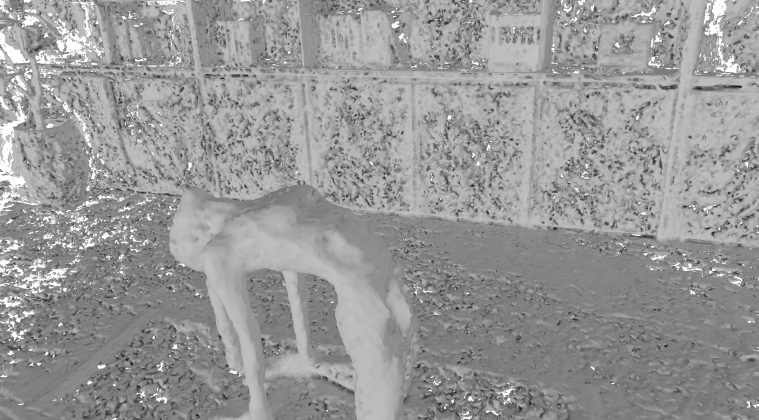}} & \raisebox{-0.5\height}{\includegraphics[height=\mcrealrow]{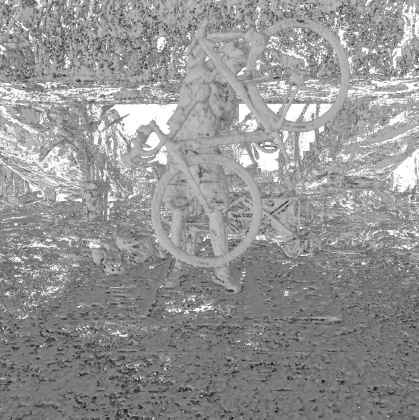}} & \raisebox{-0.5\height}{\includegraphics[height=\mcrealrow]{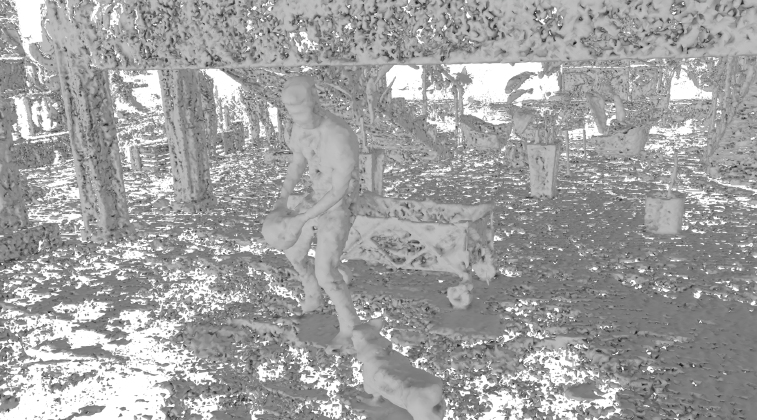}} & \raisebox{-0.5\height}{\includegraphics[height=\mcrealrow]{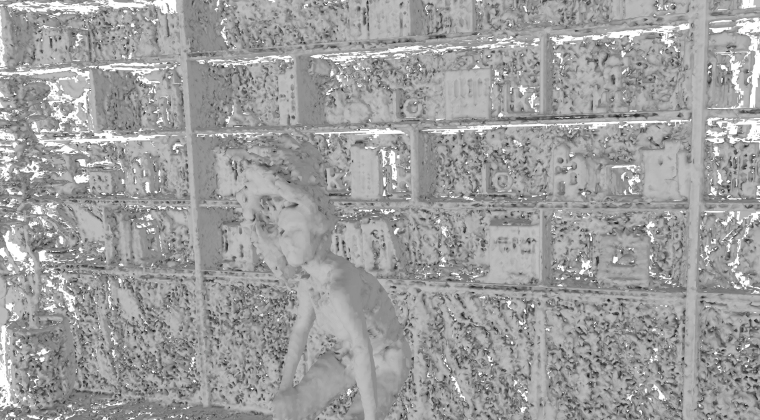}} \\[2pt]
    \rotatebox{90}{\makebox[\mcrealrow]{\footnotesize AST-3DGS}} & \raisebox{-0.5\height}{\includegraphics[height=\mcrealrow]{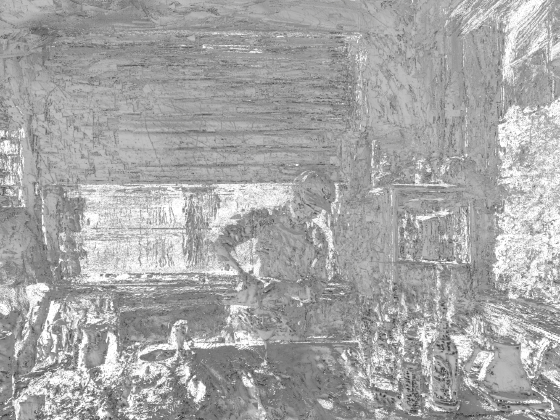}} & \raisebox{-0.5\height}{\includegraphics[height=\mcrealrow]{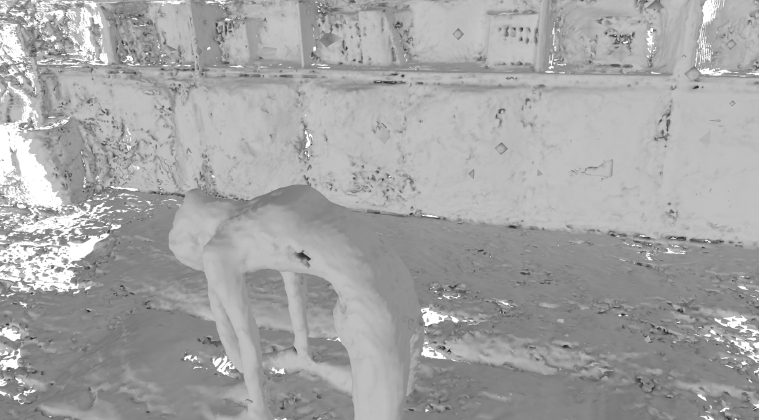}} & \raisebox{-0.5\height}{\includegraphics[height=\mcrealrow]{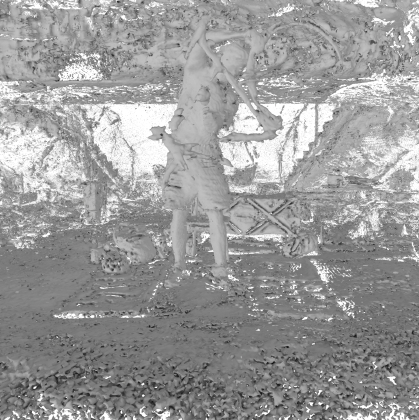}} & \raisebox{-0.5\height}{\includegraphics[height=\mcrealrow]{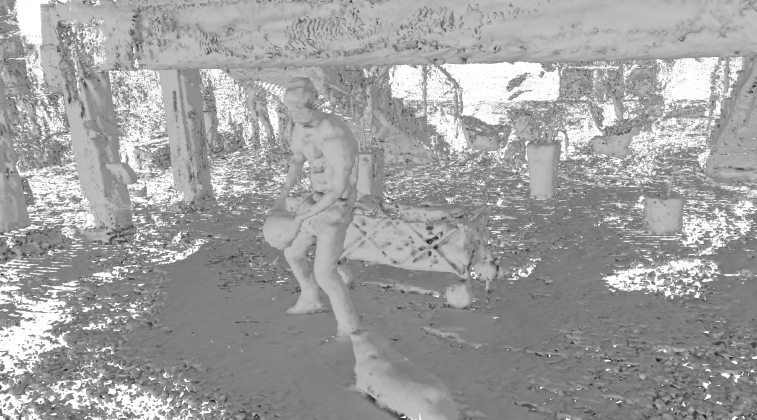}} & \raisebox{-0.5\height}{\includegraphics[height=\mcrealrow]{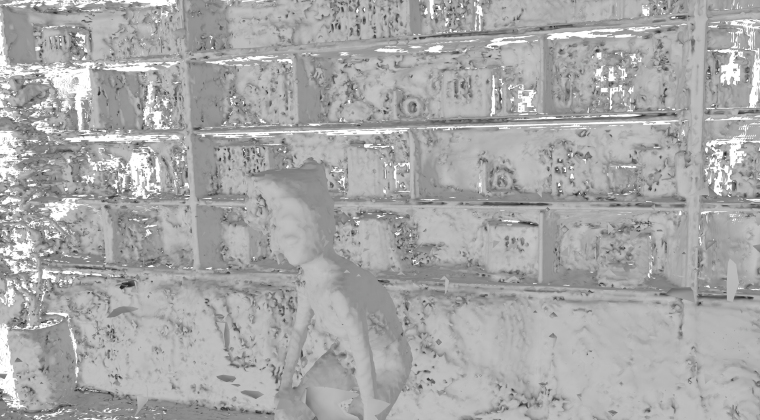}} \\[2pt]
    \rotatebox{90}{\makebox[\mcrealrow]{\footnotesize D-2DGS}} & \raisebox{-0.5\height}{\includegraphics[height=\mcrealrow]{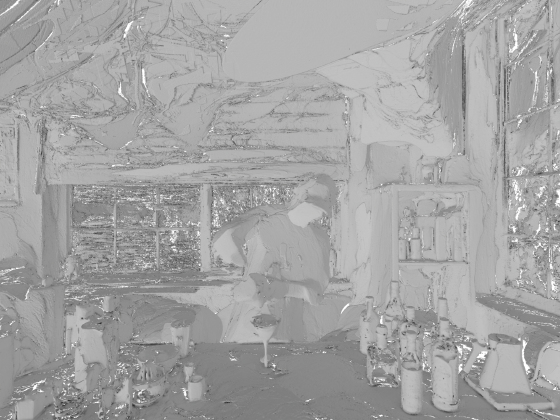}} & \raisebox{-0.5\height}{\includegraphics[height=\mcrealrow]{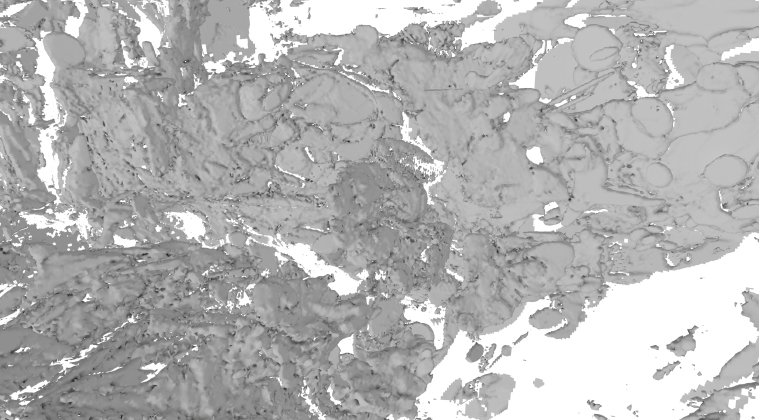}} & \raisebox{-0.5\height}{\includegraphics[height=\mcrealrow]{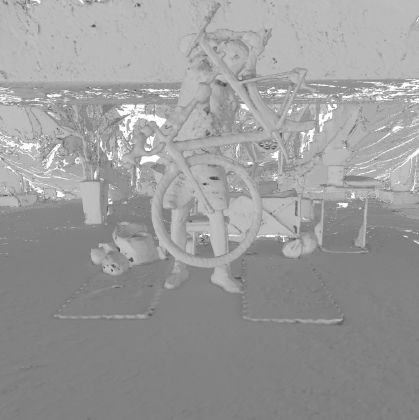}} & \raisebox{-0.5\height}{\includegraphics[height=\mcrealrow]{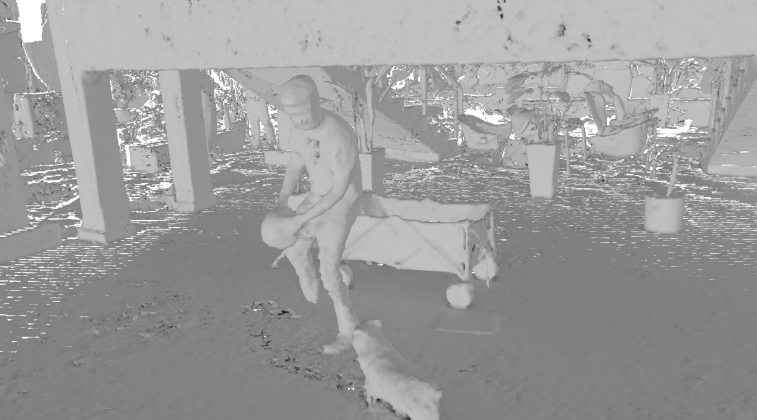}} & \raisebox{-0.5\height}{\includegraphics[height=\mcrealrow]{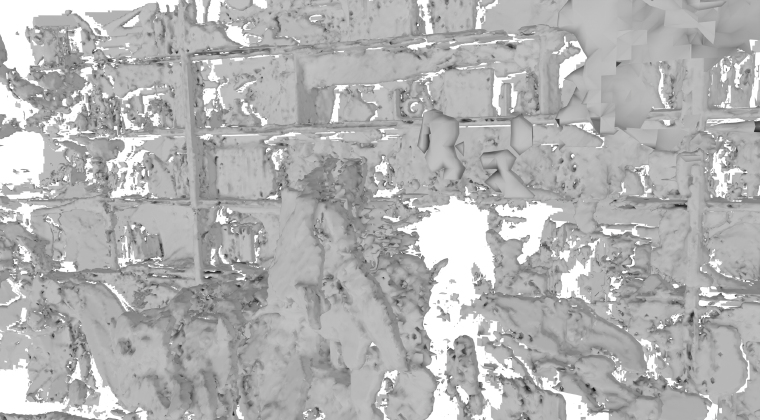}} \\[2pt]
    \rotatebox{90}{\makebox[\mcrealrow]{\footnotesize \ourmethod}} & \raisebox{-0.5\height}{\includegraphics[height=\mcrealrow]{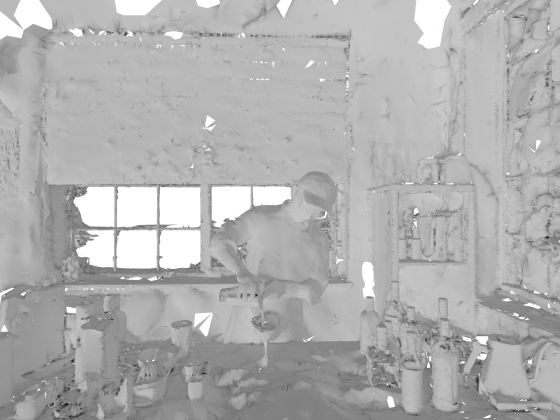}} & \raisebox{-0.5\height}{\includegraphics[height=\mcrealrow]{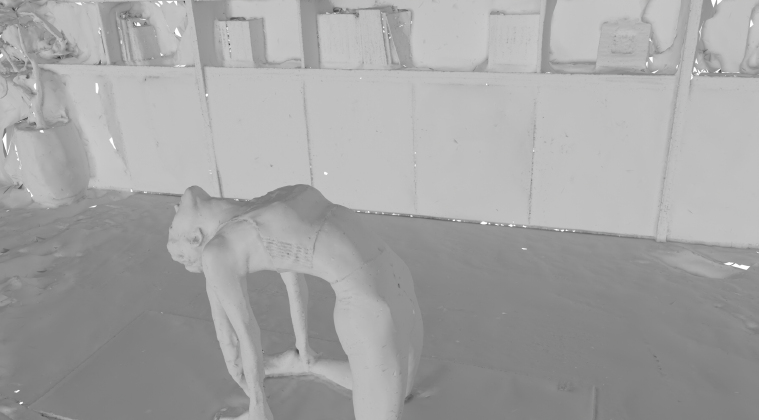}} & \raisebox{-0.5\height}{\includegraphics[height=\mcrealrow]{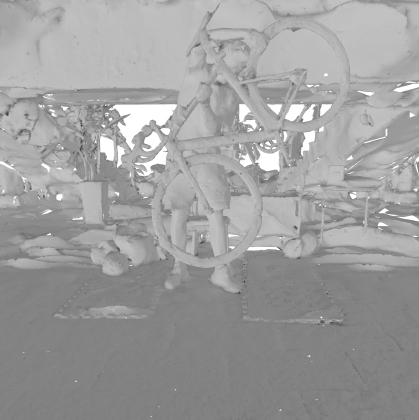}} & \raisebox{-0.5\height}{\includegraphics[height=\mcrealrow]{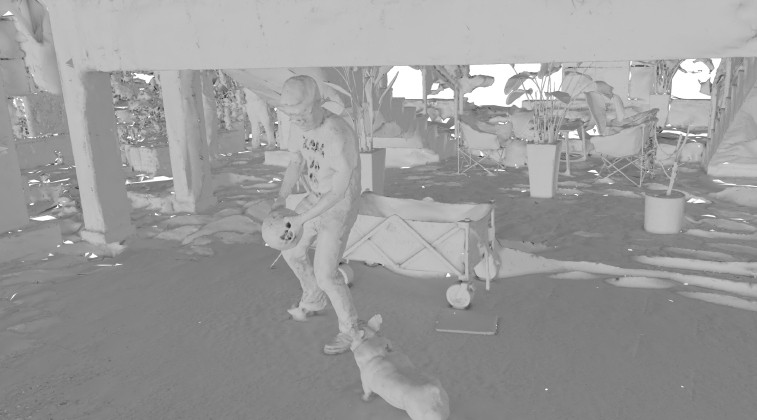}} & \raisebox{-0.5\height}{\includegraphics[height=\mcrealrow]{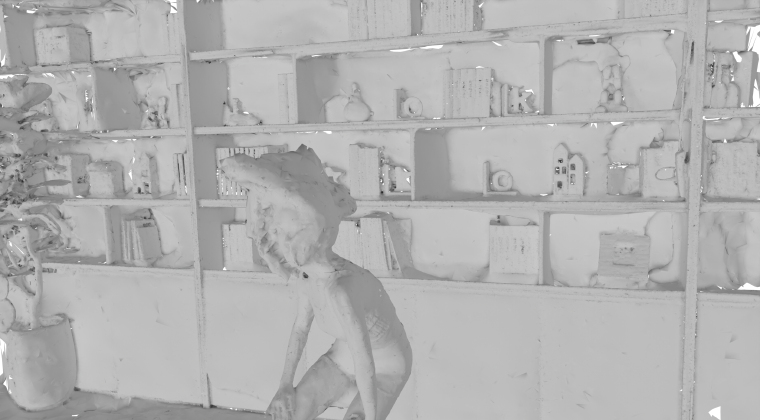}} \\[2pt]
    \rotatebox{90}{\makebox[\mcrealrow]{\footnotesize Test image}} & \raisebox{-0.5\height}{\includegraphics[height=\mcrealrow]{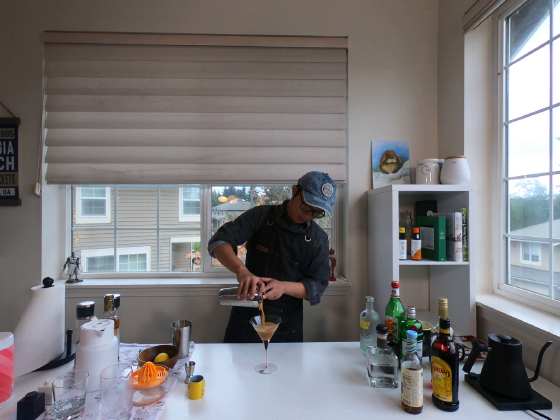}} & \raisebox{-0.5\height}{\includegraphics[height=\mcrealrow]{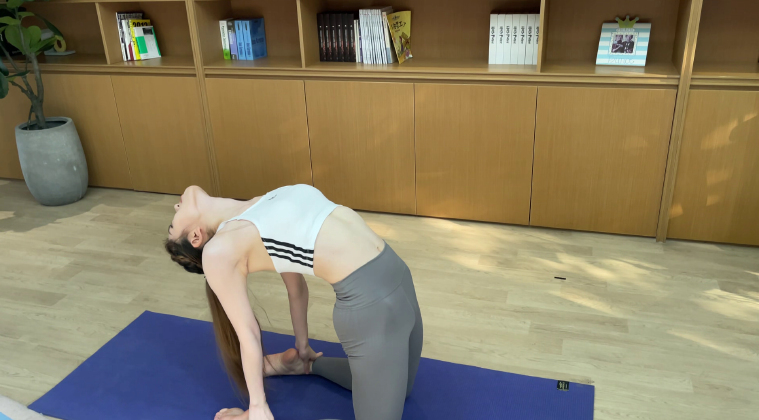}} & \raisebox{-0.5\height}{\includegraphics[height=\mcrealrow]{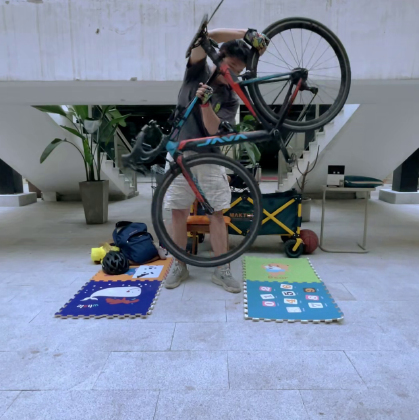}} & \raisebox{-0.5\height}{\includegraphics[height=\mcrealrow]{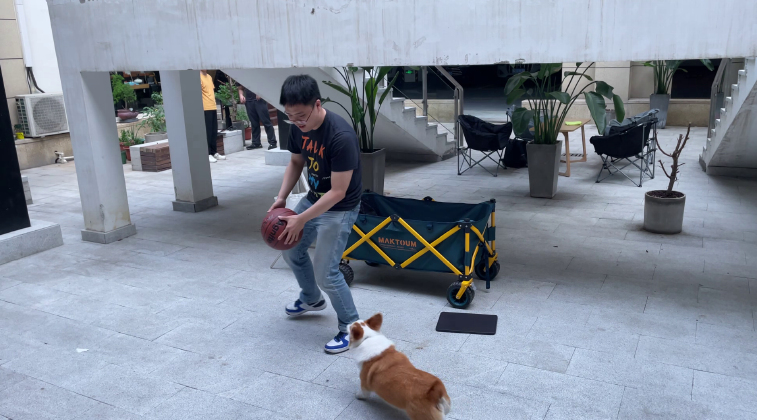}} & \raisebox{-0.5\height}{\includegraphics[height=\mcrealrow]{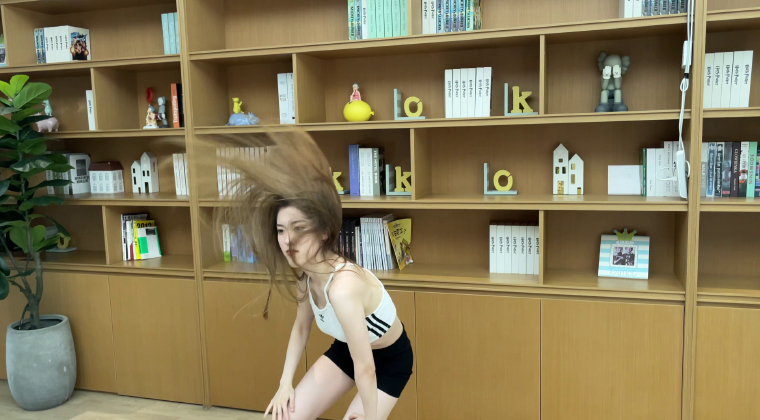}} \\[2pt]
  \end{tabular}
  \caption{Meshes extracted by all methods on the real scenes, seen from the test camera. Please zoom in for details.}
  \label{fig:mesh_comparison_real}
\end{figure*}

%% file: figures/nvs_real.tex
\begin{figure*}[p]
  \centering
  \setlength{\tabcolsep}{1pt}
  \renewcommand{\arraystretch}{0.6}
  \newlength{\nvsrow}\setlength{\nvsrow}{\dimexpr(\linewidth-1.2em-10\tabcolsep)*1000/6329\relax}
  \ifdim\nvsrow>0.95in\setlength{\nvsrow}{0.95in}\fi
  \begin{tabular}{@{}m{1.2em}ccccc@{}}
    & \multicolumn{1}{c}{\footnotesize Cook spinach} & \multicolumn{1}{c}{\footnotesize Cut roasted beef} & \multicolumn{1}{c}{\footnotesize Bike1} & \multicolumn{1}{c}{\footnotesize Corgi1} & \multicolumn{1}{c}{\footnotesize Hair1} \\
    \rotatebox{90}{\makebox[\nvsrow]{\footnotesize 4DGS}} & \raisebox{-0.5\height}{\includegraphics[height=\nvsrow]{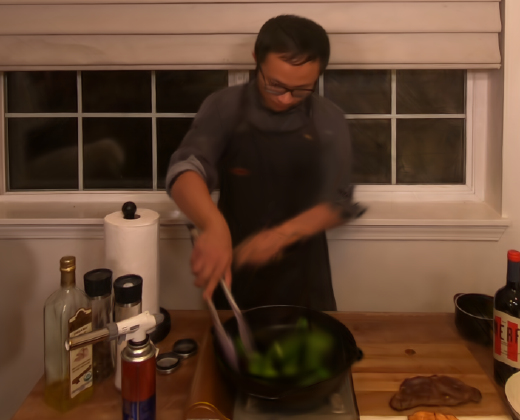}} & \raisebox{-0.5\height}{\includegraphics[height=\nvsrow]{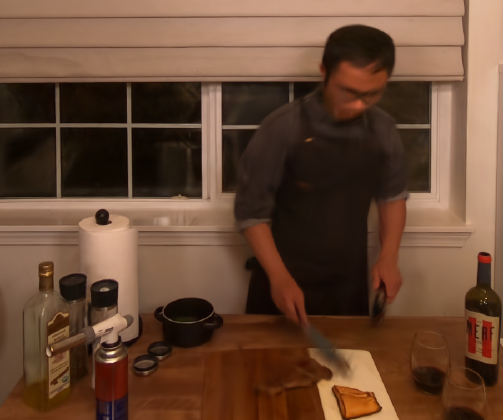}} & \raisebox{-0.5\height}{\includegraphics[height=\nvsrow]{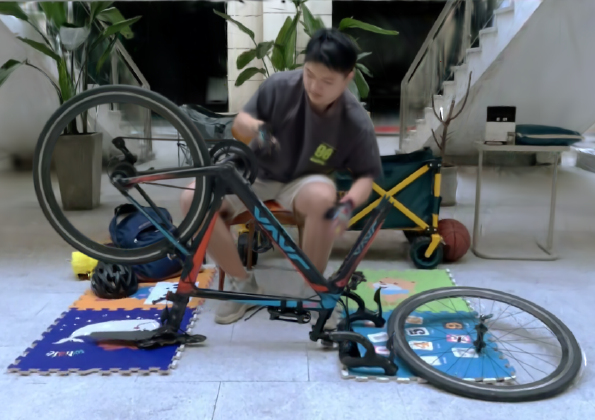}} & \raisebox{-0.5\height}{\includegraphics[height=\nvsrow]{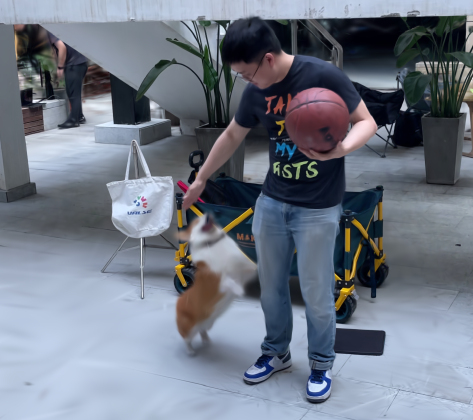}} & \raisebox{-0.5\height}{\includegraphics[height=\nvsrow]{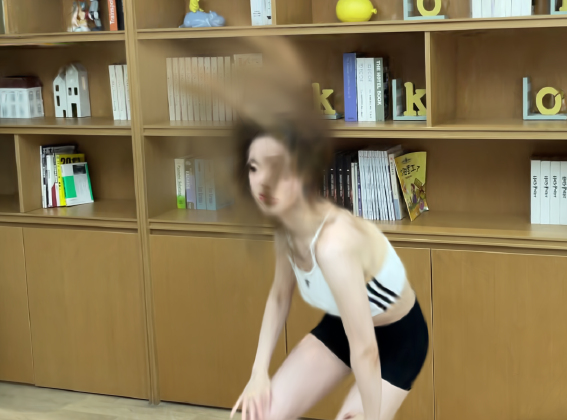}} \\[2pt]
    \rotatebox{90}{\makebox[\nvsrow]{\footnotesize Deformable-3DGS}} & \raisebox{-0.5\height}{\includegraphics[height=\nvsrow]{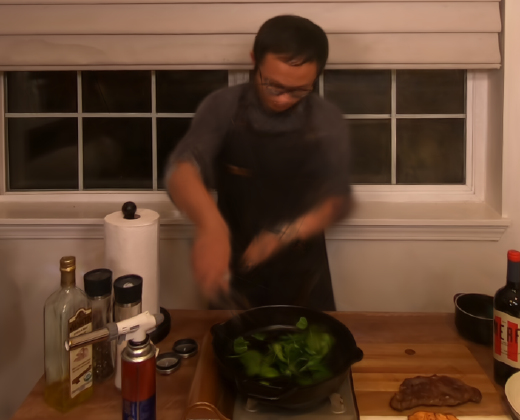}} & \raisebox{-0.5\height}{\includegraphics[height=\nvsrow]{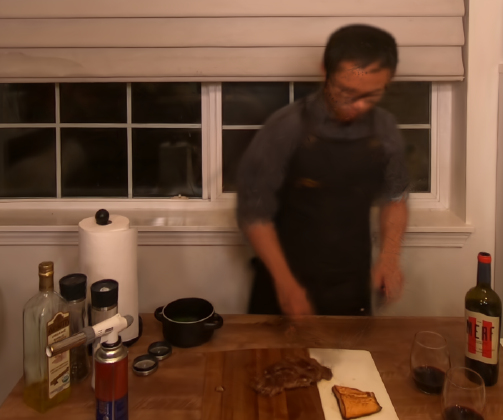}} & \raisebox{-0.5\height}{\includegraphics[height=\nvsrow]{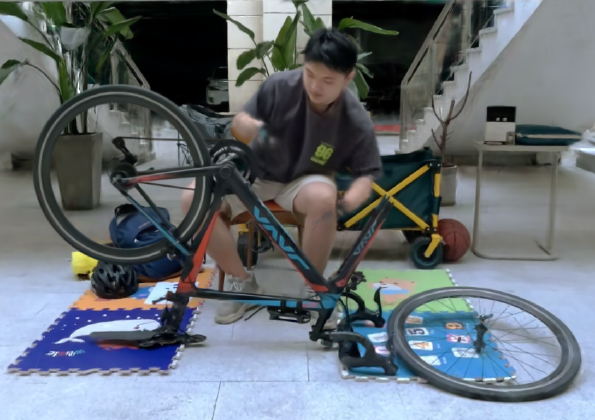}} & \raisebox{-0.5\height}{\includegraphics[height=\nvsrow]{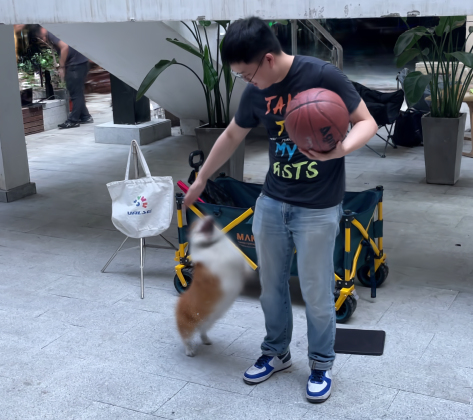}} & \raisebox{-0.5\height}{\includegraphics[height=\nvsrow]{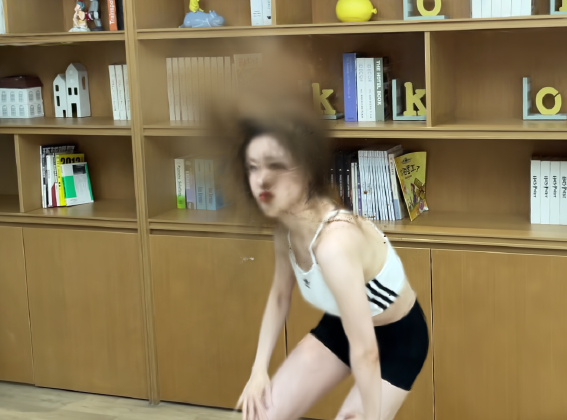}} \\[2pt]
    \rotatebox{90}{\makebox[\nvsrow]{\footnotesize SC-GS}} & \raisebox{-0.5\height}{\includegraphics[height=\nvsrow]{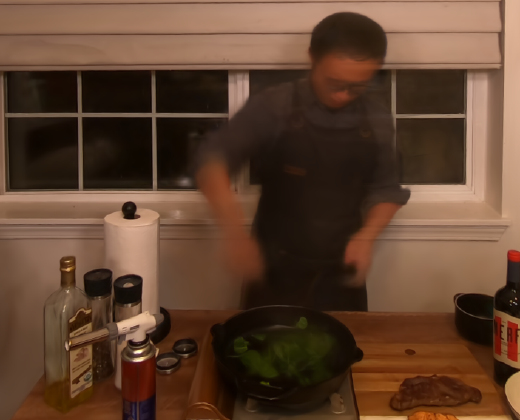}} & \raisebox{-0.5\height}{\includegraphics[height=\nvsrow]{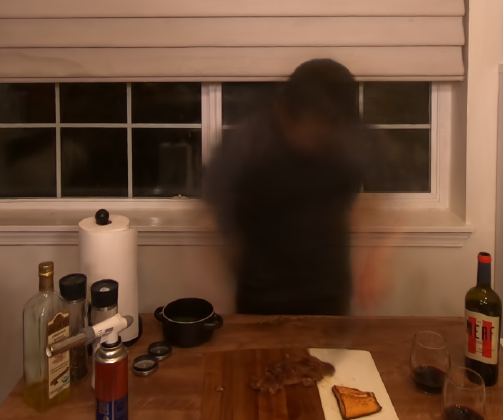}} & \raisebox{-0.5\height}{\includegraphics[height=\nvsrow]{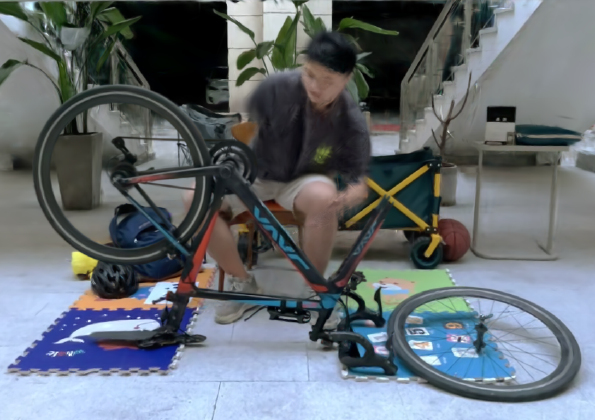}} & \raisebox{-0.5\height}{\includegraphics[height=\nvsrow]{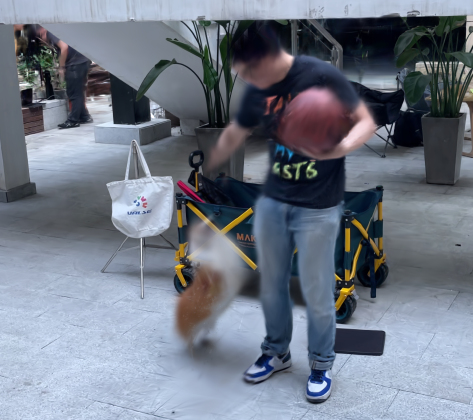}} & \raisebox{-0.5\height}{\includegraphics[height=\nvsrow]{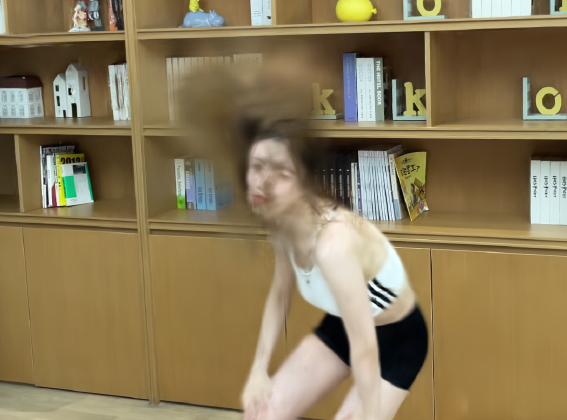}} \\[2pt]
    \rotatebox{90}{\makebox[\nvsrow]{\footnotesize FreeTimeGS}} & \raisebox{-0.5\height}{\includegraphics[height=\nvsrow]{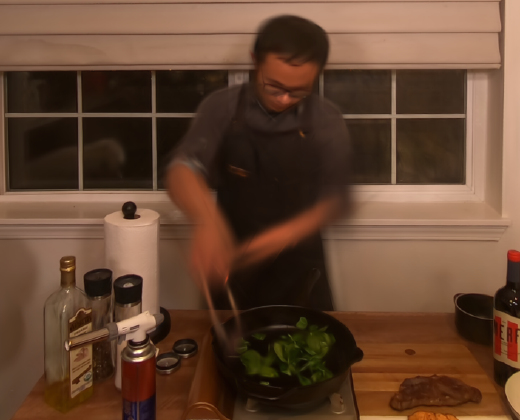}} & \raisebox{-0.5\height}{\includegraphics[height=\nvsrow]{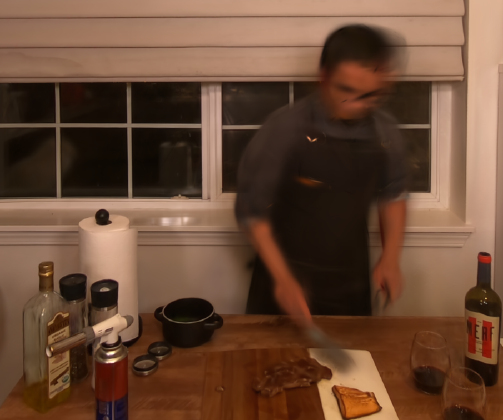}} & \raisebox{-0.5\height}{\includegraphics[height=\nvsrow]{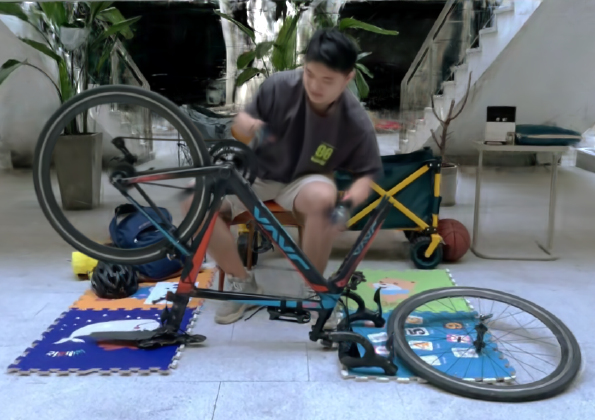}} & \raisebox{-0.5\height}{\includegraphics[height=\nvsrow]{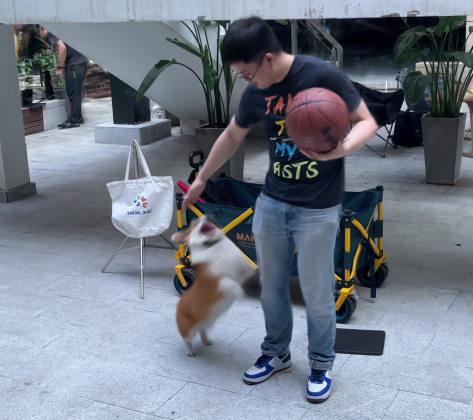}} & \raisebox{-0.5\height}{\includegraphics[height=\nvsrow]{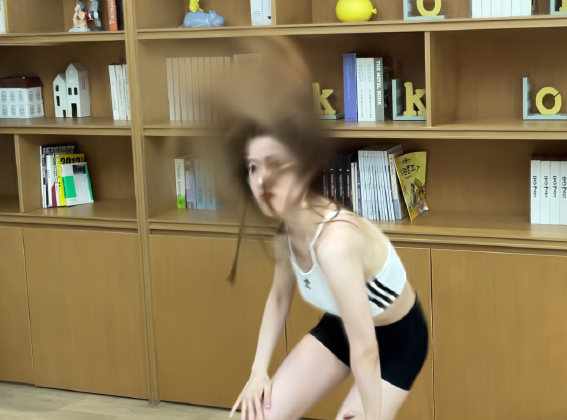}} \\[2pt]
    \rotatebox{90}{\makebox[\nvsrow]{\footnotesize AST-3DGS}} & \raisebox{-0.5\height}{\includegraphics[height=\nvsrow]{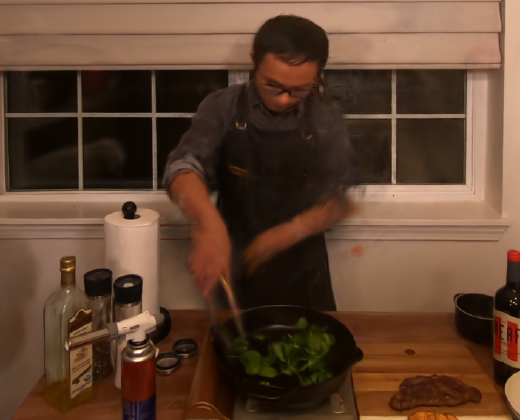}} & \raisebox{-0.5\height}{\includegraphics[height=\nvsrow]{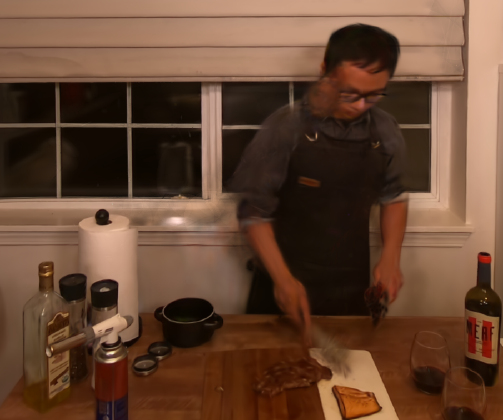}} & \raisebox{-0.5\height}{\includegraphics[height=\nvsrow]{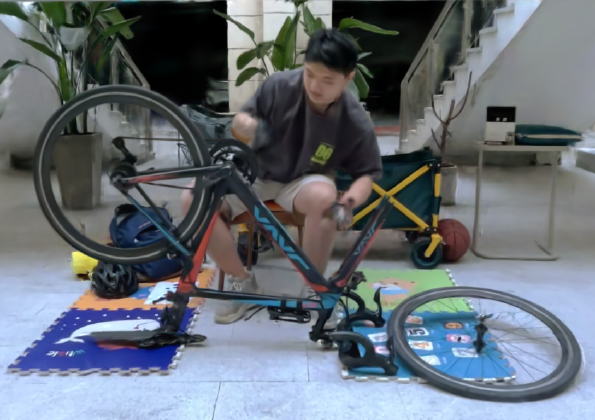}} & \raisebox{-0.5\height}{\includegraphics[height=\nvsrow]{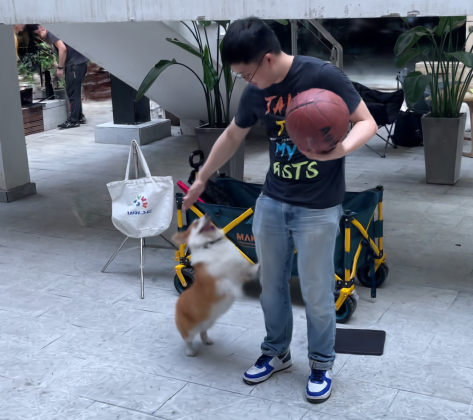}} & \raisebox{-0.5\height}{\includegraphics[height=\nvsrow]{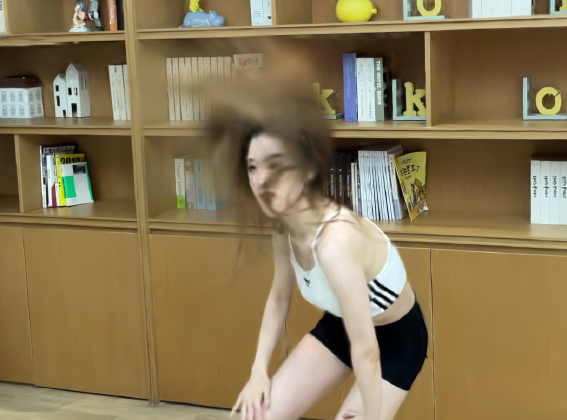}} \\[2pt]
    \rotatebox{90}{\makebox[\nvsrow]{\footnotesize D-2DGS}} & \raisebox{-0.5\height}{\includegraphics[height=\nvsrow]{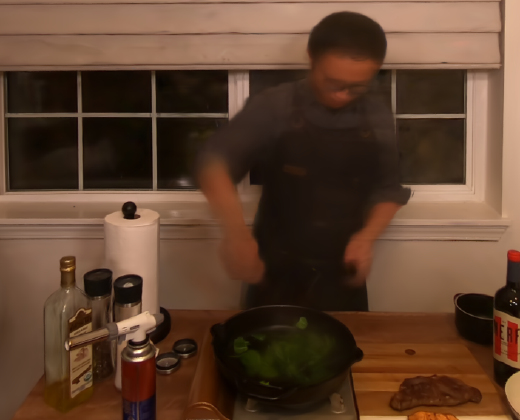}} & \raisebox{-0.5\height}{\includegraphics[height=\nvsrow]{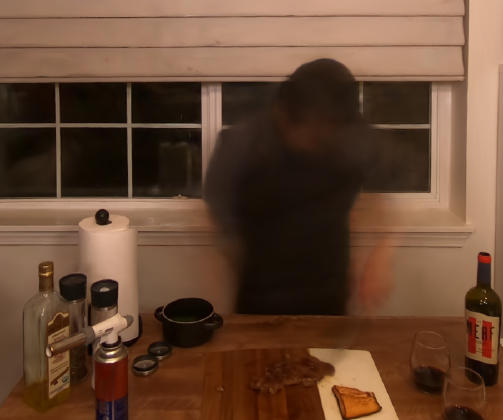}} & \raisebox{-0.5\height}{\includegraphics[height=\nvsrow]{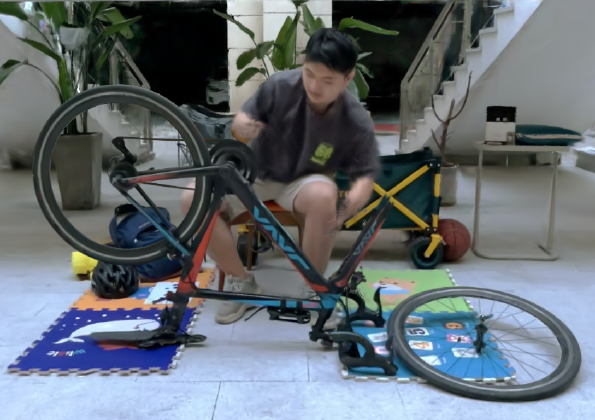}} & \raisebox{-0.5\height}{\includegraphics[height=\nvsrow]{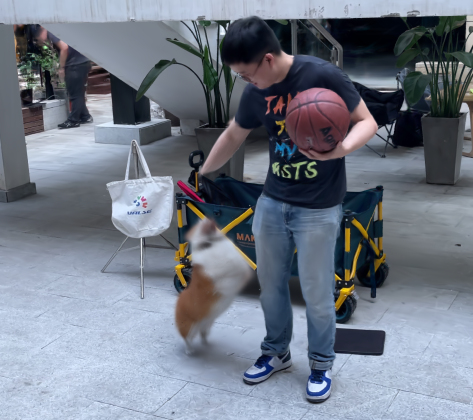}} & \raisebox{-0.5\height}{\includegraphics[height=\nvsrow]{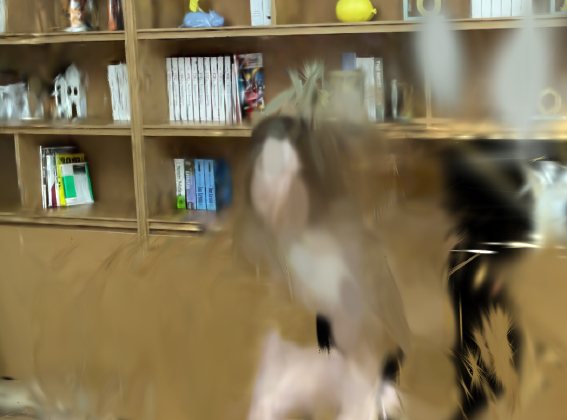}} \\[2pt]
    \rotatebox{90}{\makebox[\nvsrow]{\footnotesize \ourmethod}} & \raisebox{-0.5\height}{\includegraphics[height=\nvsrow]{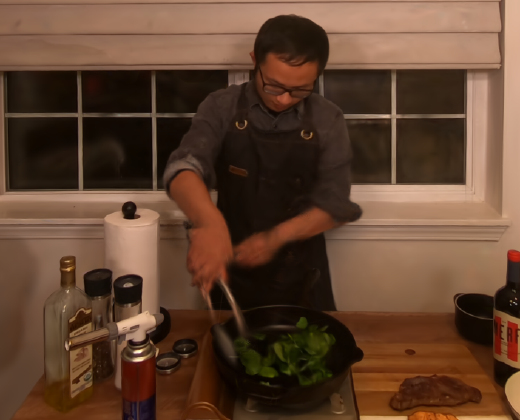}} & \raisebox{-0.5\height}{\includegraphics[height=\nvsrow]{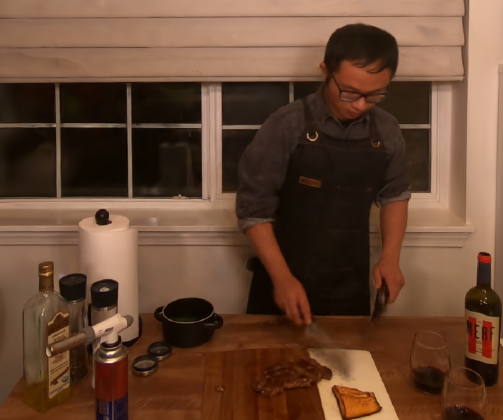}} & \raisebox{-0.5\height}{\includegraphics[height=\nvsrow]{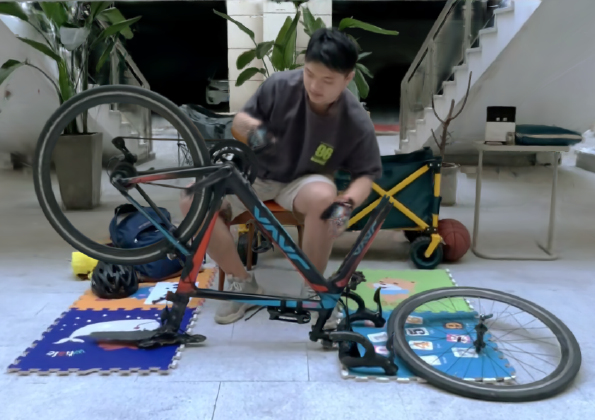}} & \raisebox{-0.5\height}{\includegraphics[height=\nvsrow]{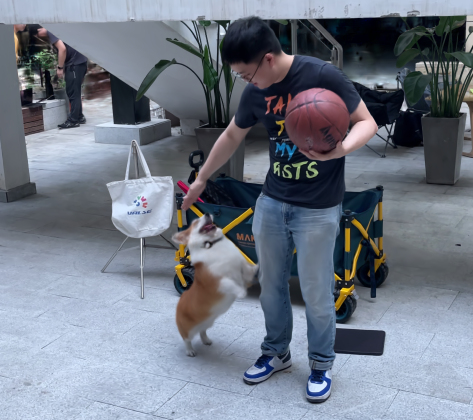}} & \raisebox{-0.5\height}{\includegraphics[height=\nvsrow]{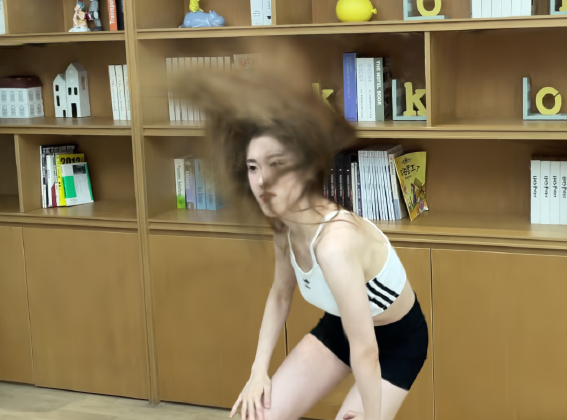}} \\[2pt]
    \rotatebox{90}{\makebox[\nvsrow]{\footnotesize Ground truth}} & \raisebox{-0.5\height}{\includegraphics[height=\nvsrow]{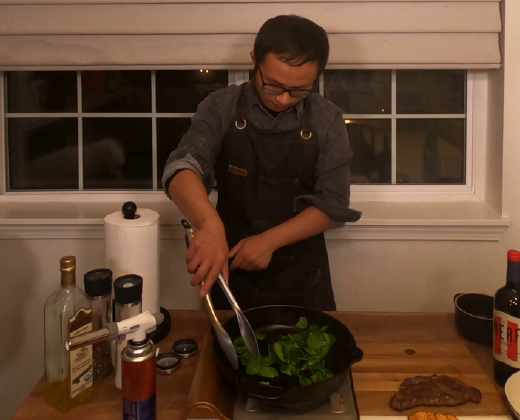}} & \raisebox{-0.5\height}{\includegraphics[height=\nvsrow]{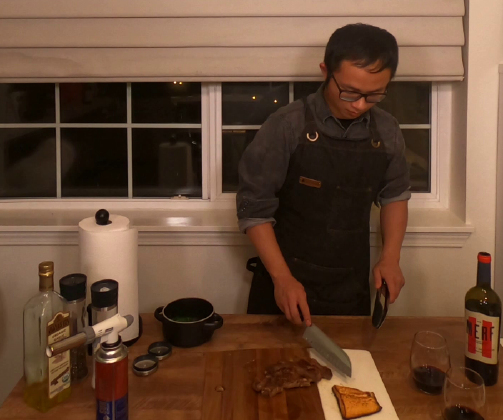}} & \raisebox{-0.5\height}{\includegraphics[height=\nvsrow]{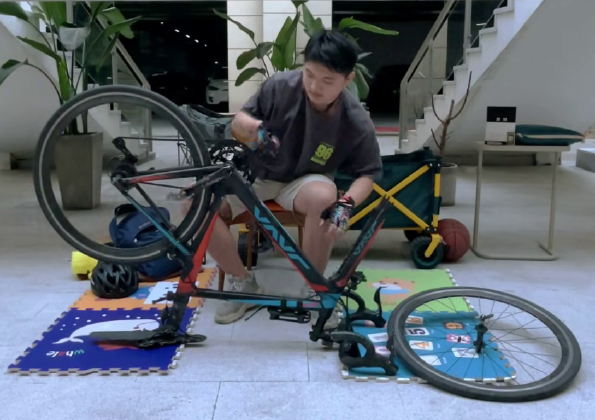}} & \raisebox{-0.5\height}{\includegraphics[height=\nvsrow]{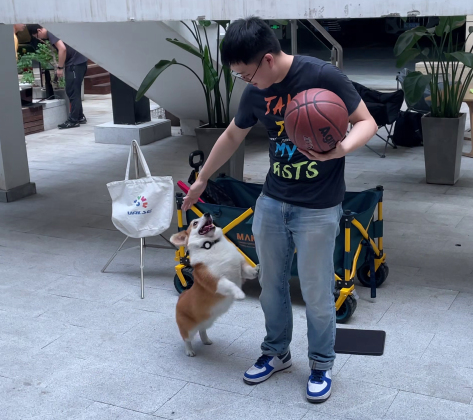}} & \raisebox{-0.5\height}{\includegraphics[height=\nvsrow]{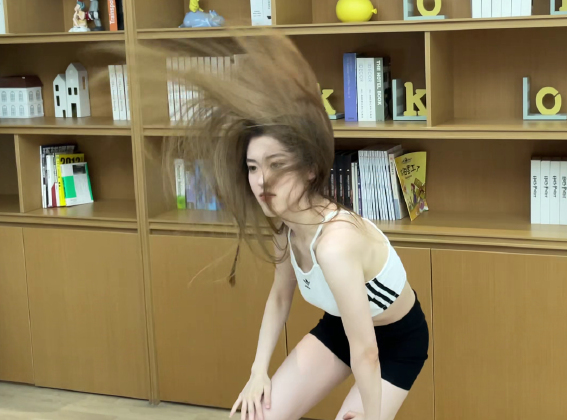}} \\[2pt]
  \end{tabular}
  \caption{Novel views rendered by all methods on the real scenes, seen from the test camera. Please zoom in for details.}
  \label{fig:nvs_real}
\end{figure*}

%% file: figures/meshes_over_time.tex
\begin{figure*}[p]
  \centering
  \includegraphics[width=\linewidth]{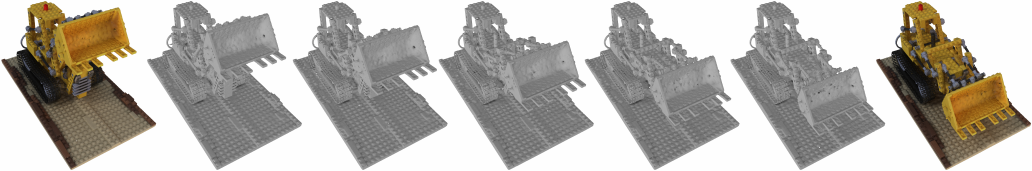}\\[27pt]
  \includegraphics[width=\linewidth]{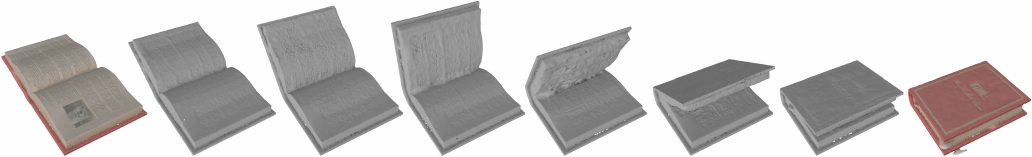}\\[27pt]
  \includegraphics[width=\linewidth]{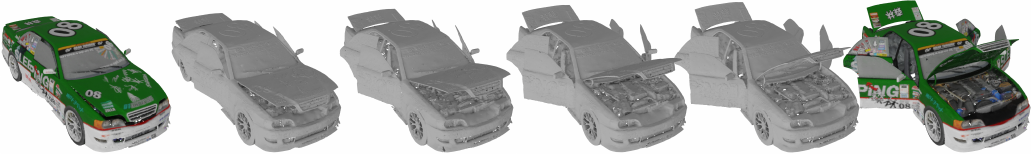}\\[27pt]
  \includegraphics[width=\linewidth]{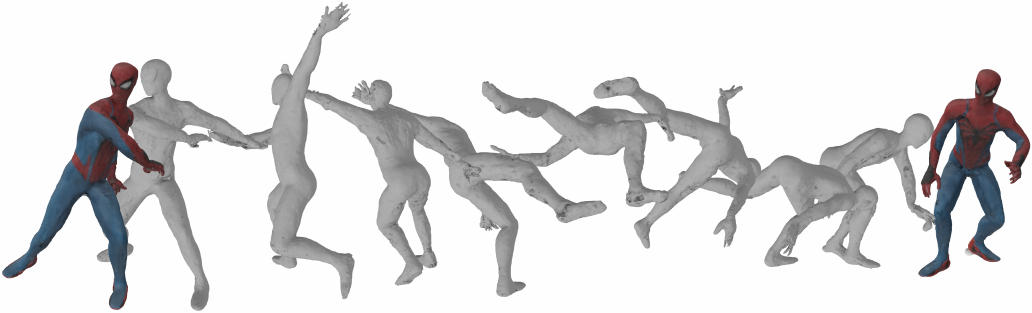}\\[27pt]
  \includegraphics[width=\linewidth]{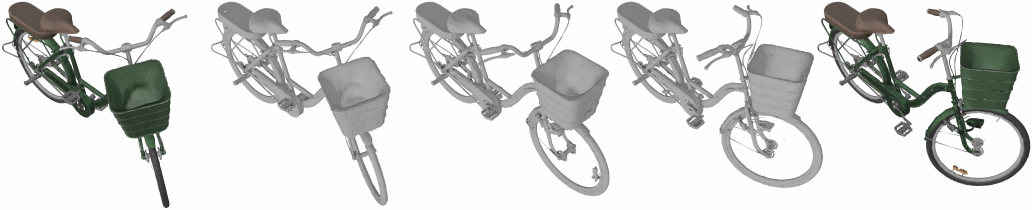}\\[2pt]
  \begin{tikzpicture}
    \draw[line width=1.2pt, -{Latex[length=6pt, width=5pt]}] (0, 0) -- (\linewidth, 0)
      node[midway, fill=white, inner xsep=4pt, inner ysep=0pt, font=\footnotesize] {Time};
  \end{tikzpicture}
  \caption{\ourmethod's meshes over time on five more \benchmark scenes, with the first and last frames in vertex colors and the frames between in gray.}
  \label{fig:meshes_over_time}
\end{figure*}